\documentclass[a4paper,fleqn]{cas-dc}

\usepackage[authoryear,longnamesfirst]{natbib}
\usepackage{subcaption}

\def\tsc#1{\csdef{#1}{\textsc{\lowercase{#1}}\xspace}}
\tsc{WGM}
\tsc{QE}

\begin{document}
\let\WriteBookmarks\relax
\def\floatpagepagefraction{1}
\def\textpagefraction{.001}

\shorttitle{Domain-aware Triplet loss in Domain Generalization}    

\shortauthors{Kaiyu Guo, Brian Lovell}  

\title [mode = title]{Domain-aware Triplet loss in Domain Generalization}  



%

\author[1]{Kaiyu Guo}[orcid=0000-0002-4187-2839]



\ead{kaiyu.guo@uqconnect.edu.au}


\credit{<Credit authorship details>}

\affiliation[1]{organization={University of Queensland},
            addressline={}, 
            city={Brisbane},
            postcode={4072}, 
            state={Queensland},
            country={Australia}}

\author[2]{Brian Lovell}

\fnmark[2]

\ead{lovell@itee.uq.edu.au}

\ead[url]{https://staff.itee.uq.edu.au/lovell/}

\credit{}

\affiliation[2]{organization={University of Queensland},
            addressline={}, 
            city={Brisbane},
            postcode={4072}, 
            state={Queensland},
            country={Australia}}

\cortext[1]{Brian Lovell}



\begin{abstract}
Despite much progress being made in the field of object recognition with the advances of deep learning, there are still several factors negatively affecting the performance of deep learning models. Domain shift is one of these factors and is caused by discrepancies in the distributions of the testing and training data. In this paper, we focus on the problem of compact feature clustering in domain generalization to help optimize the embedding space from multi-domain data. We design a domain-aware triplet loss for domain generalization to help the model to not only cluster similar semantic features, but also to disperse features arising from the domain. Unlike previous methods focusing on distribution alignment, our algorithm is designed to disperse domain information in the embedding space. The basic idea is motivated based on the assumption that embedding features can be clustered based on domain information, which is mathematically and empirically supported in this paper.

In addition, during our exploration of feature clustering in domain generalization, we note that factors affecting the convergence of metric learning loss in domain generalization are more important than the pre-defined domains. To solve this issue, we utilize two methods to normalize the embedding space, reducing the internal covariate shift of the embedding features. The ablation study demonstrates the effectiveness of our algorithm. Moreover, the experiments on the benchmark datasets, including PACS, VLCS and Office-Home, show that our method outperforms related methods focusing on domain discrepancy. In particular, our results on RegnetY-16 are significantly better than state-of-the-art methods on the benchmark datasets. Our code will be released at \href{https://github.com/workerbcd/DCT}{https://github.com/workerbcd/DCT}.
\end{abstract}




\maketitle

\section{Introduction}
\label{sec: Introduction}
With the development of deep learning, many computer vision tasks have achieved astonishing progress, such as image classification \cite{resnet,vit} and object detection \cite{fastrcnn}. However, the gap between training domains and testing domains may have significant negative impact on the model --- this is called domain shift. To deal with this issue, research on cross-domain problems, such as domain adaptation, is a popular research theme. In addition, with the increasing practical application of deep learning algorithms, the need for applying models to unseen domains is also increasing. The field of domain generalization, which has evolved from domain adaptation, has been proposed to respond to these requirements.

\begin{figure}
\centering 
\includegraphics[width=0.45\textwidth]{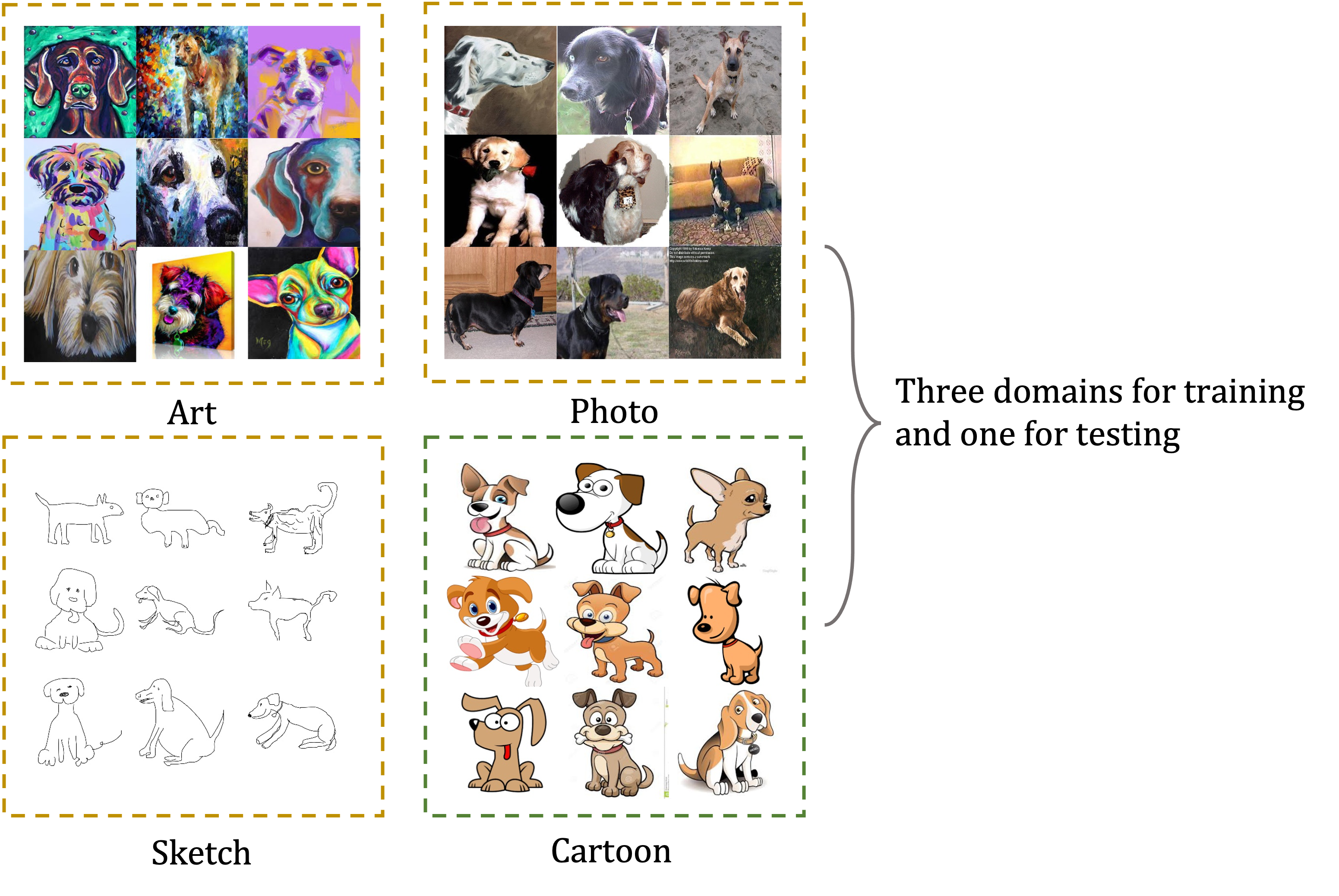} 
\caption{Some examples from the domain generalization dataset PACS\cite{PACS} which comprises four domains. For the domain generalization task in PACS, each model is trained on three source domains and then tested on the remaining source domain of the PACS dataset in a leave one out strategy.} 
\label{Fig1} 
\end{figure}

Figure \ref{Fig1} shows the PACS\cite{PACS} dataset designed for domain generalization and introduces the procedure for domain generalization. Domain generalization aims to learn a model from several source domains so the model will generalize well to an unseen domain \cite{domaingen}. There are many algorithms proposed to solve the problem of domain generalization\cite{mixstyle,swad,coral,DICA}. Generally, the main insight behind these methods is to learn domain-invariant models or features, where the semantic information should be unified in every domain.  

Many researchers try to align the distributions with different metrics~\cite{CDT, MSL} to assist in clustering features from different domains. In our experiments, the different domain distributions can be visualized as different domain clusters. Even though the alignment process can merge the source domains in the embedding space, the unseen domain may still be distant from the merged domain.  In this paper, the motivation is not to align different domains but to weaken the effect of the domain information. Based on the explanation and visualization of the embedding space in domain generalization, a domain-aware triplet loss is proposed to scatter features from the same domain and to cluster features with the same semantic information. In addition, we explore the invalidation of feature clustering in domain generalization and propose two solutions to solve the issue of non-convergnece in the training process.  By both empirically clustering features based on semantic information and also scattering the features based on domain information, our Domain-Class Triplet (DCT) loss outperforms other algorithms focusing primarily on domain discrepancy in domain generalization. \par
 The main contributions of our paper can be divided into three categories:
\begin{enumerate}
    \item This paper presents a novel approach to feature clustering in domain generalization. We believe that the pretrained model without sufficient semantic priors may cause domain clusters in the embedding space. We visualize the feature distribution influenced by the domain discrepancy in domain generalization to support our explanation.  
    \item Based on our exploration, we propose a simple but effective pair mining method applied to triplet loss, which helps disperse domain-clusters  to reduce the domain discrepancy in domain generalization. In addition, we explain how internal covariate shift (ICS) affects the convergence of triplet loss in domain generalization and propose two ways to solve this issue.
    \item We test our theory and the effectiveness of our algorithm. The ablation study supports the validity of our algorithm. Moreover, comparison with other methods shows that our algorithm outperforms algorithms based primarily on domain discrepancy for domain generalization.  We will release our code upon acceptance.
\end{enumerate}

\section{Related Works}
\label{sec: Related Works}
\textbf{Domain Generalization}
 Domain Generalization is a popular task to handle the domain shift issue in deep learning models. Unlike the transductive transfer learning methods \cite{pan}, not only the labels but also any other information about the target domain is not available in the training stage of domain generalization. Generally, domain generalization algorithms can be divided into three categories \cite{dgsurvey}: 1) data manipulation, 2) representation learning, and 3) learning strategy.

  \noindent \textbf{Data manipulation:} methods \cite{crossgrad,mixstyle} mostly utilize data augmentation or data generation to help enrich or unify the distribution in the training data. The limitation of these methods is the performance of the data-generating methods which are expected to mix the domain information and retain the semantic information.
  
 \noindent \textbf{Representation learning:} methods, adversarial learning \cite{adv1,adv2}, invariant risk minimization \cite{IRM}, kernel methods \cite{DMA, mmd} and feature disentangling methods \cite{DG-net-PP}  have all been proposed. Such methods focus on generating invariant feature representations to different domains.
 
\noindent \textbf{Learning strategy:} methods include meta-learning \cite{steam,selfbalance}, gradient operation \cite{rsc,fishr}, self-supervised learning \cite{selfreg}, etc.

These methods are designed to improve model generalization via different learning strategies. Actually, all the methods mentioned above focus more on the domain discrepancy in domain generalization. Recently, a method with dense stochastic weight
averaging \cite{swad} is also proposed to help improve the performance of domain generalization. Unlike other methods in domain generalization, this method aims at searching for a robust risk minimization with flat minima using SWA \cite{swa}.  \par

\textbf{Contrastive Learning} 
Contrastive learning~\cite{contrastiveloss} is a research topic in the field of deep metric learning. The general insight behind contrastive learning is to minimize the distance among positive pairs and maximize the distance among negative pairs. With the development of face recognition, algorithms like triplet loss \cite{tripletloss} and center loss ~\cite{centerloss} were proposed to solve the N-pair problems in contrastive learning. Then, instance discrimination\cite{instancediscrimination} was proposed and is widely used in self-supervised learning tasks.  

In domain generalization research, there are several algorithms exploiting contrastive learning~\cite{CDT, pcl}. CDT~\cite{CDT} utilizes Mahalanobis distance to align the source domains in triplet loss and PCL~\cite{pcl} used a proxy-based contrastive loss with two MLP layers. Unlike these methods, the method we propose does not consider the implicit distribution alignment or require more linear layers in the training phase, which makes our method more interpretable and also adaptable to different backbones.
\section{Motivation}
\label{sec: Motivation}
\begin{figure*}
\subcaptionbox{}{
\begin{minipage}[b]{0.23\linewidth}
\includegraphics[width=0.8\linewidth,trim={5cm 2cm 2cm 4cm},clip]{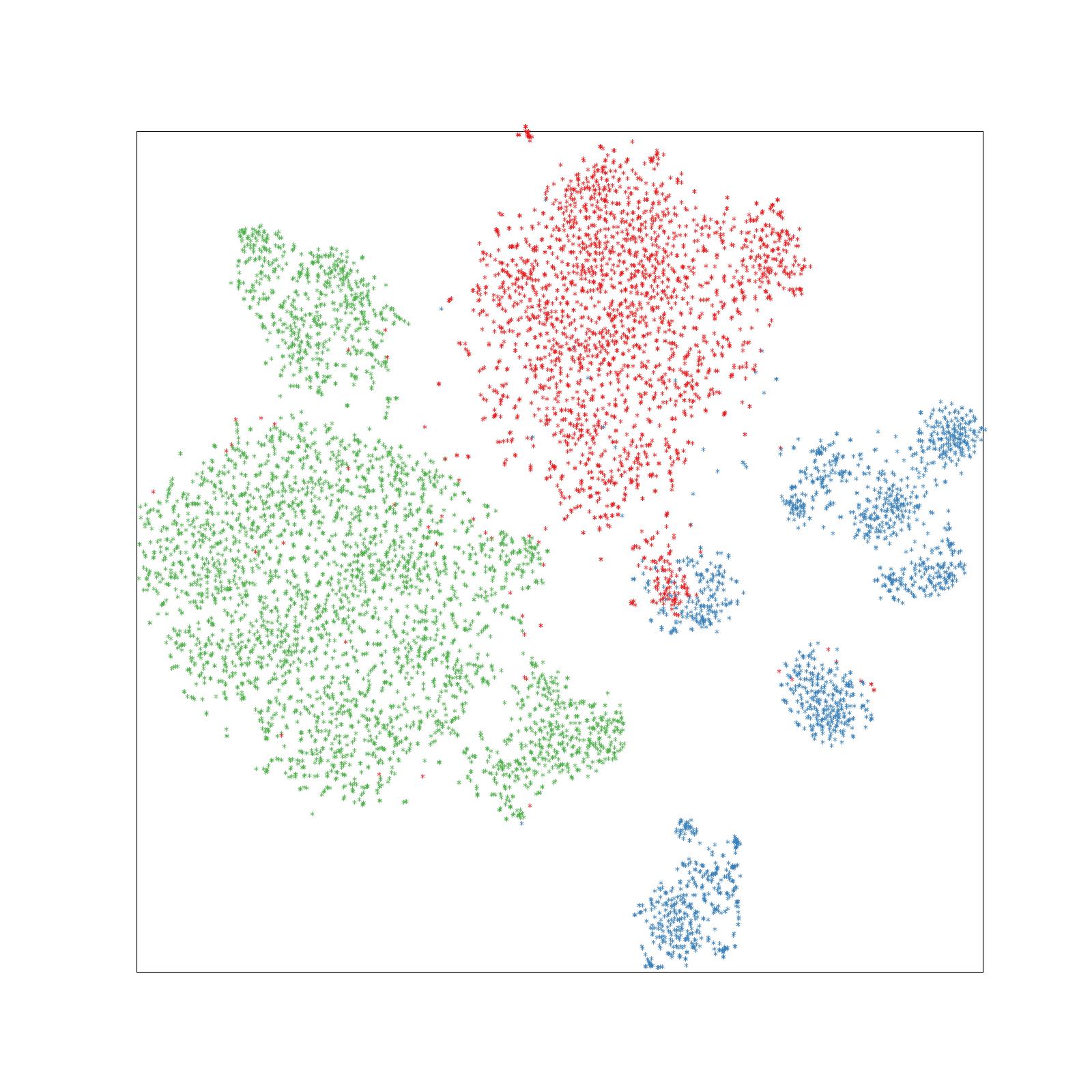}\vspace{1pt}
\includegraphics[width=0.8\linewidth,trim={0cm 0cm 0cm 0cm},clip]{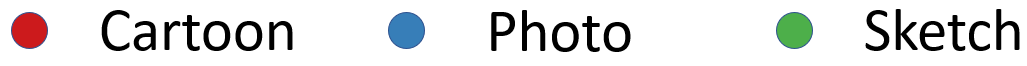}\vspace{1pt}
\includegraphics[width=0.8\linewidth,trim={5cm 2cm 2cm 4cm},clip]{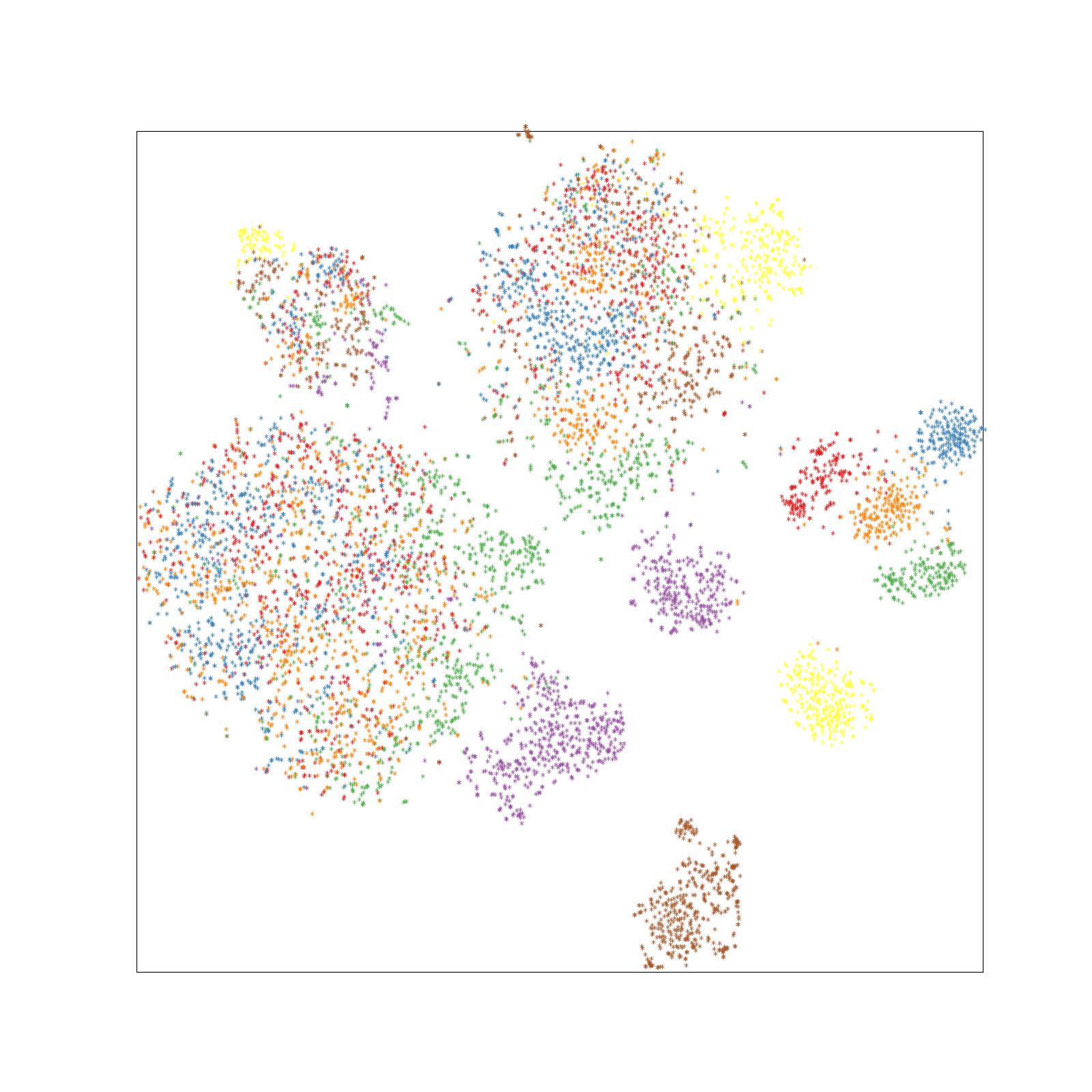}\vspace{1pt}
\includegraphics[width=0.8\linewidth,trim={0cm 0cm 0cm 0cm},clip]{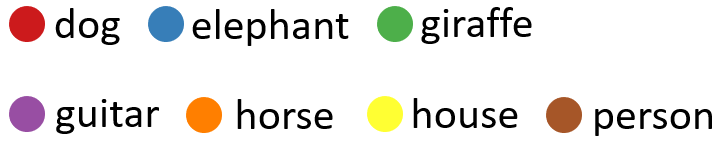}
\end{minipage}}
\subcaptionbox{}{
\begin{minipage}[b]{0.23\linewidth}
\includegraphics[width=0.8\linewidth,trim={5cm 2cm 2cm 4cm},clip]{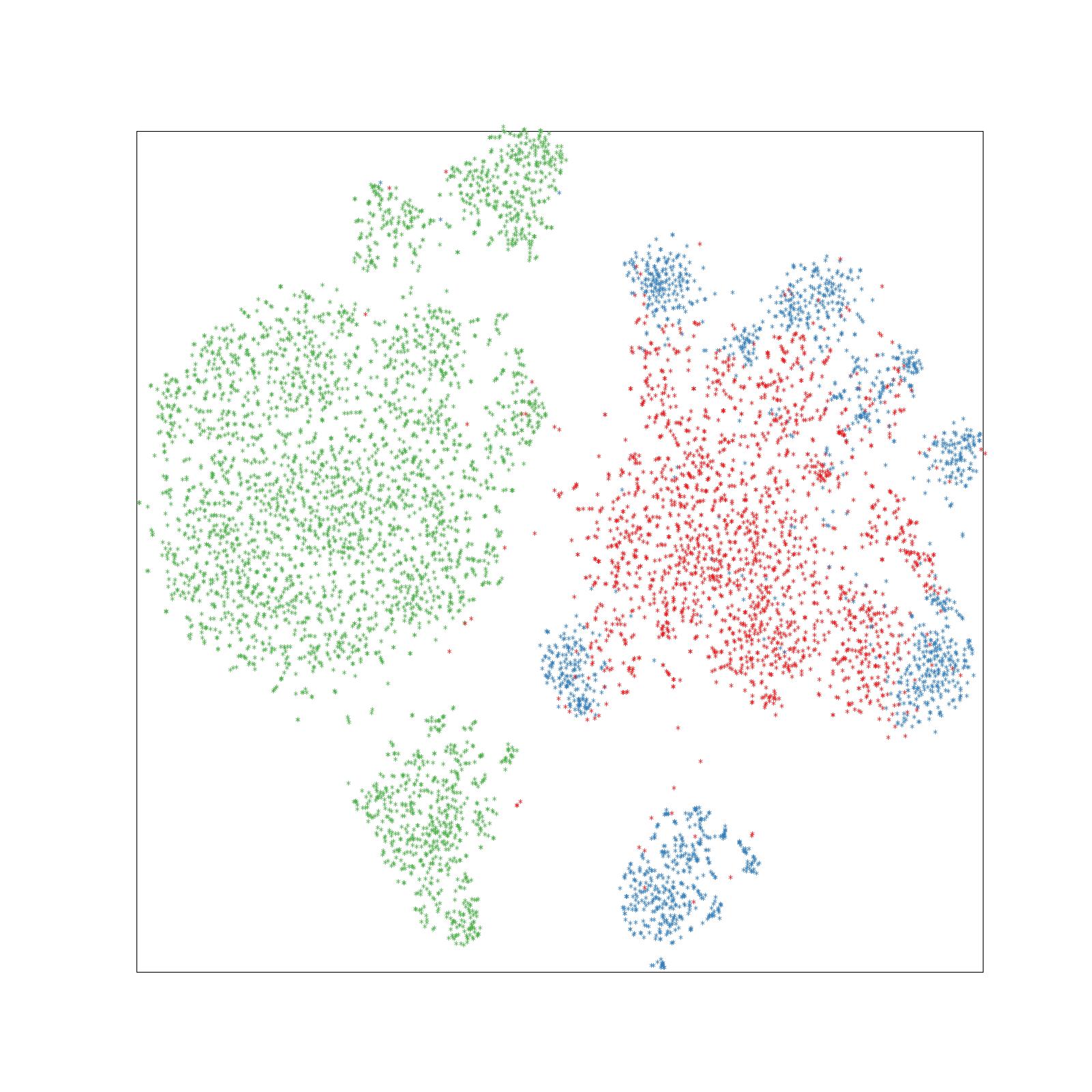}\vspace{1pt}
\includegraphics[width=0.8\linewidth,trim={0cm 0cm 0cm 0cm},clip]{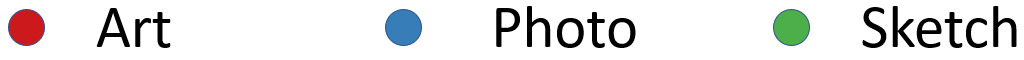}\vspace{1pt}
\includegraphics[width=0.8\linewidth,trim={5cm 2cm 2cm 4cm},clip]{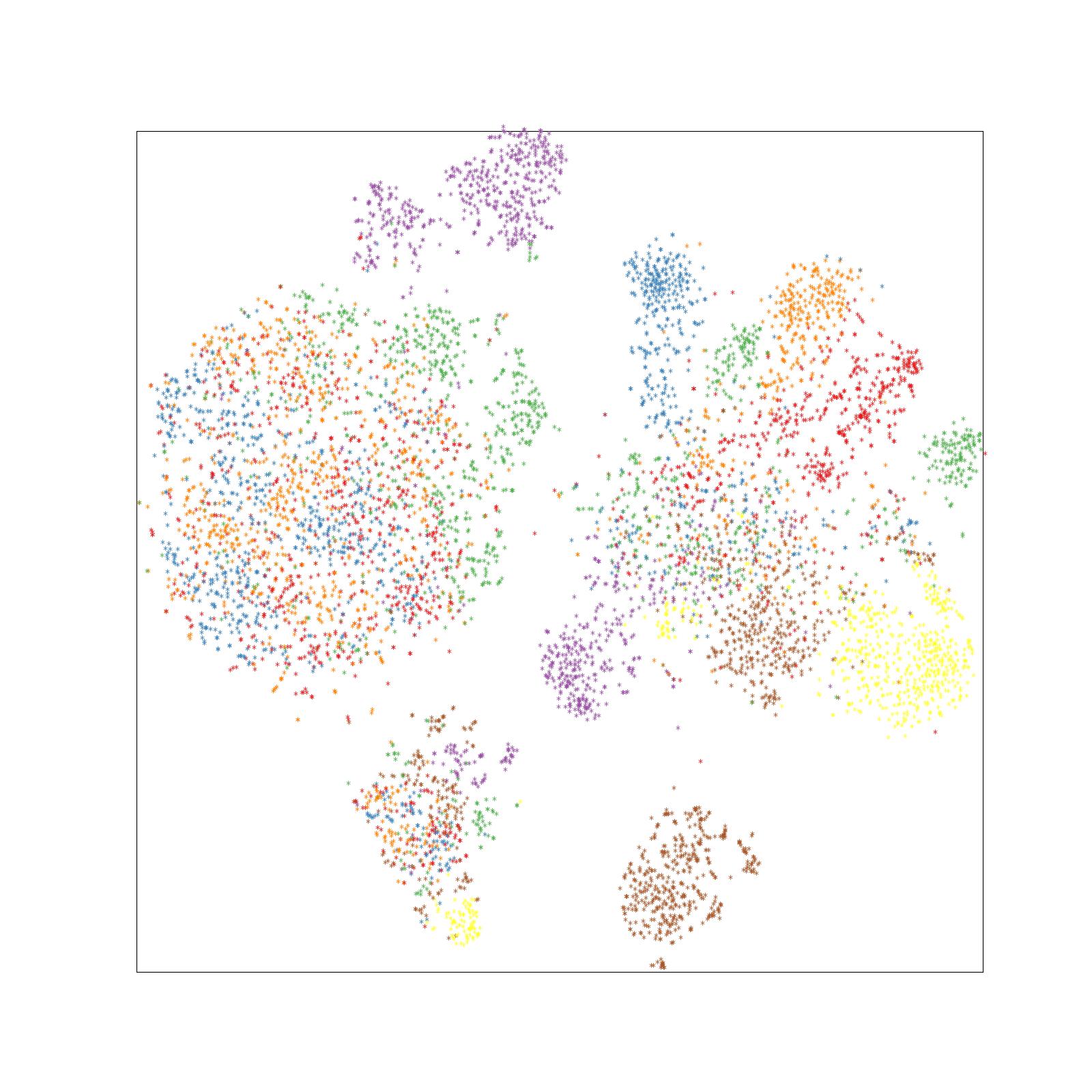}\vspace{1pt}
\includegraphics[width=0.8\linewidth,trim={0cm 0cm 0cm 0cm},clip]{images/classlabel.png}
\end{minipage}}
\subcaptionbox{}{
\begin{minipage}[b]{0.23\linewidth}
\includegraphics[width=0.8\linewidth,trim={5cm 2cm 2cm 4cm},clip]{images/ERM-1step/eval1_show_tsne.jpg}\vspace{1pt}
\includegraphics[width=0.8\linewidth,trim={0cm 0cm 0cm 0cm},clip]{images/label2.png}\vspace{1pt}
\includegraphics[width=0.8\linewidth,trim={5cm 2cm 2cm 4cm},clip]{images/ERM-1step-id/eval1_show_tsne.jpg}\vspace{1pt}
\includegraphics[width=0.8\linewidth,trim={0cm 0cm 0cm 0cm},clip]{images/classlabel.png}
\end{minipage}}
\subcaptionbox{}{
\begin{minipage}[b]{0.23\linewidth}
\includegraphics[width=0.8\linewidth,trim={5cm 2cm 2cm 4cm},clip]{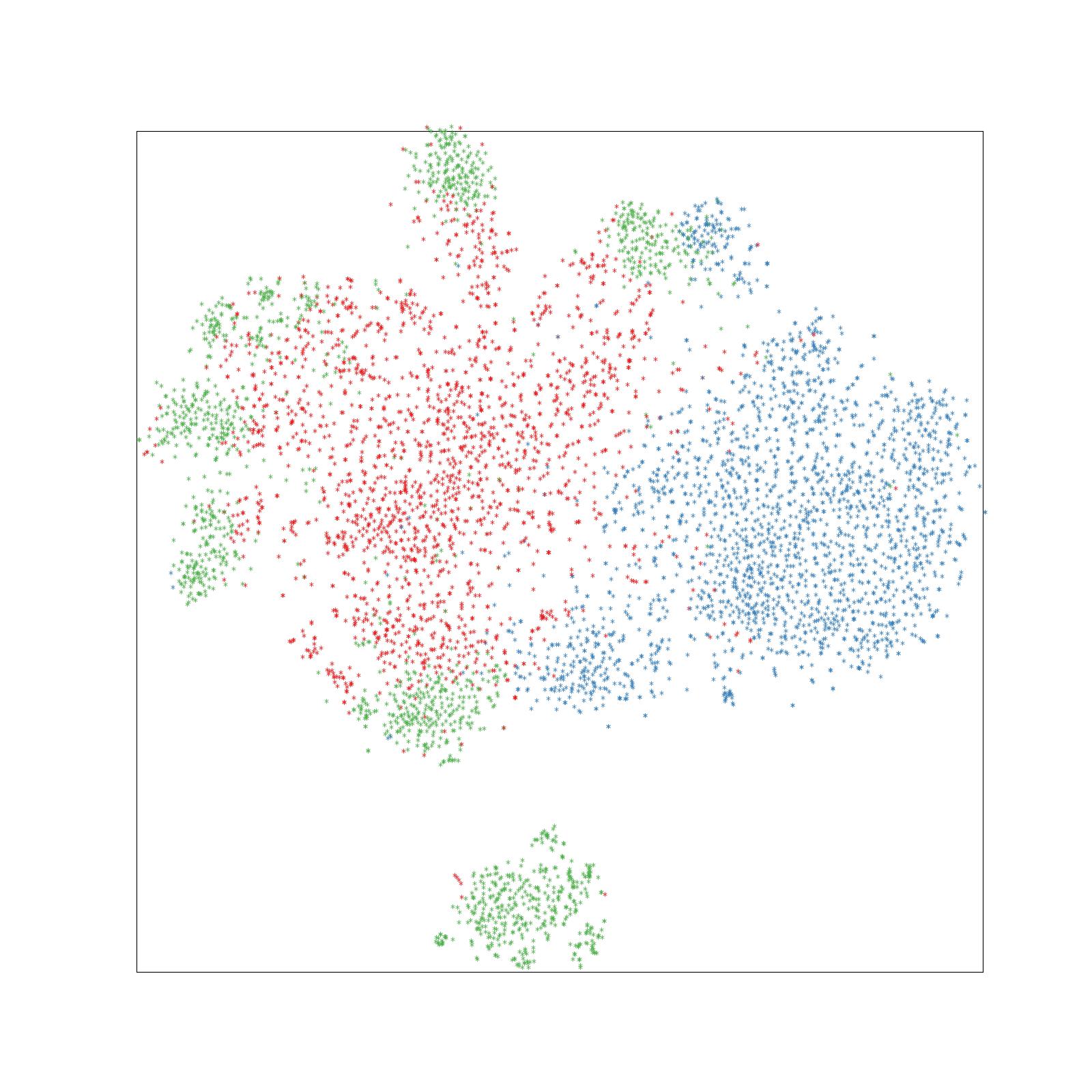}\vspace{1pt}
\includegraphics[width=0.8\linewidth,trim={0cm 0cm 0cm 0cm},clip]{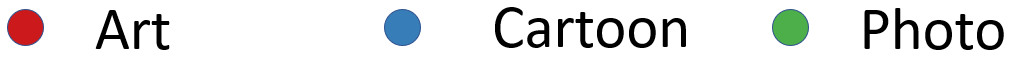}\vspace{1pt}
\includegraphics[width=0.8\linewidth,trim={5cm 2cm 2cm 4cm},clip]{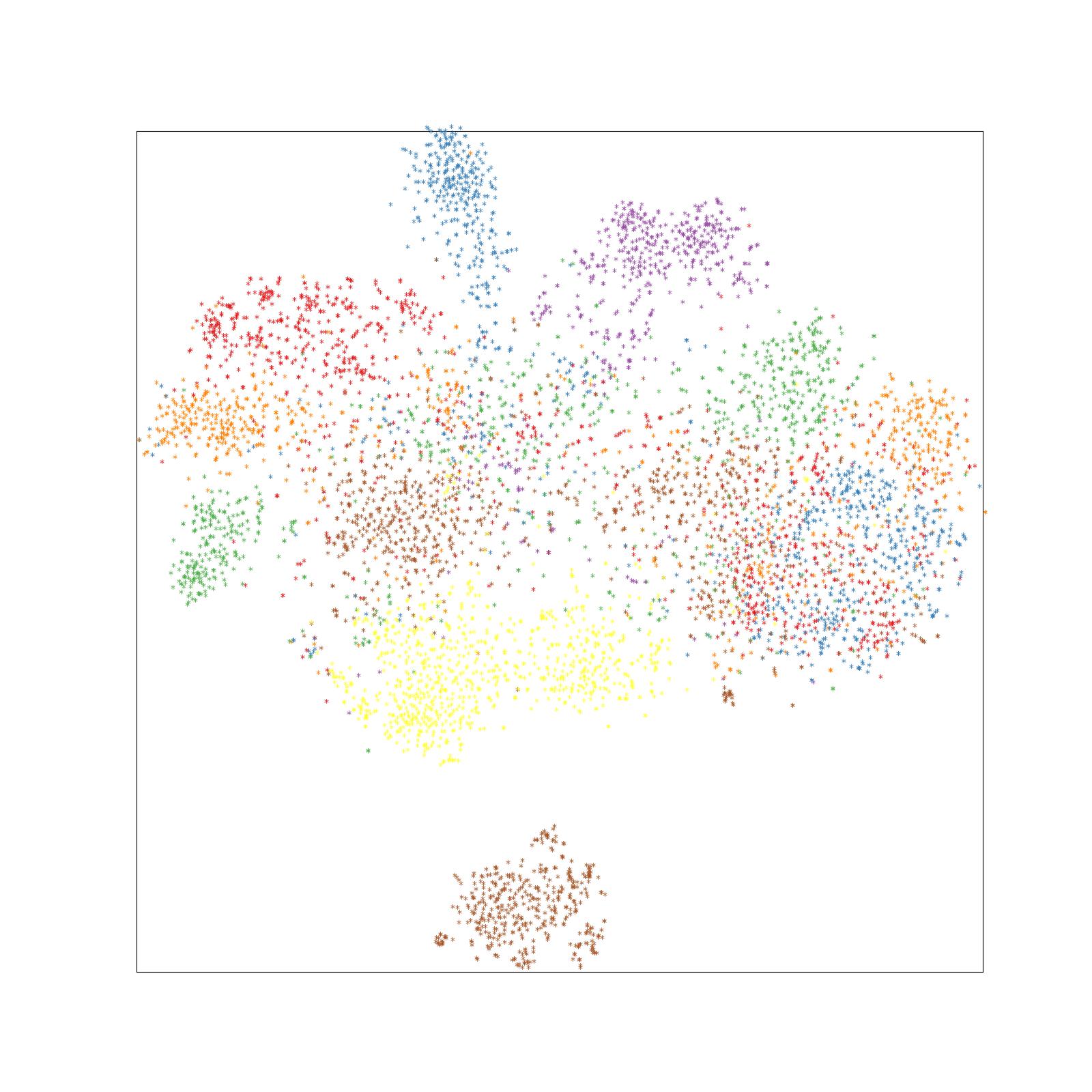}\vspace{1pt}
\includegraphics[width=0.8\linewidth,trim={0cm 0cm 0cm 0cm},clip]{images/classlabel.png}
\end{minipage}}
\caption{Visualization based on domain labels and class labels of feature clustering of trained model on PACS dataset. The training method is ERM with ImageNet~\cite{imagenet} pretrained resnet-50 and triplet loss on DomainBed~\cite{domainbed}. Features in the first row are coloured by domain labels and in the second row are coloured by class labels.  } 
\label{Fig2} 
\end{figure*}
\subsection{What affects feature clustering?}\label{explanation}
In this subsection, we give our explanation of feature clustering based on our assumptions. 
From contrastive learning, if we want to cluster the features based on the independent conditions $C=\left\{C_{k}\right\}$, the main purpose is trying to calculate the distance between $P(x_{1}|C)$ and $P(x_{2}|C)$, where $x_1, x_2$ are embedding features from different images. However, not all the prior conditions should be determinant to the feature clustering --- some are just noise such as the background information in face recognition. In this case, the prior conditions set $C$ can be divided into two parts: determinant conditions $B$ and non-determinant conditions $A$. Ideally, the non-determinant conditions $A$ should be unified, which means $P(A|x_{1})\equiv P(A|x_{2}), \forall x_{1}\ne x_{2}$. By Bayes' Theorem, we have the following equations:
\begin{equation}
\centering
\begin{split}
     & P(x|A,B_{1}) - P(x|A,B_{2})\\ 
    = & \frac{P(A|x)(P(x|B_{1})-P(x|B_{2}))}{P(A)}
\end{split}
\label{form1}
\end{equation}
When calculating the distribution distance in a metric space, the Wasserstein distance~\cite{wdistance} is commonly used as follows:
\begin{equation}
\centering
\begin{split}
   &W_{p}(P(x|A,B_{1}),P(x|A,B_{2})) \\= &\left(\int_{\mathbf{S}} |P(x|A,B_{1}) - P(x|A,B_{2})|^{p}dx)\right) ^{1/p}\\ 
    = & \left(\int_{\mathbf{S}}\frac{P(A|x)^{p}|P(x|B_{1})-P(x|B_{2})|^{p}}{P(A)^{p}}dx\right)^{1/p}
\end{split}
\label{form2}
\end{equation}
where $S$ is a subspace of the $L_{p}$ space. So, the relative distance will be determined by the distance between $P(x|B_{1})$ and $P(x|B_{2})$, which is also what the models are expected to achieve. However, $P(A|x)$ can not be totally the same in practice. Generally, the posterior $P(A|x)$ can be generalized by the quality and quantity of the data, which means $P(A|x_{1}) \approx P(A|x_{2}), \forall x_{1}\ne x_{2}$. But in some special situations, if the infomration $A$ cannot be generalized, for example if $A$ can be recognized as $A_{1}, A_{2}, A_{1}\ne A_{2}$ by the model, obviously it will affect feature clustering in an unexpected way. To find out the circumstance when $A$ affects feature clustering, we represent distance $d_{A}=|P(x|A_{1})-P(x|A_{2})|$ and distance $d_{B}=|P(x|B_{1})-P(x|B_{2})|$.  Then, with Equation \ref{form1}, when $d_{A} \gg max\big(\frac{P(x|A_{1})d_{B}}{P(x|B_{2})},\frac{P(x|A_{2})d_{B}}{P(x|B_{1})}\big)$, the non-determinant information $A$ will dominate the factors affecting the distance between $P(x|A,B_{1})$ and $P(x|A,B_{2})$, which is not what we expected. In the following, we show the more details.\par
In the case where $A$ can be divided into $A_{1}$ and $A_{2}$, which means the non-determinant condition cannot be generalized from the training data, we have 
\begin{align} \nonumber
    &P(x|A_{1},B_{1}) - P(x|A_{2},B_{2})\\  \nonumber
    =& \frac{P(A_{1}|x)P(x|B_{1})}{P(A_{1})} -\frac{P(A_{2}|x)P(x|B_{2})}{P(A_{2})} \\ \nonumber
    =&\frac{P(x|A_{1})P(x|B_{1})}{P(x)} -\frac{P(x|A_{2})P(x|B_{2})}{P(x)} \\ \nonumber
    =& \frac{(P(x|A_{1})-P(x|A_{1}))P(x|B_{1})}{P(x)}\\ \nonumber
     &-\frac{(P(x|B_{1})-P(x|B_{2}))P(x|A_{2})}{P(x)} \\ \nonumber
\end{align}
So, if
\[|P(x|A_{1})-P(x|A_{2})| \gg \frac{P(x|A_{2})|P(x|B_{1})-P(x|B_{2})|}{P(x|B_{1})},\]
the determinant conditions $A$ will dramatically affect the distance between the two distributions. Using the same argument, we can also say 
\[|P(x|A_{1})-P(x|A_{2})| \gg \frac{P(x|A_{1})|P(x|B_{1})-P(x|B_{2})|}{P(x|B_{2})}.\]
Overall, if we represent distance $d_{A}=|P(x|A_{1})-P(x|A_{2})|$ and distance $d_{B}=|P(x|B_{1})-P(x|B_{2})|$, we have the conclusion that $A$ will dominate the distance between $P(x|A,B_{1})$ and $P(x|A,B_{2})$ when \[d_{A} \gg max\big(\frac{P(x|A_{1})d_{B}}{P(x|B_{2})},\frac{P(x|A_{2})d_{B}}{P(x|B_{1})}\big).\]
\par
With the assumption above, the feature will be clustered based on domain information in domain generalization, if the model is more sensitive to domain distributions. In our exploration, the pretrained models are exactly such kind of sensitive models. We give a visualization of feature clustering to illustrate this statement in the following. 

\subsection{Visualization of domain discrepancy}
  To determine the initial embedding distribution in a pretrained model, we adopt a pattern of feature visualization to domain generalization from object re-identification tasks. In this subsection, this pattern is used to visualize the feature embedding from the 1-step-trained model by t-SNE~\cite{tsne}, which is shown as Figure \ref{Fig2}. In Figure \ref{Fig2}, the images in the first row are the results coloured by domain labels, and the images in the second row are the results coloured by class labels. The labels are shown at the bottom of each image. The correspondences of colours and labels are shown in Figure \ref{Fig2} as well. All the visualizations of feature embedding in this paper are based on the settings used in Figure \ref{Fig2}.
  
 Initialized with the pretrained model on ImageNet, the features are supposed to be clustered with the semantic information or be scattered in the embedding space randomly. However, from Figure \ref{Fig2}, we can tell there are apparent boundaries among different domains instead of different classes, which means the pretrained model tends to cluster the features based on the domain information to some degree. We represent domain information as A and semantic information as B. As explained with the theory above, the semantic information in the initial training stage is not amplified, which makes $D_{A}$ satisfy the condition that the domain discrepancy dominates the feature distribution in the embedding space. This visualization of embedding space in pretrained networks is a demonstration of domain discrepancy in feature clustering, which demonstrates the relation between feature clustering and domain discrepancy in domain generalization. 
 
This visualization of domain discrepancy strongly demonstrates that domain discrepancy is one of factors affecting feature clustering in the pretrained models. Based on the explanation and visualization above, if we want to generalize the domain information in the embedding space, the domain clusters should be dispersed during the training process.

\section{Algorithm}
\label{sec: Algorithm}

\subsection{Domain-Class pair mining for triplet loss}
From the perspective above, to obtain the expected feature clusters, the domain discrepancy in source domains should be considered. Motivated by this idea, we propose a new pair mining method called Domain-Class pair mining to reduce the unexpected domain information in the contrastive losses. In the original pair mining methods, we define the triplet as $\left\{\alpha,p,n\right\}$, where $\alpha$ is the anchor image feature, $p$ is the positive image feature with same class label as $\alpha$ and $n$ is the negative image feature with a different class label from $\alpha$. In our Domain-Class pair mining, we re-define the pairs with domain information to help scatter features from the same domain in the semantic clusters. In the re-defined pair mining, $p$ means the positive image feature with the same class label  but a different domain label with $\alpha$ and $n$ means the negative image feature with the same domain label but a different class label with $\alpha$. \par
With the pair mining method above, the updated triplet loss can be represented as follows:
\begin{equation}
\centering
Loss_{DCT} = [D(a_{i_{c},i_{d}},p_{i_{c},j_{d}})-D(a_{i_{c},i_{d}},n_{j_{c},i_{d}})+\tau]_{+}
\label{eq1}
\end{equation}
where $c$ and $d$ represent class label and domain label respectively and $i\neq j$. $D$ represents the Euclidean distance (Wasserstein distance with moment 2) between features. The parameter $\tau$ is the margin of triplet loss. \par
 This domain-aware pair mining will help not only cluster the features based on class labels, but will also scatter the features by domain labels to generalize the domain information. This modification enables triplet loss to disperse the domain information, which further helps generalize the model to different domains.  Otherwise, there is the possibility that the positive pairs help cluster features in the same domain and the negative pairs help push features from different domains, which is unexpected in domain generalization.  
 
\subsection{Normalized Embedding Space}
~\cite{batchnorm}  explained how the Internal Covariate Shift (ICS)~\cite{batchnorm} moves the feature distribution in the training process. If the Euclidean distance  between features $f,g$ is $\left\|f-g\right\|_{2}$, the feature distance with ICS is $\left\|f-g+\Delta f- \Delta g\right\|_{2}$. Ideally, with the convergence of the model, there should be $\left\|f-g+\Delta f- \Delta g\right\|_{2} \leq \left\|f-g\right\|_{2}$ in the positive pairs. But with the strong uncertainty in the dataset, there can be significant shifts affecting the convergence of the distance in positive pairs. Actually, since the distance can not converge even if the positive pairs are fixed in the same domain in our experiments, there may be factors other than the pre-defined domain distribution causing uncertainty in the dataset. In this case, it is hard to handle the shift of each feature without more specific information, but it is possible to deal with shift of the embedding space or constrain the shift in an acceptable range in the training process. Two methods are utilized to deal with the shift as follows. \par
\noindent \textbf{No-bias Batch Normalization} The no-bias batch normalization is represented in the following:
\begin{equation}
\centering
\begin{split}
  f_{i}^{\ast}& = \gamma\frac{f_{i}-\mu}{\sigma}\\
  \mu& = \frac{1}{N_{B}}\sum^{N_{B}}_{i} f_{i}\\
  \sigma^{2}& = \frac{1}{N_{B}}\sum^{N_{B}}_{i} (f_{i}-\mu) 
\end{split}
\label{form3}
\end{equation}
Where $f_{i}, f_{i}^{\ast}$ represents the features from the backbone and features after BNNeck layer respectively. The term $N_{B}$ is the batch size in a mini-batch. The parameter $\gamma$ is a learnable scale parameter. The shift of the embedding space can be approximated as $E(\Delta f^{\ast})= E(f^{\ast}+\Delta f^{\ast}) - E(f^{\ast})$.  In every training step, the expectation $E(f^{\ast})$ is zero from Equation \ref{form3}. In addition, with the ICS, the expection can be represented as $E(f^{\ast}+\Delta f^{\ast})$ which is also zero from Equation \ref{form3}. Then, we have $E(\Delta f^{\ast}) \equiv 0$, which means the shift of the whole embedding space will be removed. So,  the no-bias batch normalization is used before triplet loss to constrain the embedding space in our algorithm.

\noindent\textbf{Feature Normalization} It is well known that the normalized features are distributed on the surface of the hypersphere centered at the origin  with radius 1 in $\mathbf{R}^{n}$, where $n$ is the dimensionality of the features. Even though  BNNeck is able to solve the shift of the embedding space, it can not constrain the shift of each feature. However, with normalization, the norm of shift is $\left\|\Delta f\right\| \leq \left\|f+\Delta f\right\|+\left\|f\right\| \leq 2$. So, with the bounded shift norm, the effect from ICS can be reduced. 
\subsection{Overall}
As stated above, the features used in triplet loss should be regularized in domain generalization. So, the expected triplet features should be $\left\{\alpha^{\ast},p^{\ast},n^{\ast}\right\}$, where $f^{\ast}$ means the regularized feature from the backbone-extracted feature $f$. The choice of the normalization method will be discussed in the following section. In addition, our DCT loss should be combined with cross-entropy loss. So, the total loss is $Loss=Loss_{cross-entropy}+L_{DCT}$. 
\section{Experiments}
\label{sec: Experiment}

\subsection{Datasets and Implementation Details}
We evaluate our DCT triplet method on 3 benchmark datasets composed of images for domain generalization as follows:\\
\textbf{PACS} \cite{PACS}: PACS contains 4 domains, 7 classes and 9991 images. Learning rate is set to 5e-5.\\
\textbf{OfficeHome} \cite{officehome}: OfficeHome contains 4 domains, 65 classes and 15588 images. Learning rate is set to 1e-4.\\
\textbf{VLCS} \cite{VLCS}: VLCS contains 4 domains, 5 classes and 10729 images. Learning rate is set to 1e-4.\par
The domain generalization protocol is proposed by DomainBed~\cite{domainbed} and the code framework is from SWAD~\cite{swad} which is updated from DomainBed. The DCT loss is implemented based on the triplet loss in Re-ID baseline \cite{reidsb,transreid}. In our experiments, the BN layer is initialized in the same manner as Transreid\cite{transreid}. The dataloader and evaluation index are the same as domainbed. The optimizer is SGD. Even though we build the code based on SWAD, our reported results are not combined with the SWAD algorithm. All the experiments are conducted on NVIDIA V100 or GTX 1080 Ti GPU, Linux system, Pytorch 1.10.2~\cite{pytorch}, Torchvision 0.11.3, Python 3.9.7, CUDA 11.4. Same as  SWAD, every experiment includes the leave-one-out cross-validations for all domains in each dataset. All the quantitative experiments are averaged with three trial seeds. 
\subsection{Margin Analysis}
To investigate our performance of varying hyperparameter $\tau$, we show the accuracy trend based on different $\tau$ on the PACS and OfficeHome datasets. \par
\begin{figure}[ht]
\centering 
\includegraphics[width=0.45\textwidth]{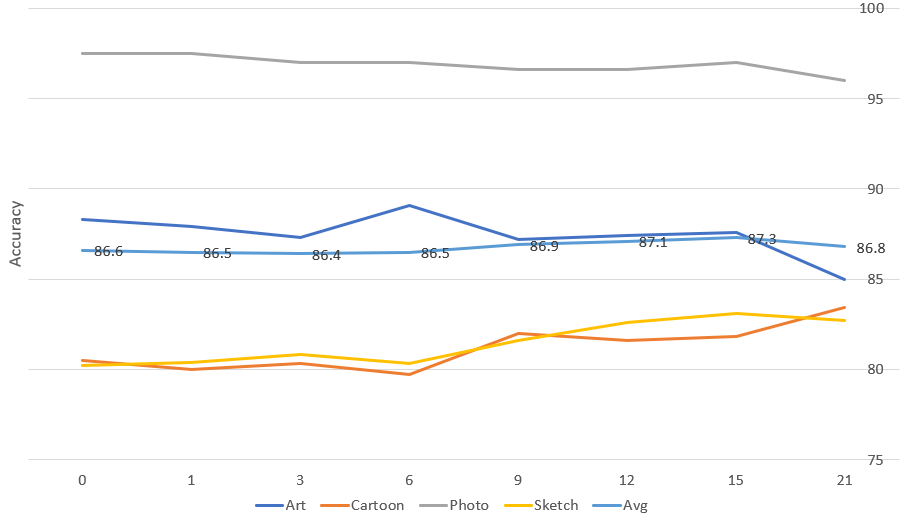}
\caption{Margin ablation on domain generalization dataset PACS. The x-axis and y-axis indicate margins and accuracy respectively.} 
\label{Fig3} 
\end{figure}
\begin{figure}[ht]
\centering 
\includegraphics[width=0.45\textwidth]{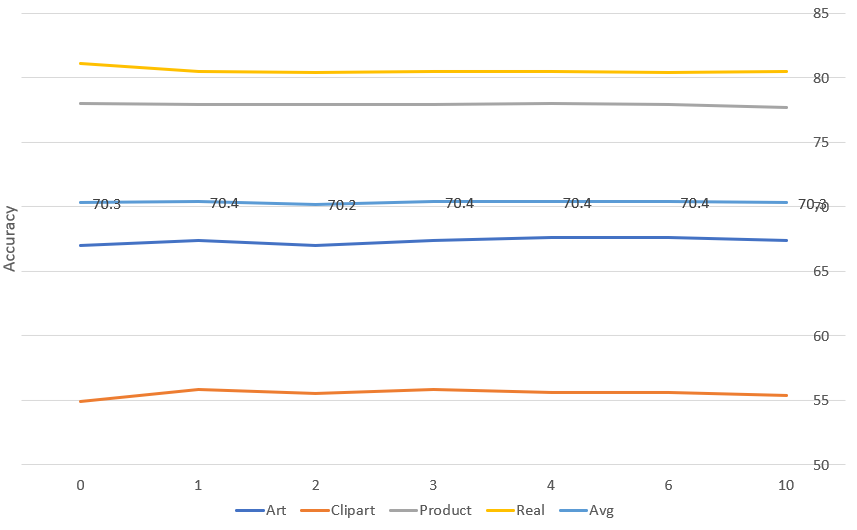} 
\caption{Margin ablation on domain generalization dataset OfficeHome. The x-axis and y-axis indicate margins and accuracy respectively.} 
\label{FigOHmargin} 
\end{figure}
Figure \ref{Fig3} shows the accuracy trend of different margins on the PACS dataset. The trends in different test domains are also shown with lines in different colours and the average results of different margins are given in Figure \ref{Fig3}. The DCT loss reaches its best performance when the margin is set to around 15 on the PACS dataset. In addition, with the increase of margin $\tau$, we can tell that the results evaluating on Cartoon and Sketch will be improved but the results on the other two domains will drop. This phenomenon can be explained by the motivation we mentioned above. From our assumption,  the Art and Photo domains, which are closer to the real image, will benefit from the pre-trained data domain. Indeed, the experiments on PACS in DomainBed~\cite{domainbed} show that the results on Art and Photo domains are always higher than the results on the other domains --- which can also supports our assumptions. As discussed, our algorithm is trying to disperse the domain information, which will negatively impact the results on the Art and Photo domains. So, with the increase of margin, the domain information will be disperseed more, explaining the phenomenon in Figure \ref{Fig3}.\par
Figure \ref{FigOHmargin} shows the accuracy of different margins on the OfficeHome dataset. The setting in this figure is the same as for igure \ref{Fig3}. For the OfficeHome dataset, we use both batch nomalization and feature normalization to regularize the embedding space.  We see that the margin does not affect DCT loss much on the OfficeHome dataset. So we do not choose margin intentionally on OfficeHome dataset for the following experiments. Theoretically, feature normalization will bound the distance of two features in the range of $[0,2]$, so the results will not be affected by the margin so much.  
\begin{figure*}
\centering
\caption{Domain-labeled visualization on ablation study of different parts in our algorithm on PACS. The first row  is the result with only cross-entropy loss. The second row is the result with cross-entropy loss and batch normalization. The third and fourth rows are the results with methods added triplet loss and DCT loss respectively based on the method of the second row. Both the margins in this experiment are 15. } 
\label{FigPACScluser}
\subcaptionbox{Cartoon,Photo, Sketch}{
\begin{minipage}[b]{0.23\linewidth}
\includegraphics[width=0.8\linewidth,trim={0cm 0cm 0cm 0cm},clip]{images/label1.png}\vspace{4pt}
\includegraphics[width=1\linewidth,trim={5cm 2cm 2cm 4cm},clip]{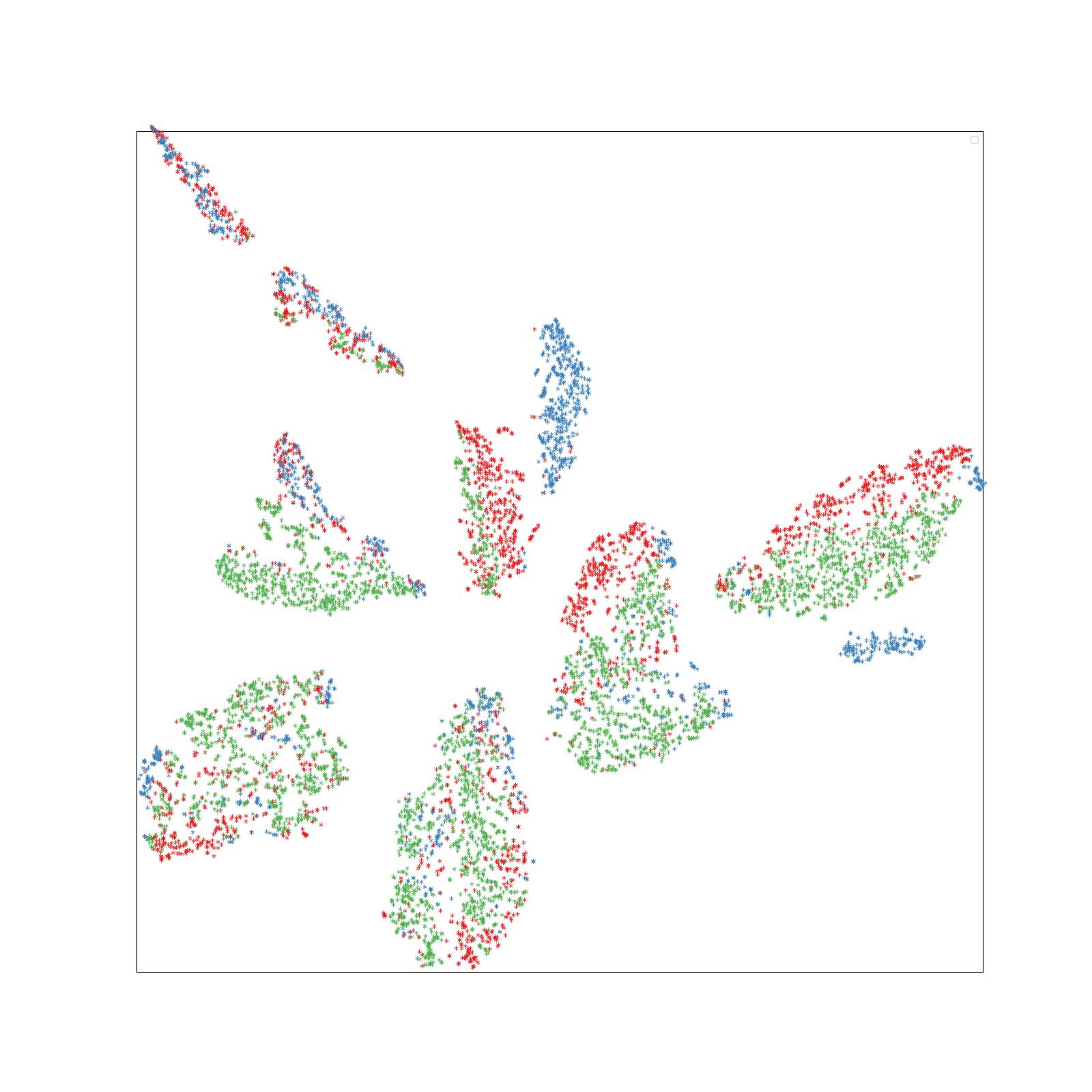}\vspace{4pt}
\includegraphics[width=1\linewidth,trim={5cm 2cm 2cm 4cm},clip]{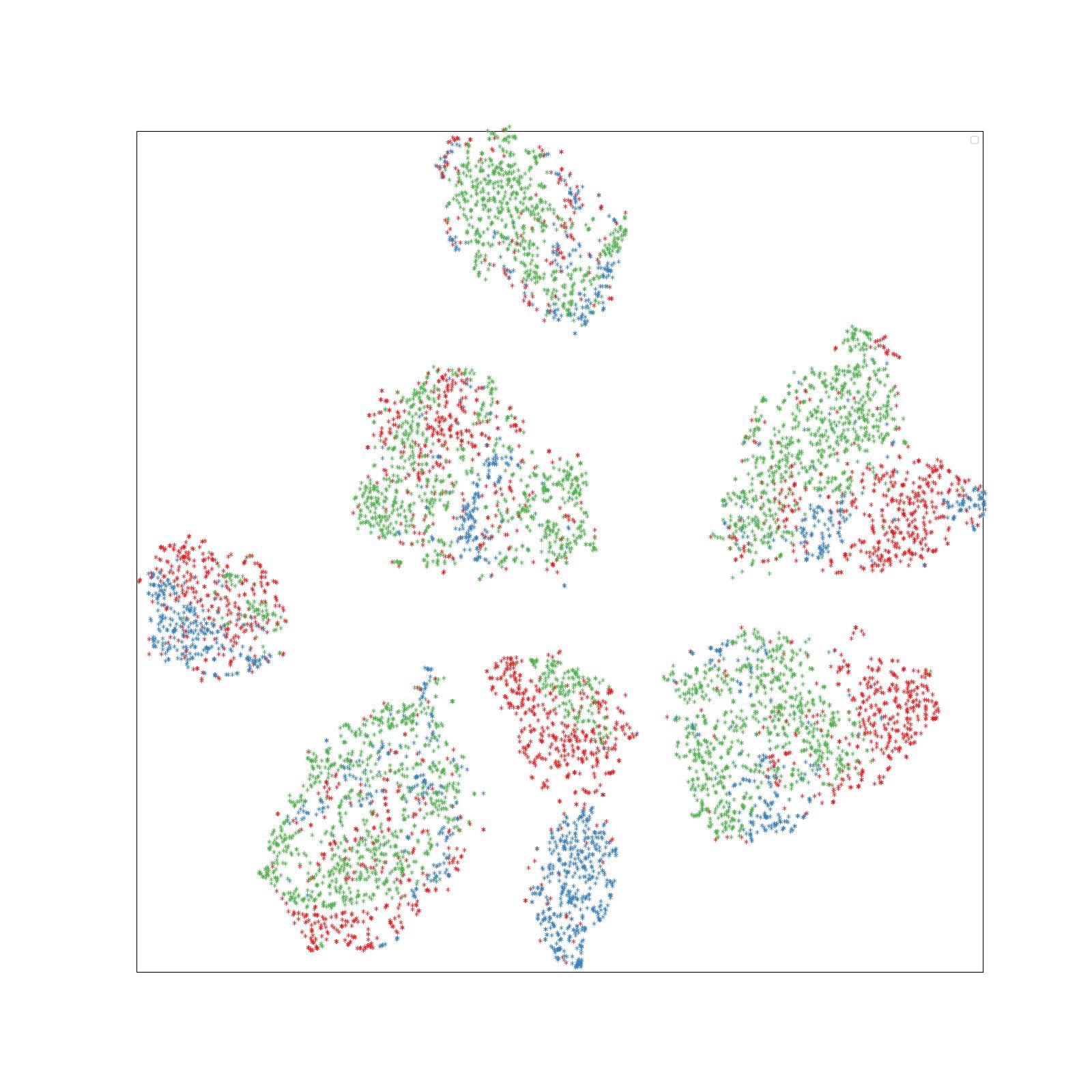}\vspace{4pt}
\includegraphics[width=1\linewidth,trim={5cm 2cm 2cm 4cm},clip]{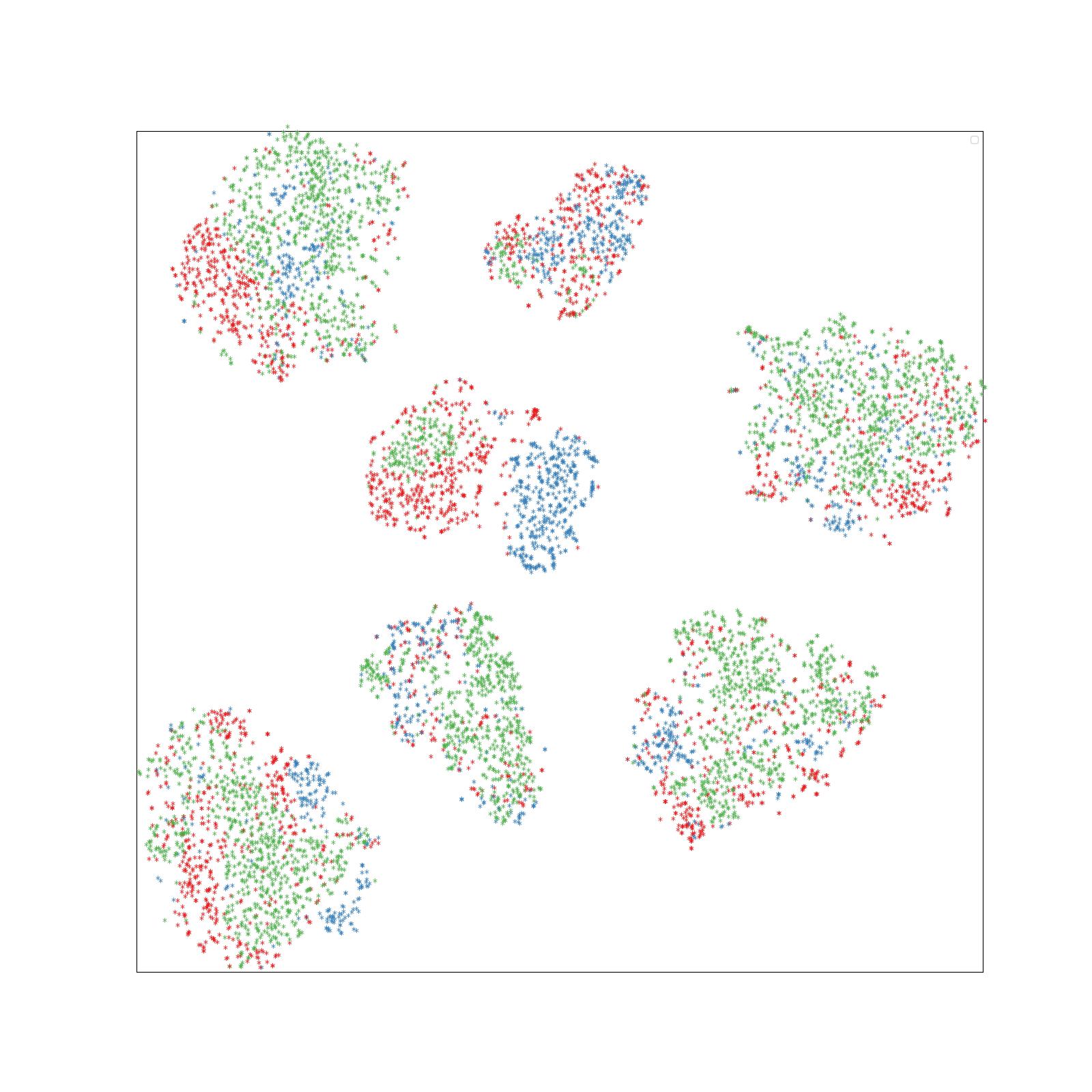}\vspace{4pt}
\includegraphics[width=1\linewidth,trim={5cm 2cm 2cm 4cm},clip]{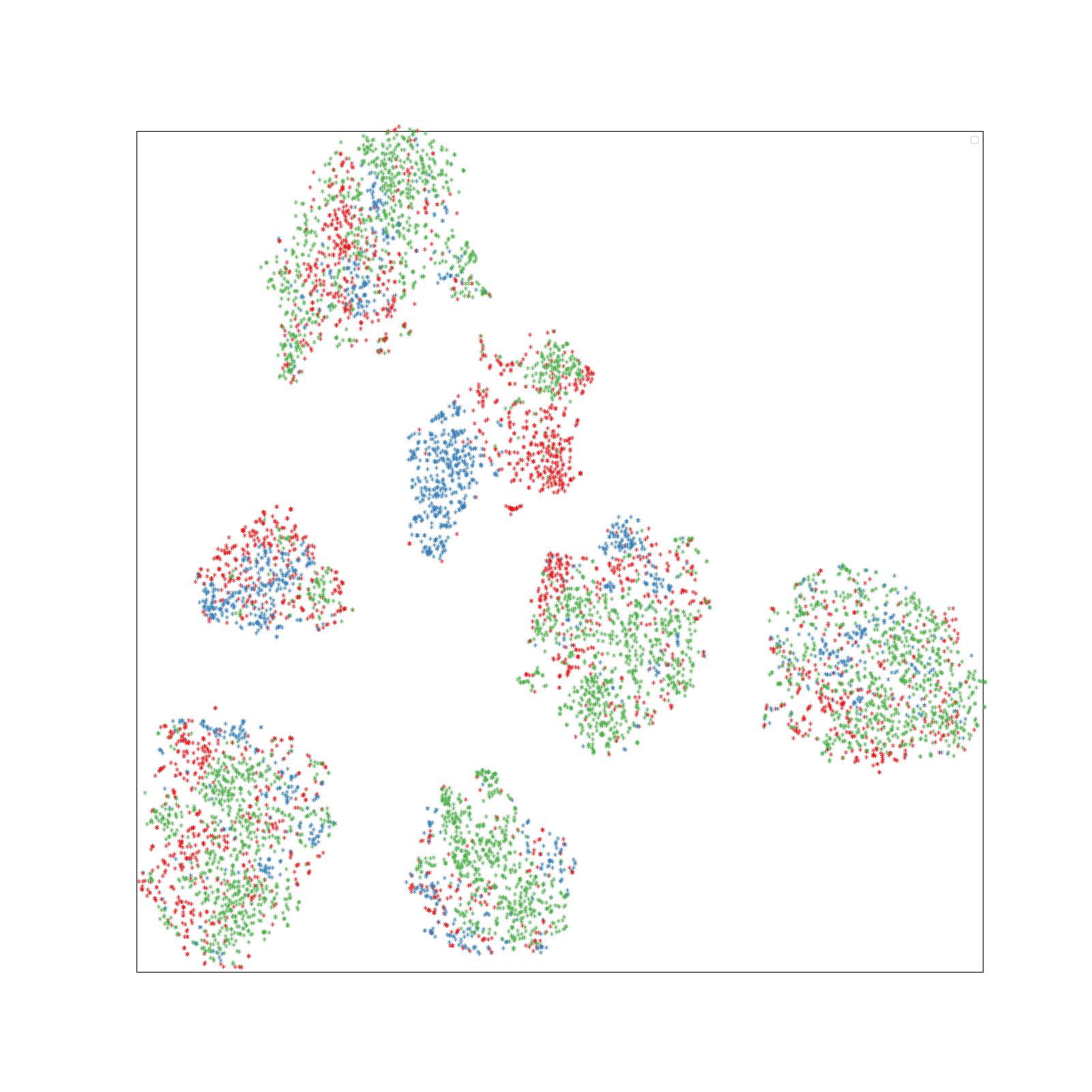}
\end{minipage}}
\subcaptionbox{Art, Photo, Sketch}{
\begin{minipage}[b]{0.23\linewidth}
\includegraphics[width=0.8\linewidth,trim={0cm 0cm 0cm 0cm},clip]{images/label2.png}\vspace{4pt}
\includegraphics[width=1\linewidth,trim={5cm 2cm 2cm 4cm},clip]{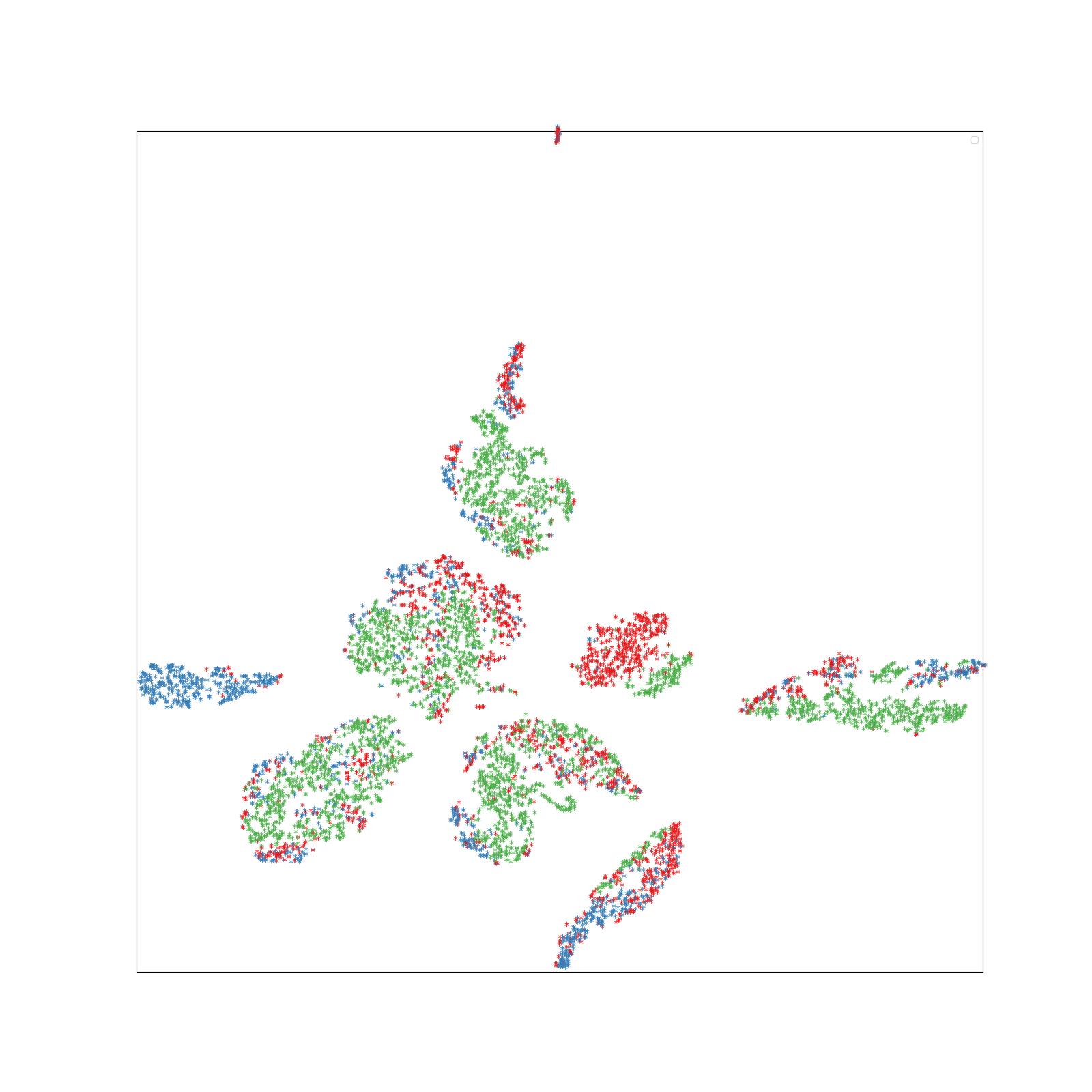}\vspace{4pt}
\includegraphics[width=1\linewidth,trim={5cm 2cm 2cm 4cm},clip]{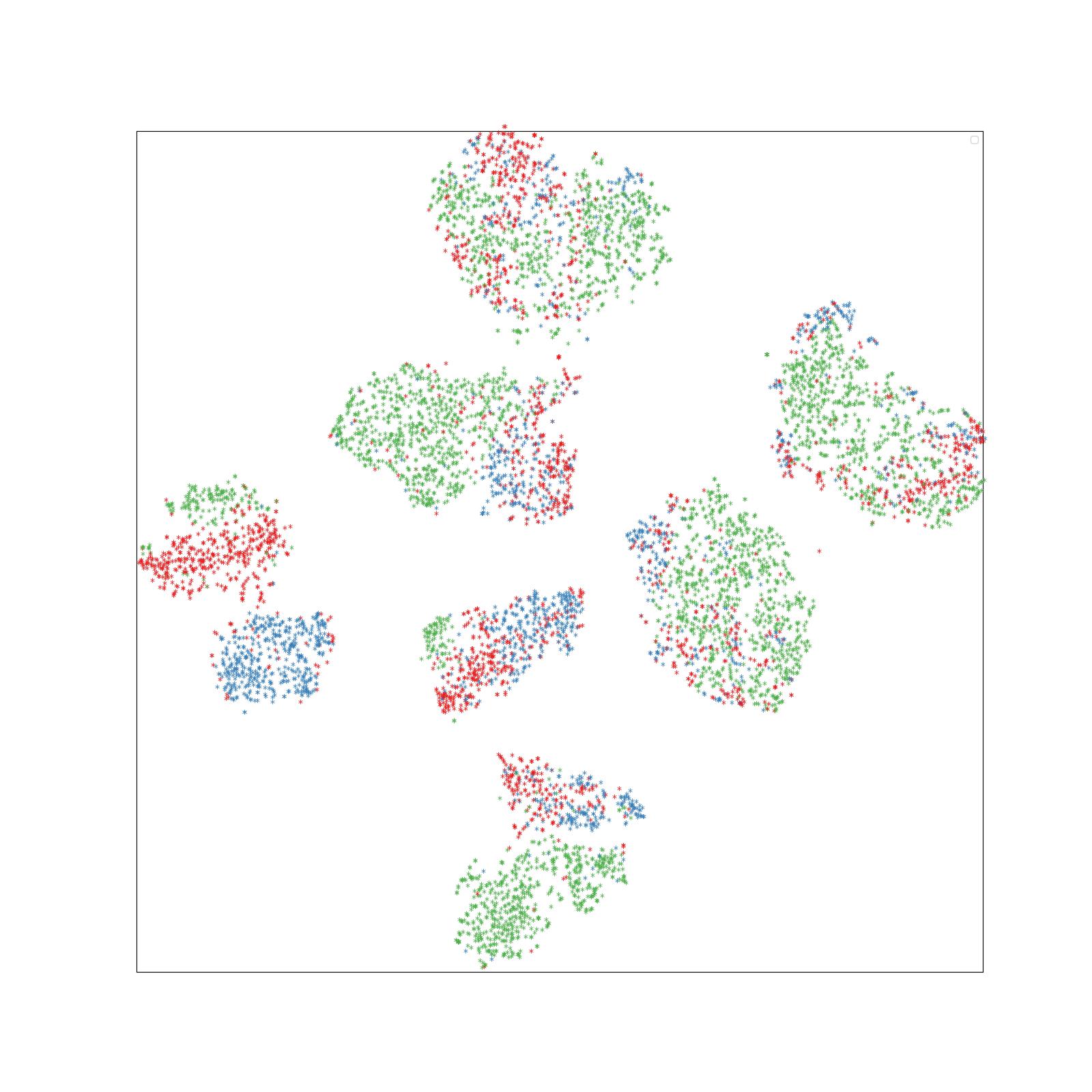}\vspace{4pt}
\includegraphics[width=1\linewidth,trim={5cm 2cm 2cm 4cm},clip]{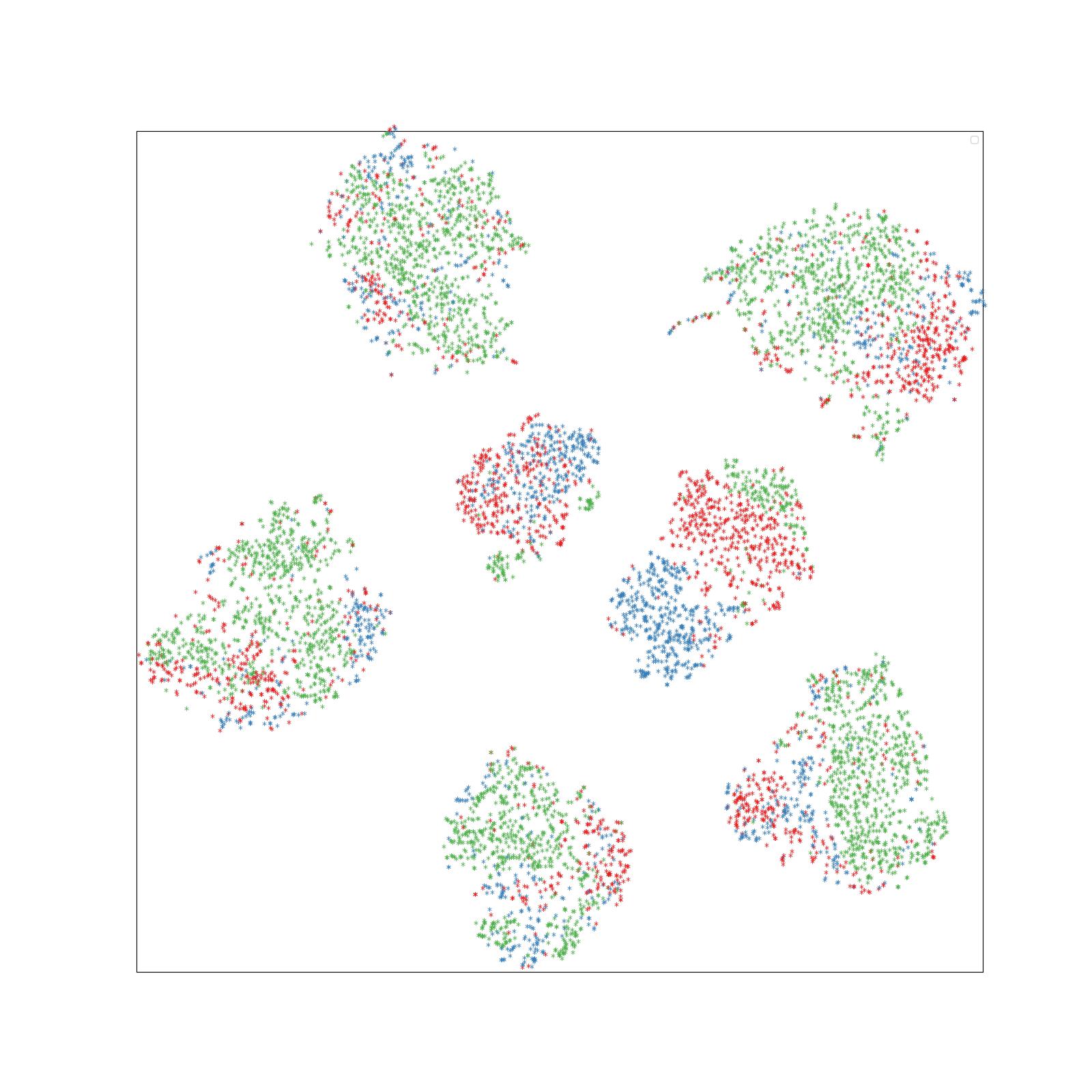}\vspace{4pt}
\includegraphics[width=1\linewidth,trim={5cm 2cm 2cm 4cm},clip]{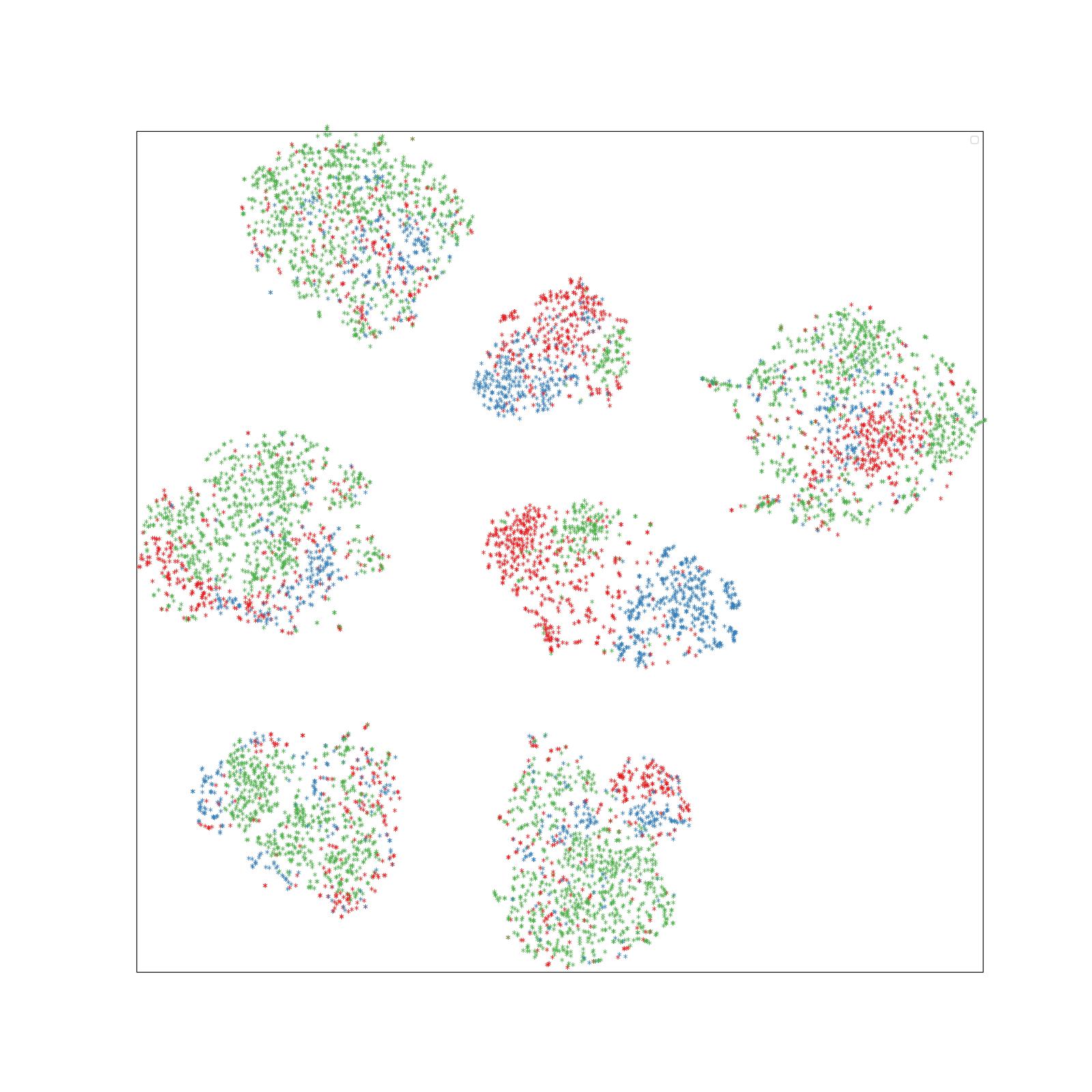}
\end{minipage}}
\subcaptionbox{Art, Cartoon, Sketch}{
\begin{minipage}[b]{0.23\linewidth}
\includegraphics[width=0.8\linewidth,trim={0cm 0cm 0cm 0cm},clip]{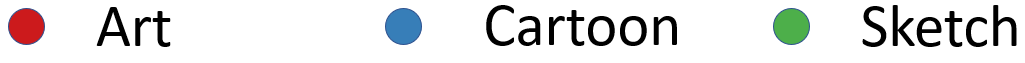}\vspace{4pt}
\includegraphics[width=1\linewidth,trim={5cm 2cm 2cm 4cm},clip]{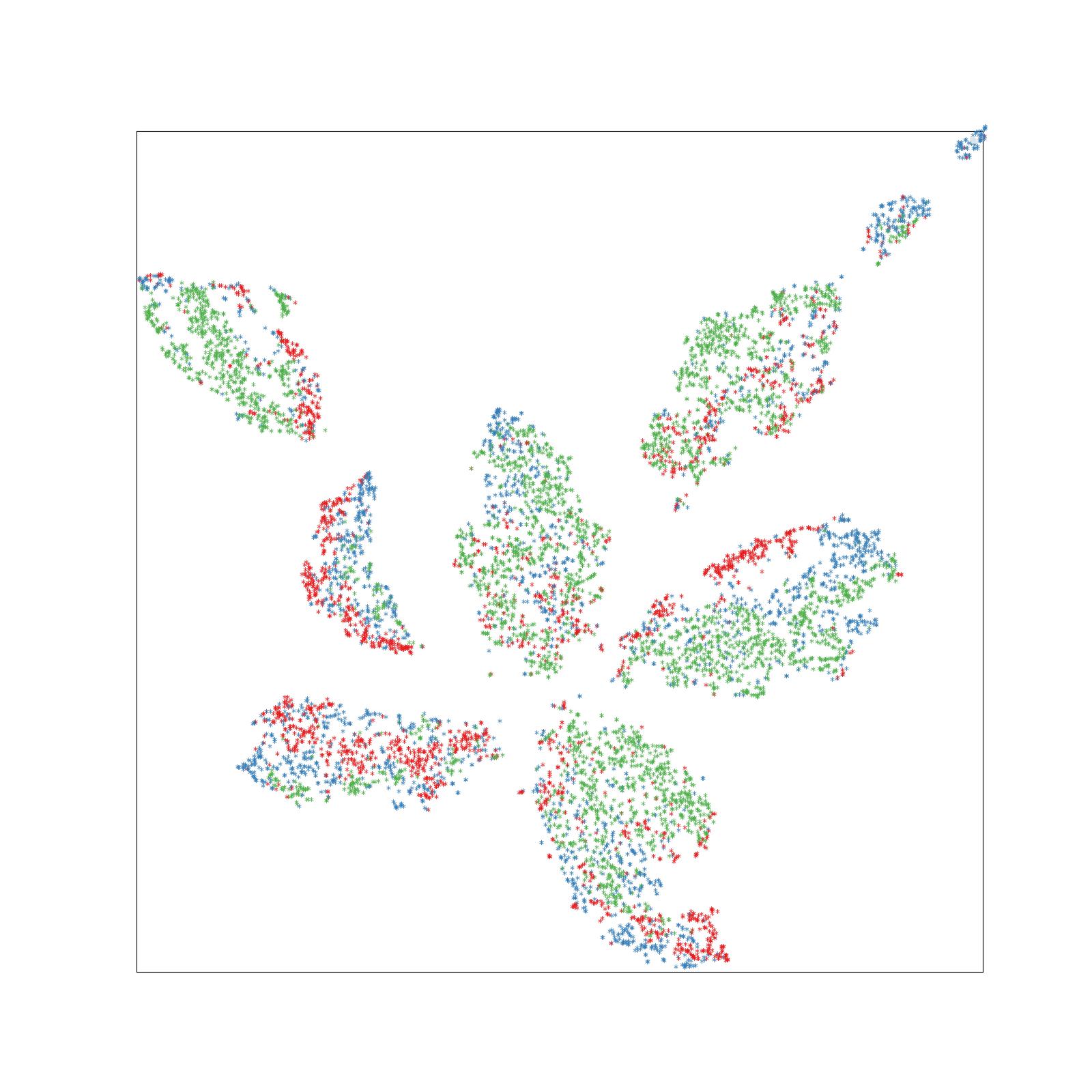}\vspace{4pt}
\includegraphics[width=1\linewidth,trim={5cm 2cm 2cm 4cm},clip]{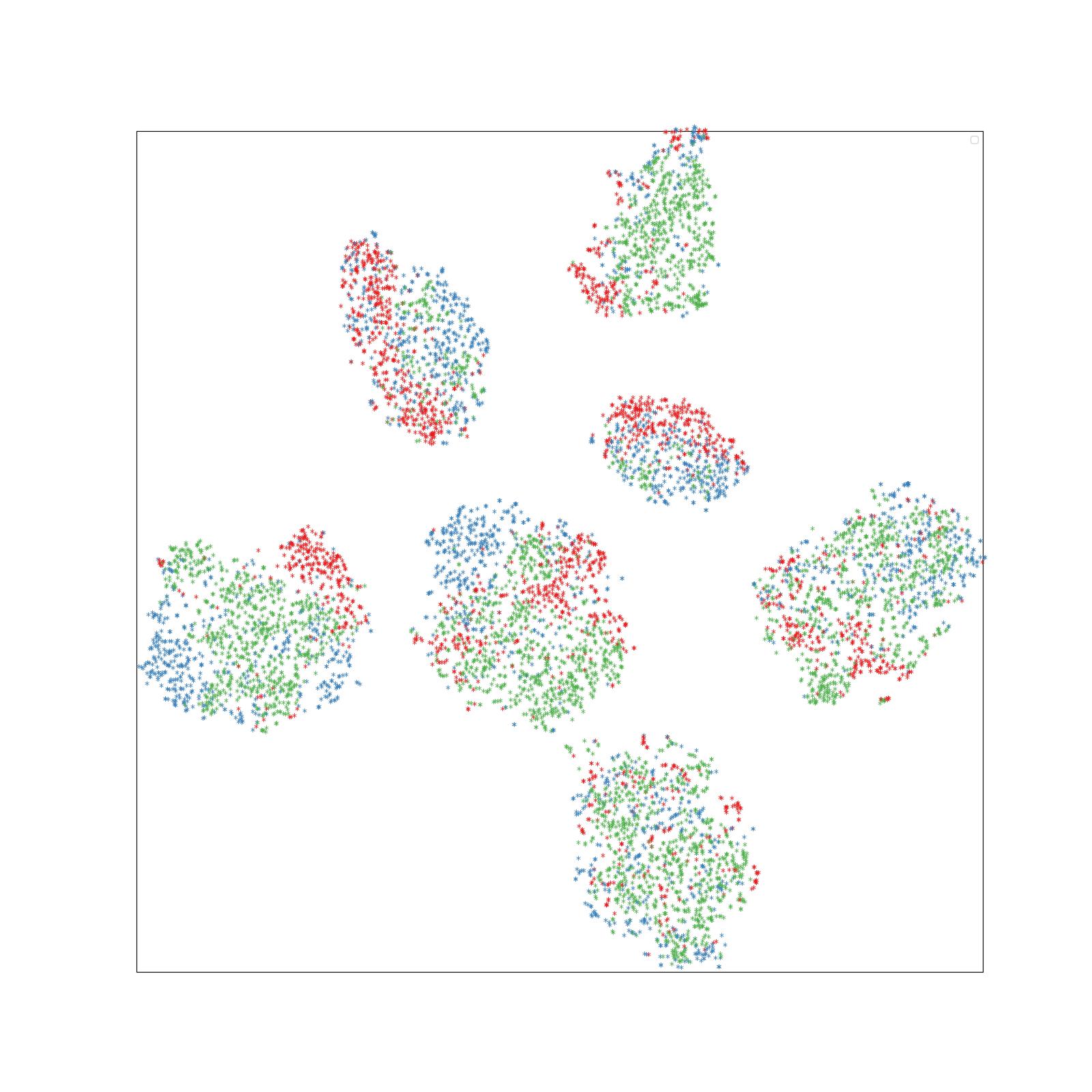}\vspace{4pt}
\includegraphics[width=1\linewidth,trim={5cm 2cm 2cm 4cm},clip]{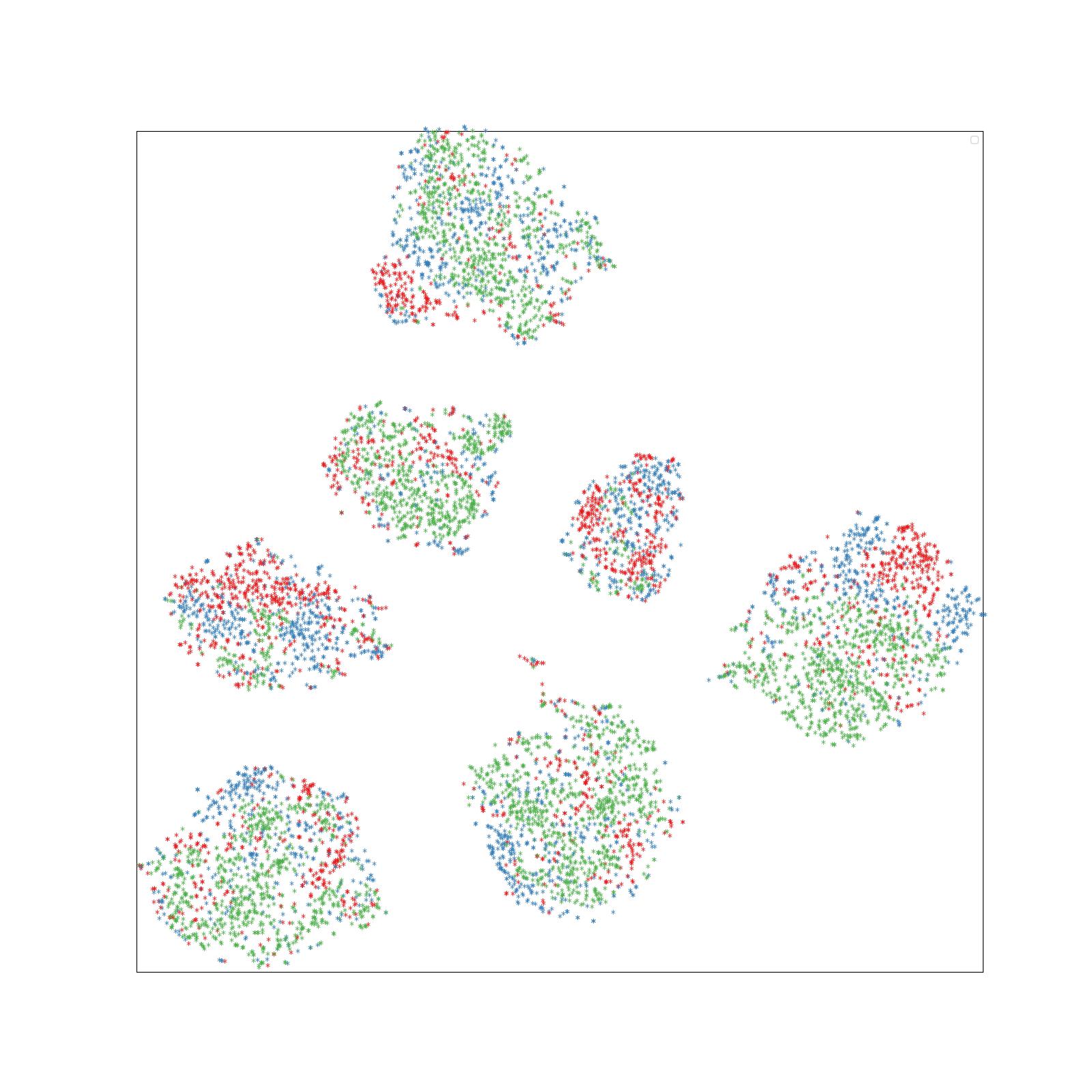}\vspace{4pt}
\includegraphics[width=1\linewidth,trim={5cm 2cm 2cm 4cm},clip]{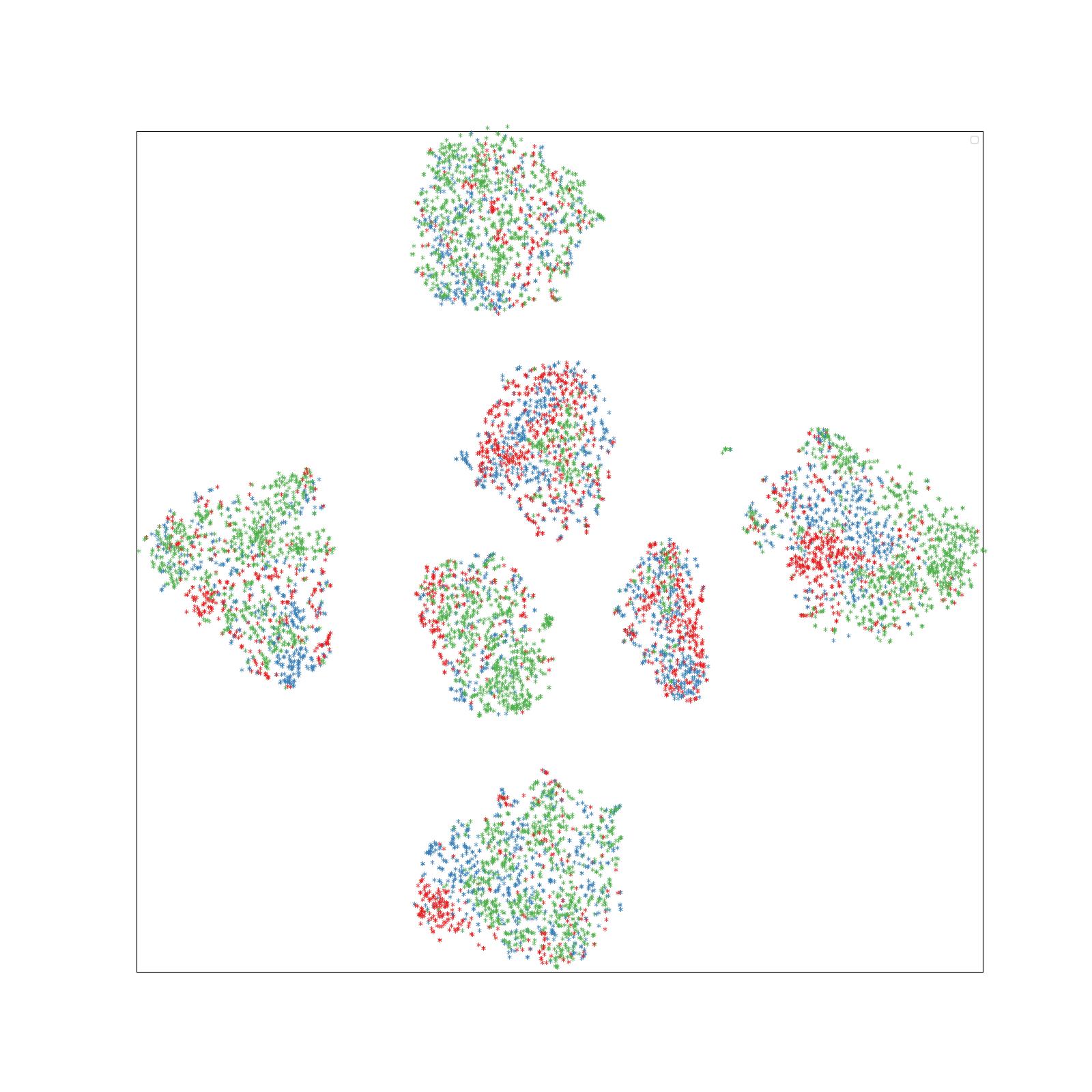}
\end{minipage}}
\subcaptionbox{Art, Cartoon, Photo}{
\begin{minipage}[b]{0.23\linewidth}
\includegraphics[width=0.8\linewidth,trim={0cm 0cm 0cm 0cm},clip]{images/label4.png}\vspace{4pt}
\includegraphics[width=1\linewidth,trim={5cm 2cm 2cm 4cm},clip]{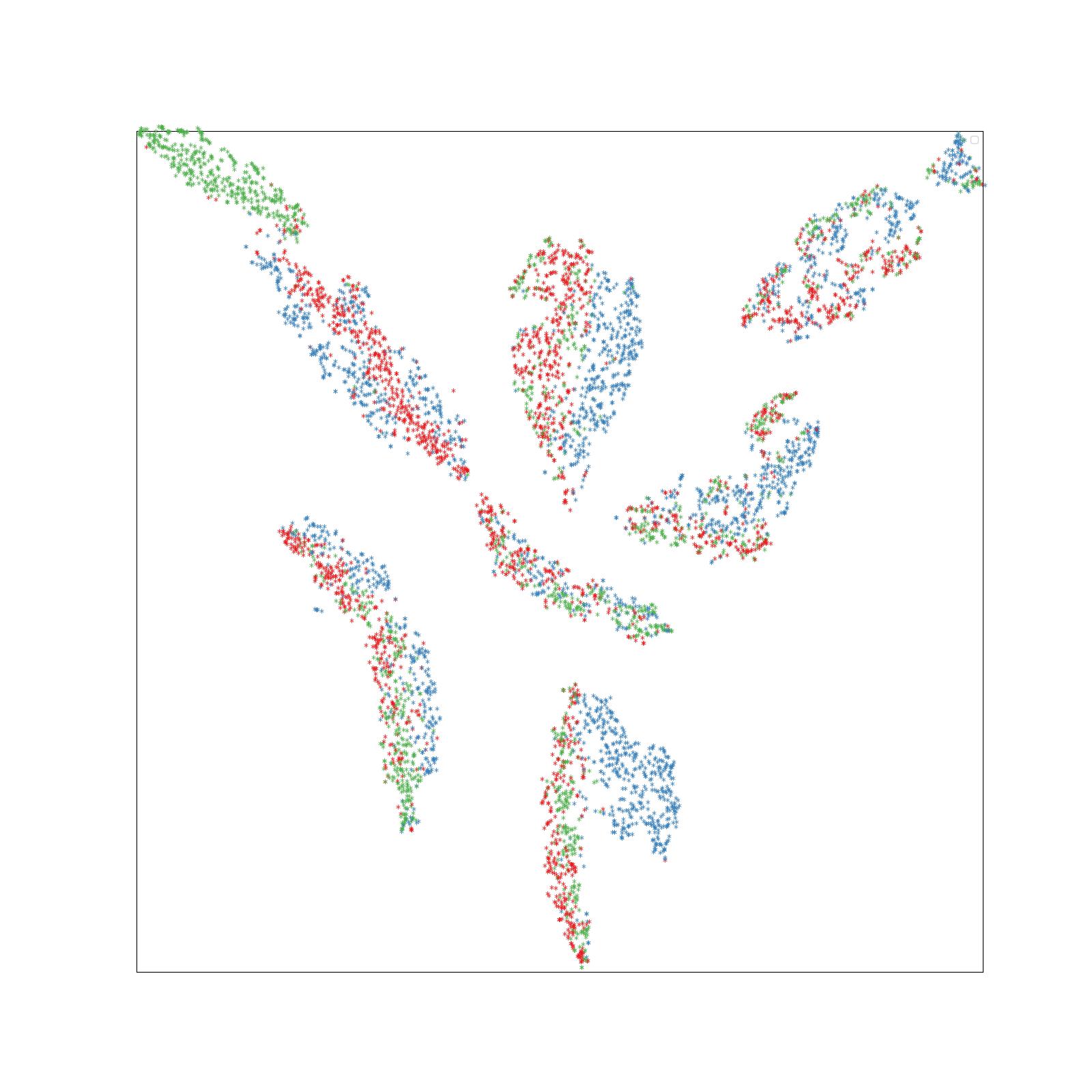}\vspace{4pt}
\includegraphics[width=1\linewidth,trim={5cm 2cm 2cm 4cm},clip]{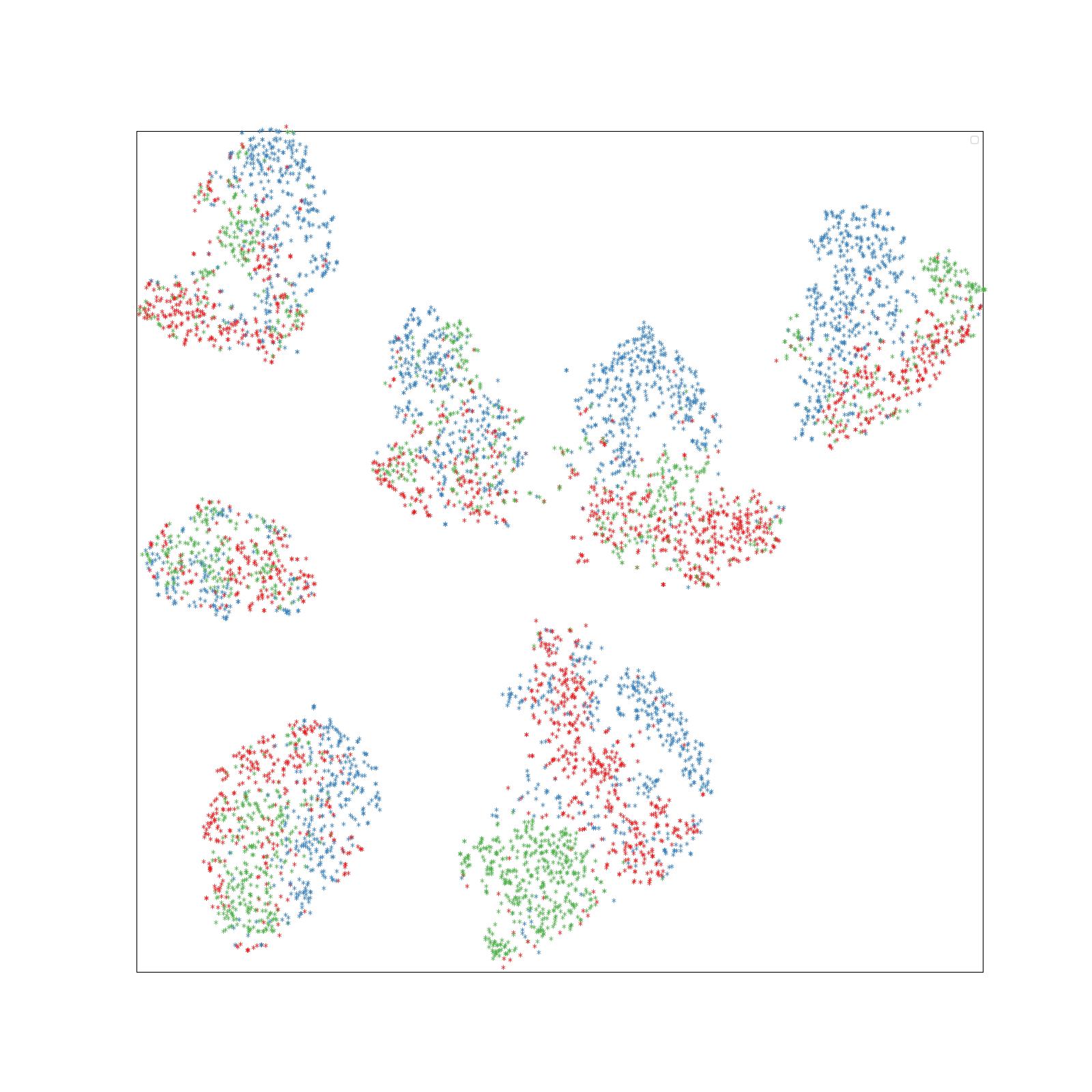}\vspace{4pt}
\includegraphics[width=1\linewidth,trim={5cm 2cm 2cm 4cm},clip]{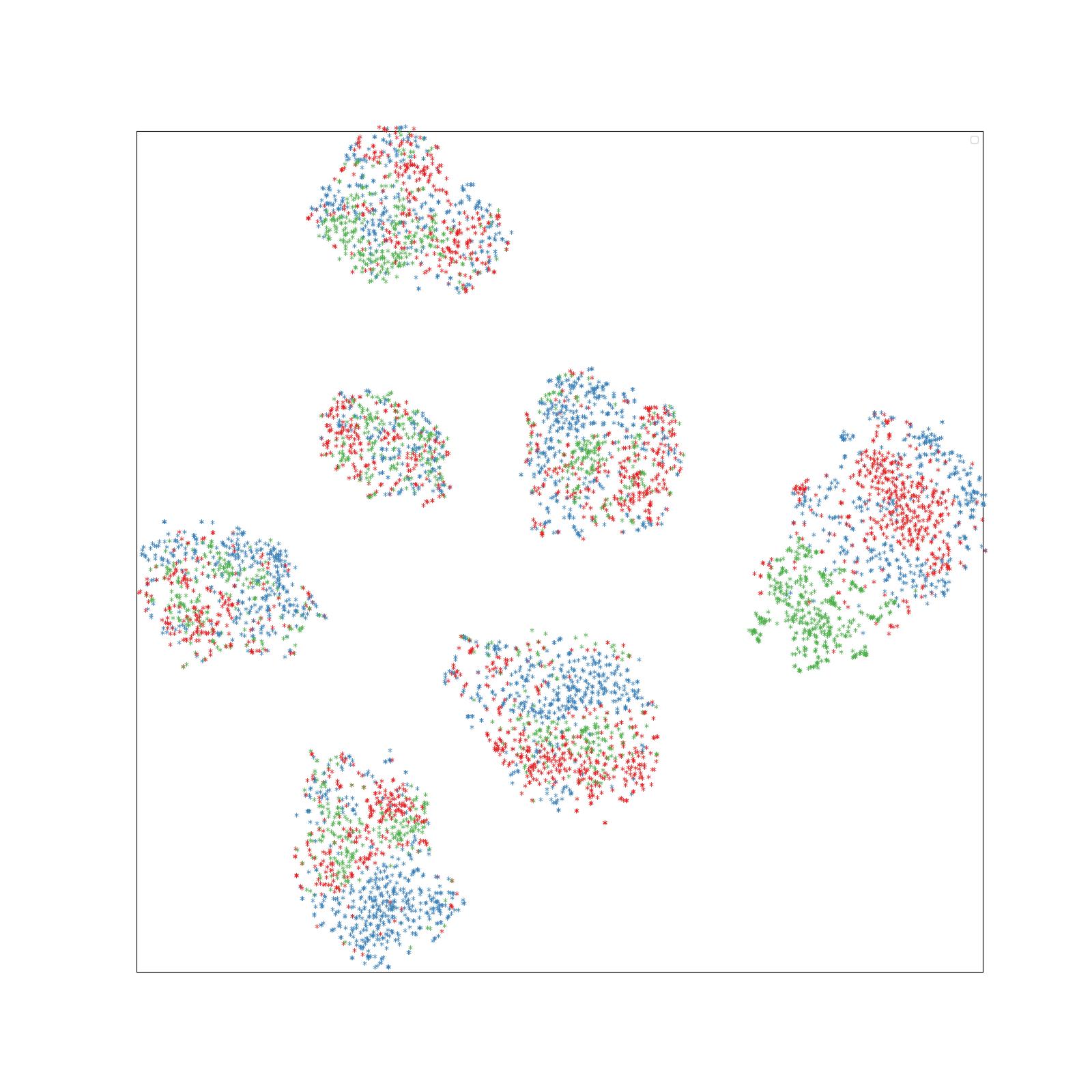}\vspace{4pt}
\includegraphics[width=1\linewidth,trim={5cm 2cm 2cm 4cm},clip]{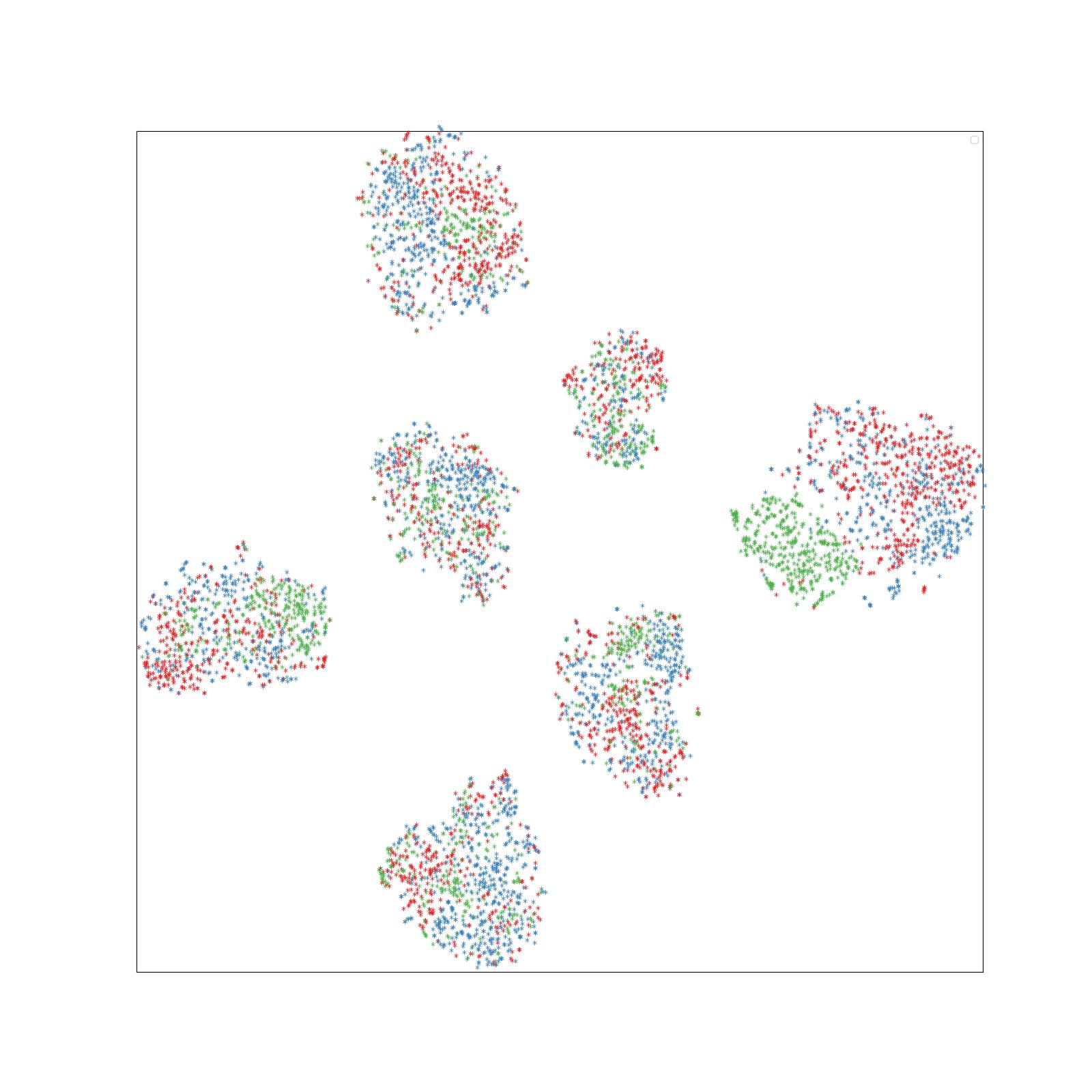}
\end{minipage}}
\end{figure*}
\begin{figure*}
\centering 
\caption{Class-labeled visualization corresponding to Figure \ref{FigPACScluser}} 
\label{FigPACScluserid} 
\subcaptionbox{Cartoon,Photo, Sketch}{
\begin{minipage}[b]{0.23\linewidth}
\includegraphics[width=0.8\linewidth,trim={0cm 0cm 0cm 0cm},clip]{images/classlabel.png}\vspace{4pt}
\includegraphics[width=1\linewidth,trim={5cm 2cm 2cm 4cm},clip]{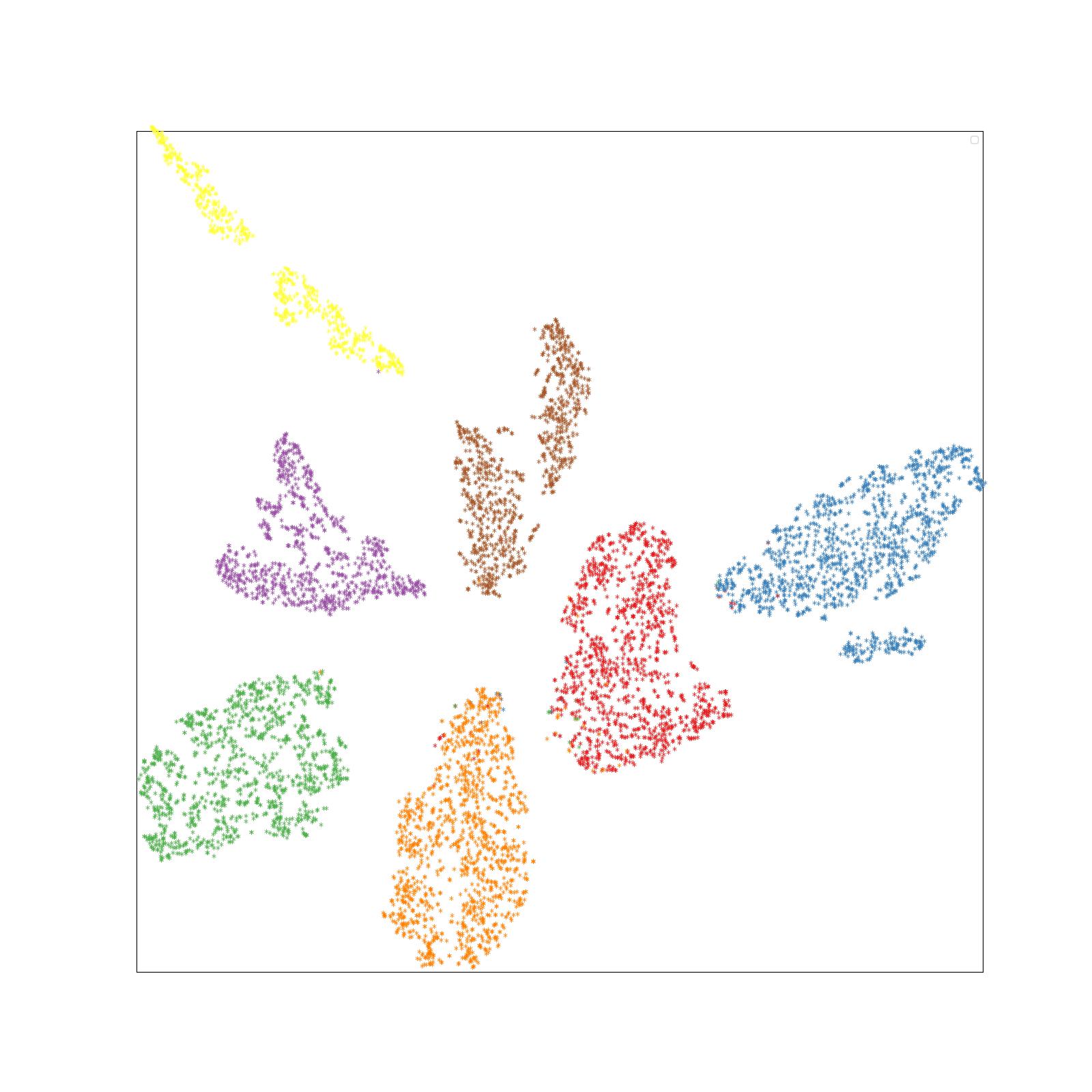}\vspace{4pt}
\includegraphics[width=1\linewidth,trim={5cm 2cm 2cm 4cm},clip]{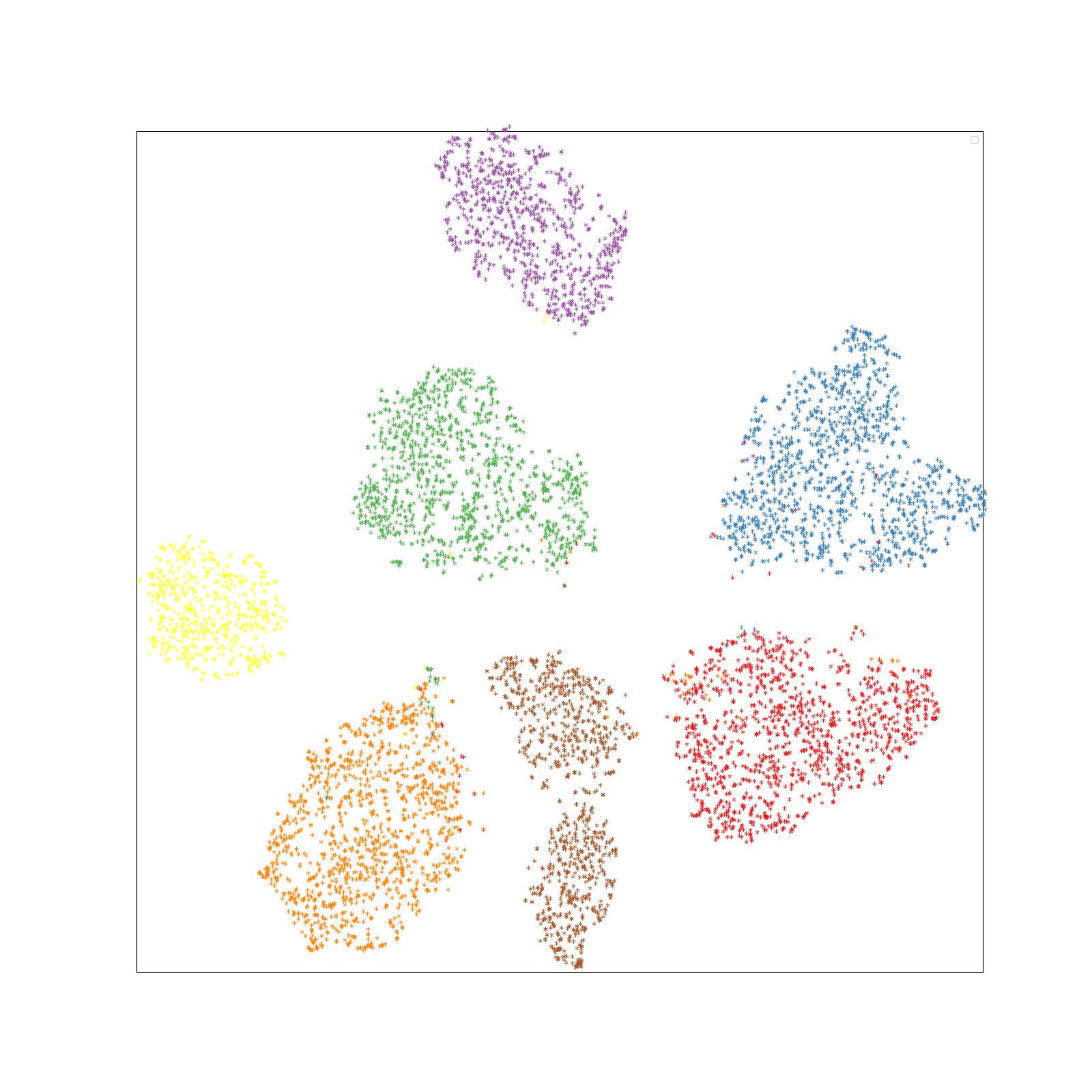}\vspace{4pt}
\includegraphics[width=1\linewidth,trim={5cm 2cm 2cm 4cm},clip]{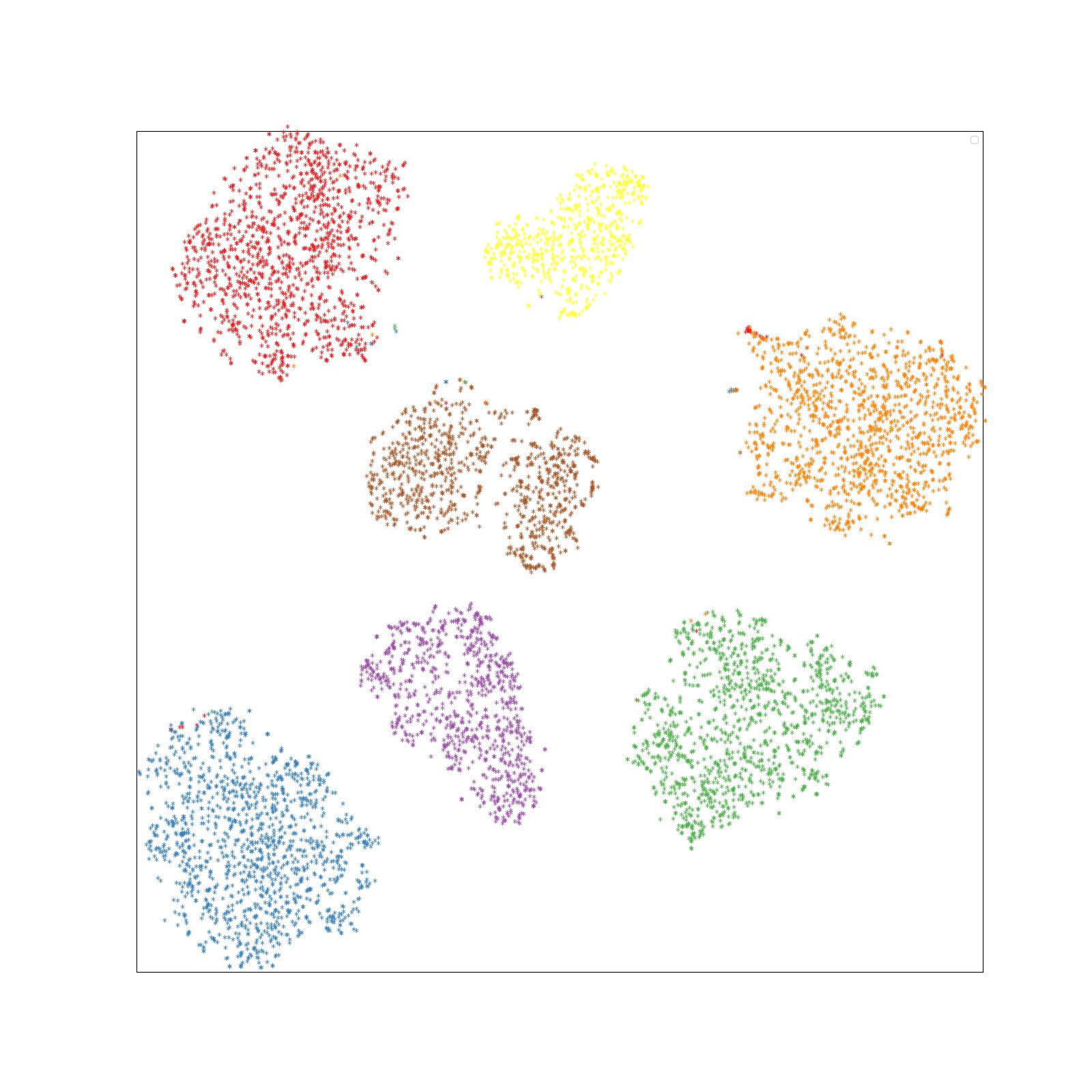}\vspace{4pt}
\includegraphics[width=1\linewidth,trim={5cm 2cm 2cm 4cm},clip]{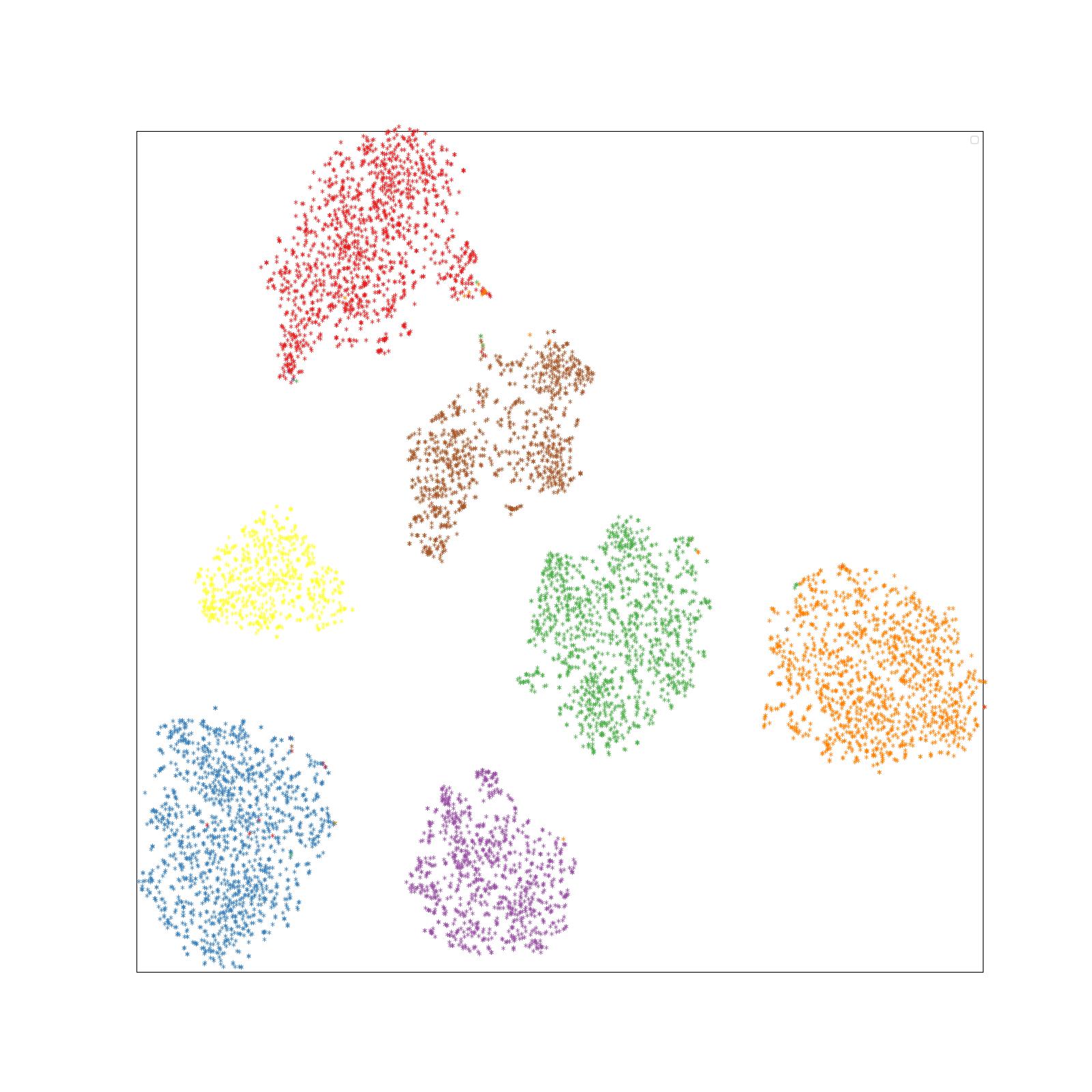}
\end{minipage}}
\subcaptionbox{Art,Photo, Sketch}{
\begin{minipage}[b]{0.23\linewidth}
\includegraphics[width=0.8\linewidth,trim={0cm 0cm 0cm 0cm},clip]{images/classlabel.png}\vspace{4pt}
\includegraphics[width=1\linewidth,trim={5cm 2cm 2cm 4cm},clip]{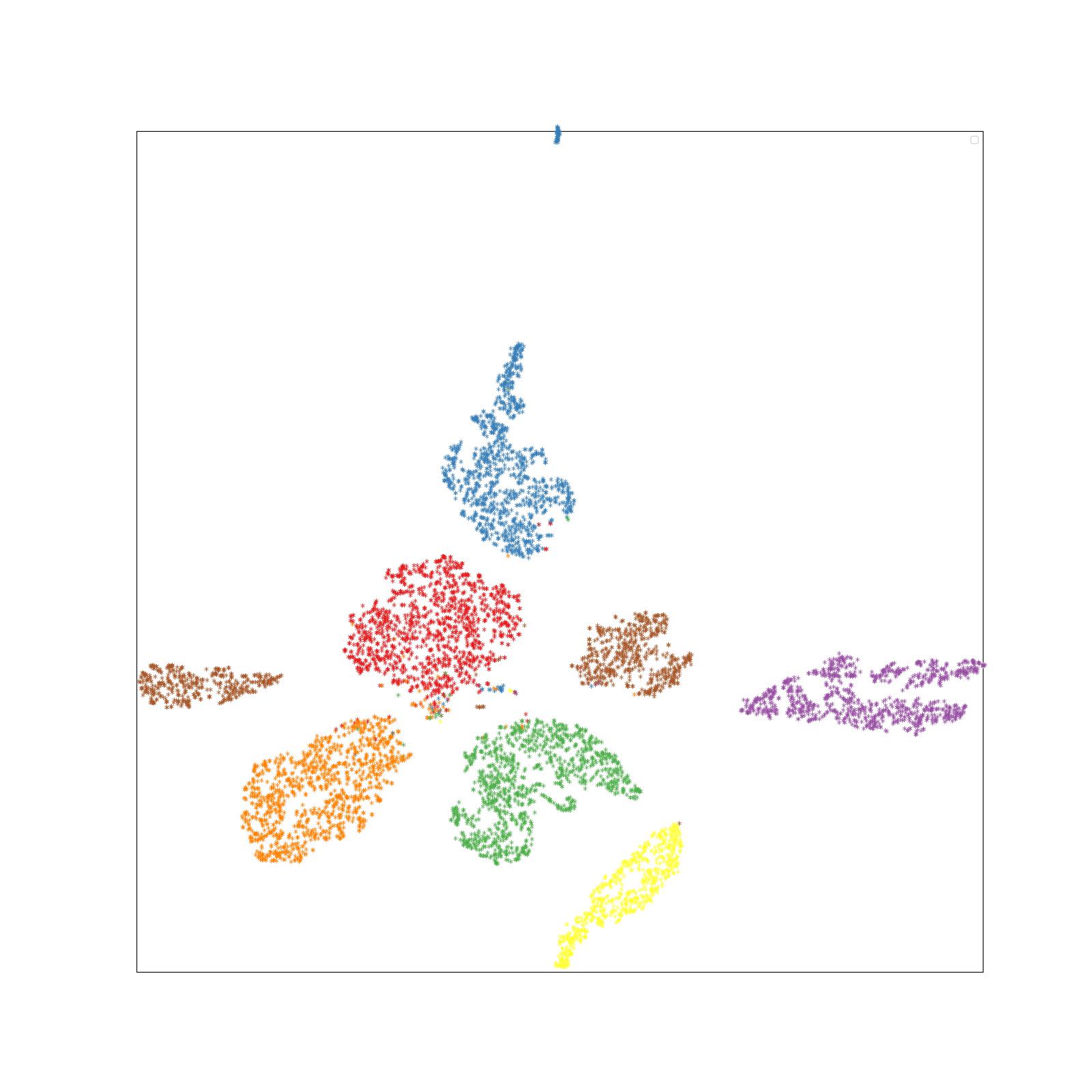}\vspace{4pt}
\includegraphics[width=1\linewidth,trim={5cm 2cm 2cm 4cm},clip]{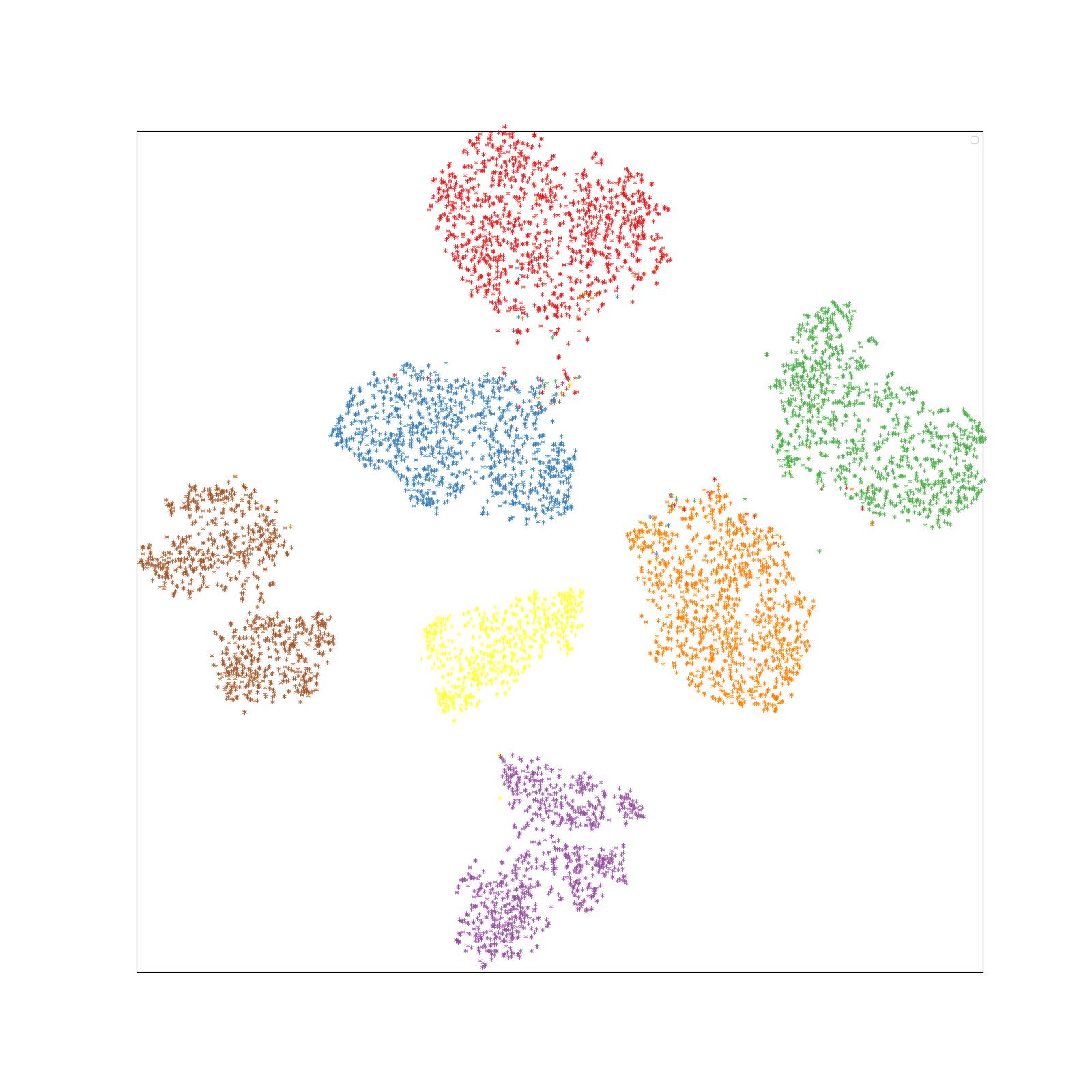}\vspace{4pt}
\includegraphics[width=1\linewidth,trim={5cm 2cm 2cm 4cm},clip]{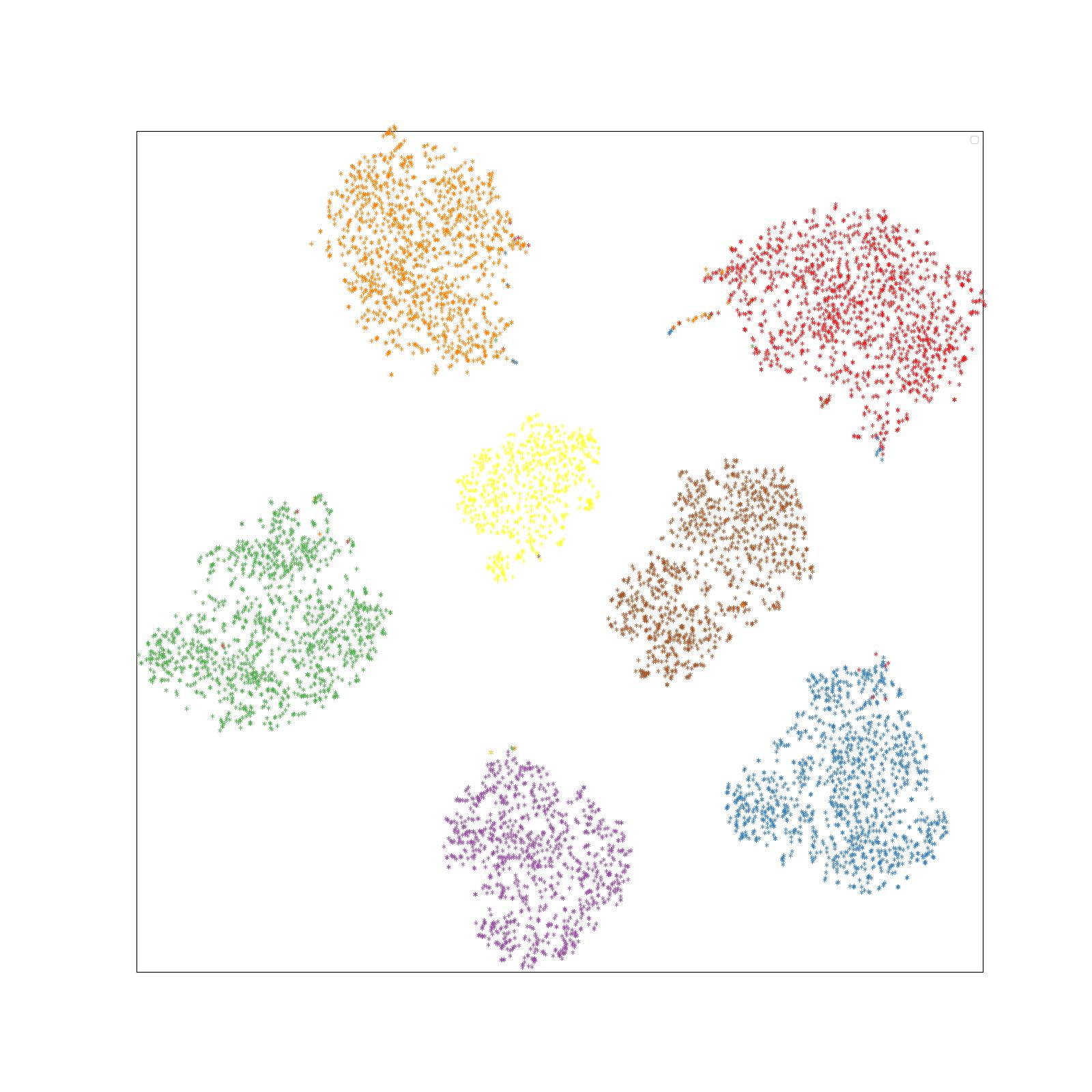}\vspace{4pt}
\includegraphics[width=1\linewidth,trim={5cm 2cm 2cm 4cm},clip]{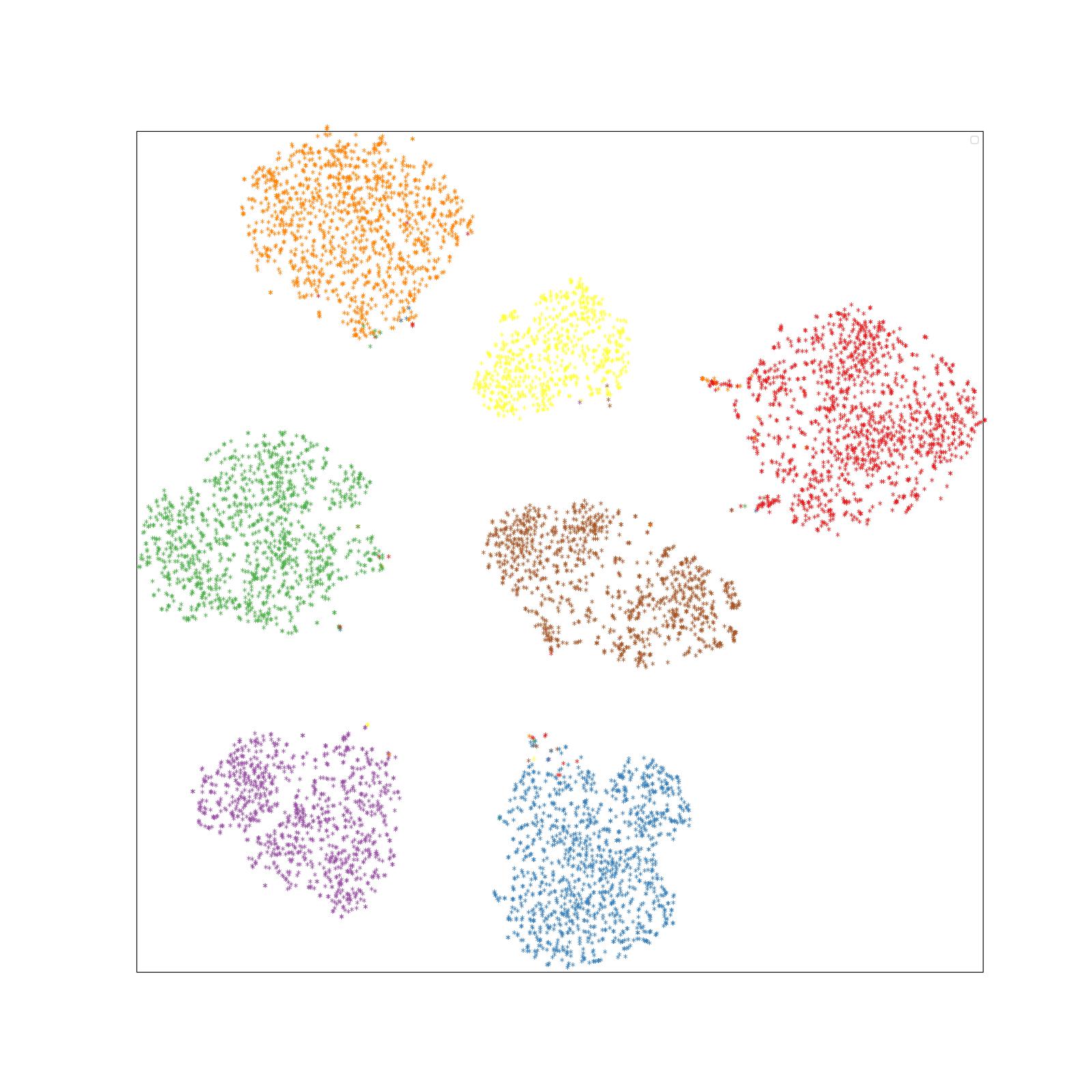}
\end{minipage}}
\subcaptionbox{Art,Cartoon,Sketch}{
\begin{minipage}[b]{0.23\linewidth}
\includegraphics[width=0.8\linewidth,trim={0cm 0cm 0cm 0cm},clip]{images/classlabel.png}\vspace{4pt}
\includegraphics[width=1\linewidth,trim={5cm 2cm 2cm 4cm},clip]{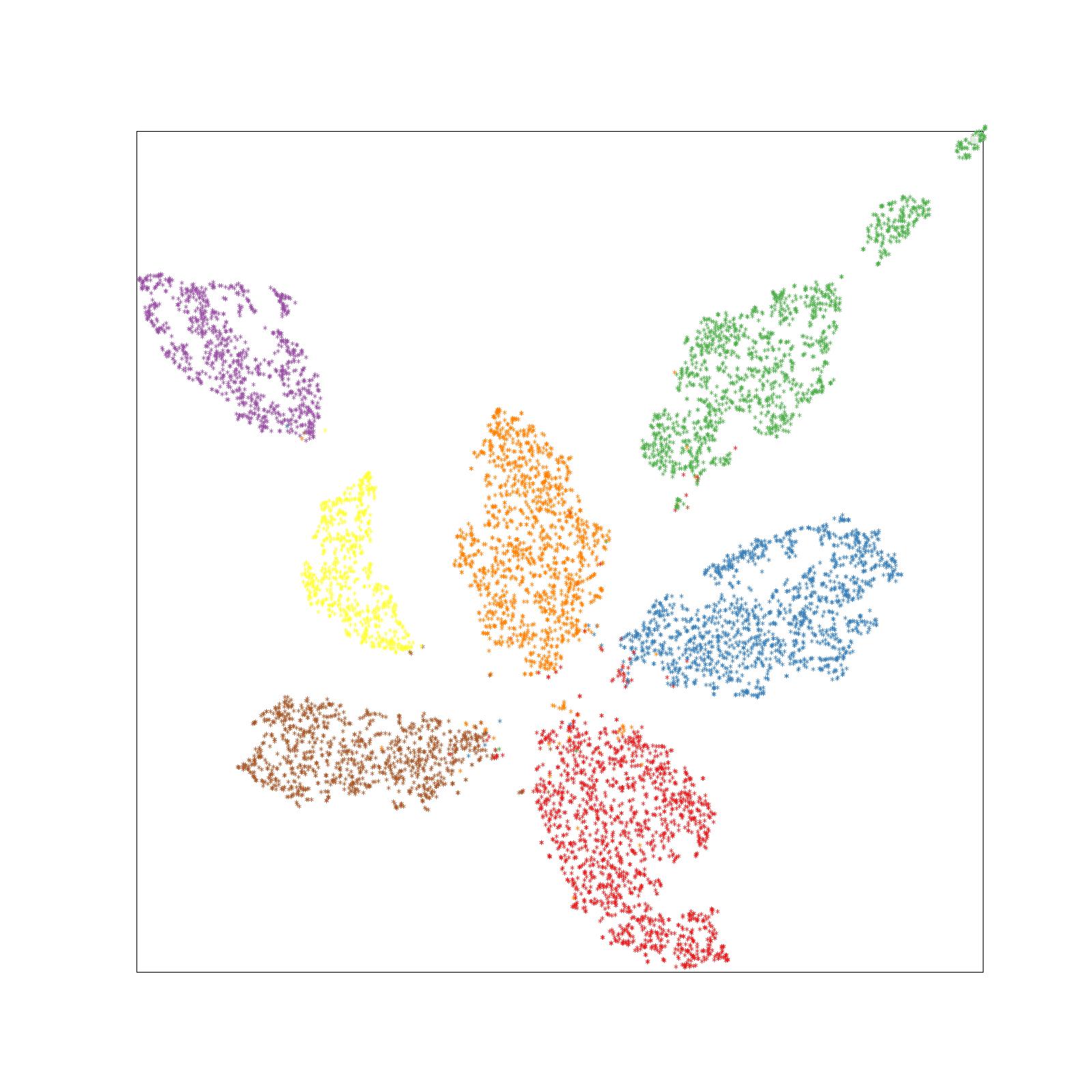}\vspace{4pt}
\includegraphics[width=1\linewidth,trim={5cm 2cm 2cm 4cm},clip]{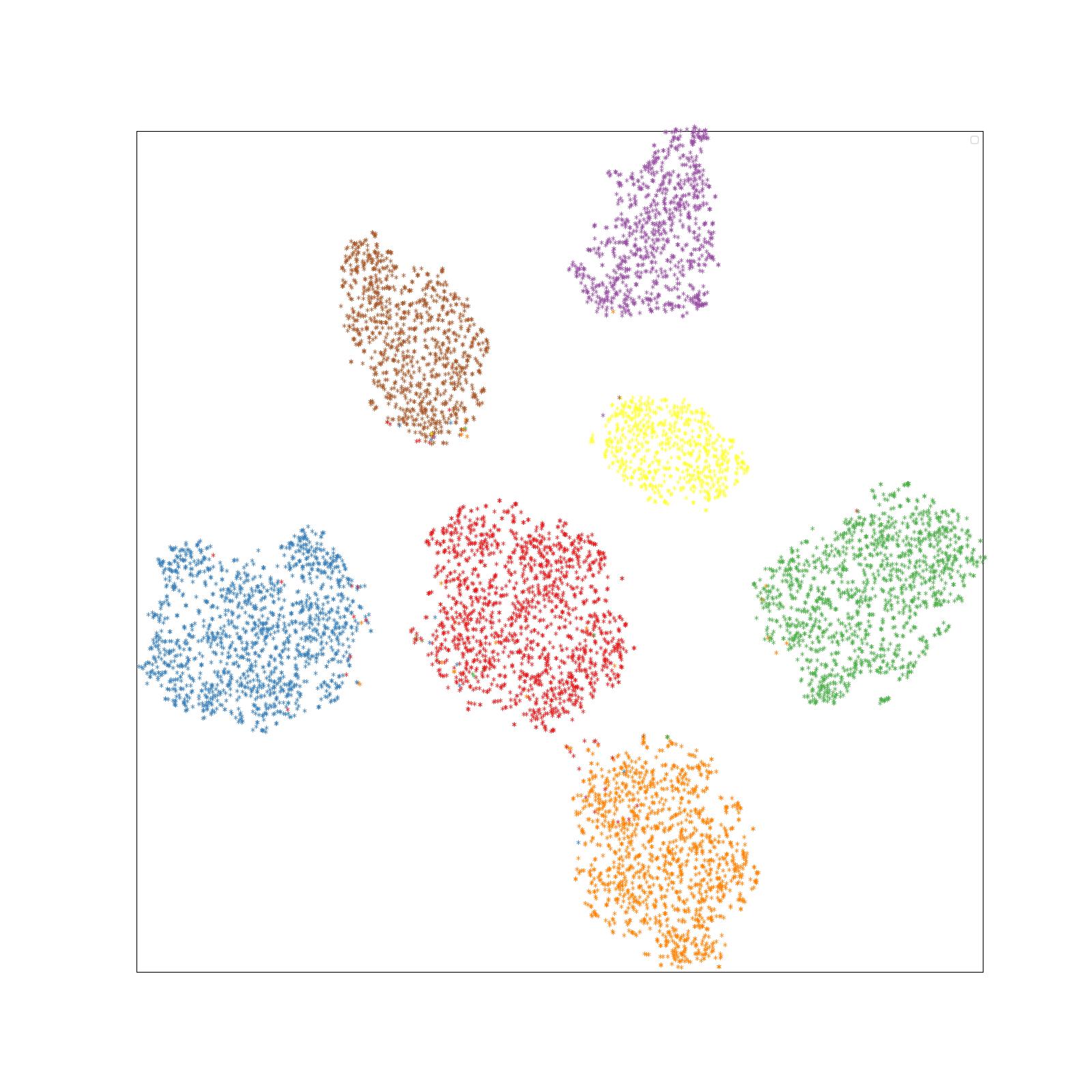}\vspace{4pt}
\includegraphics[width=1\linewidth,trim={5cm 2cm 2cm 4cm},clip]{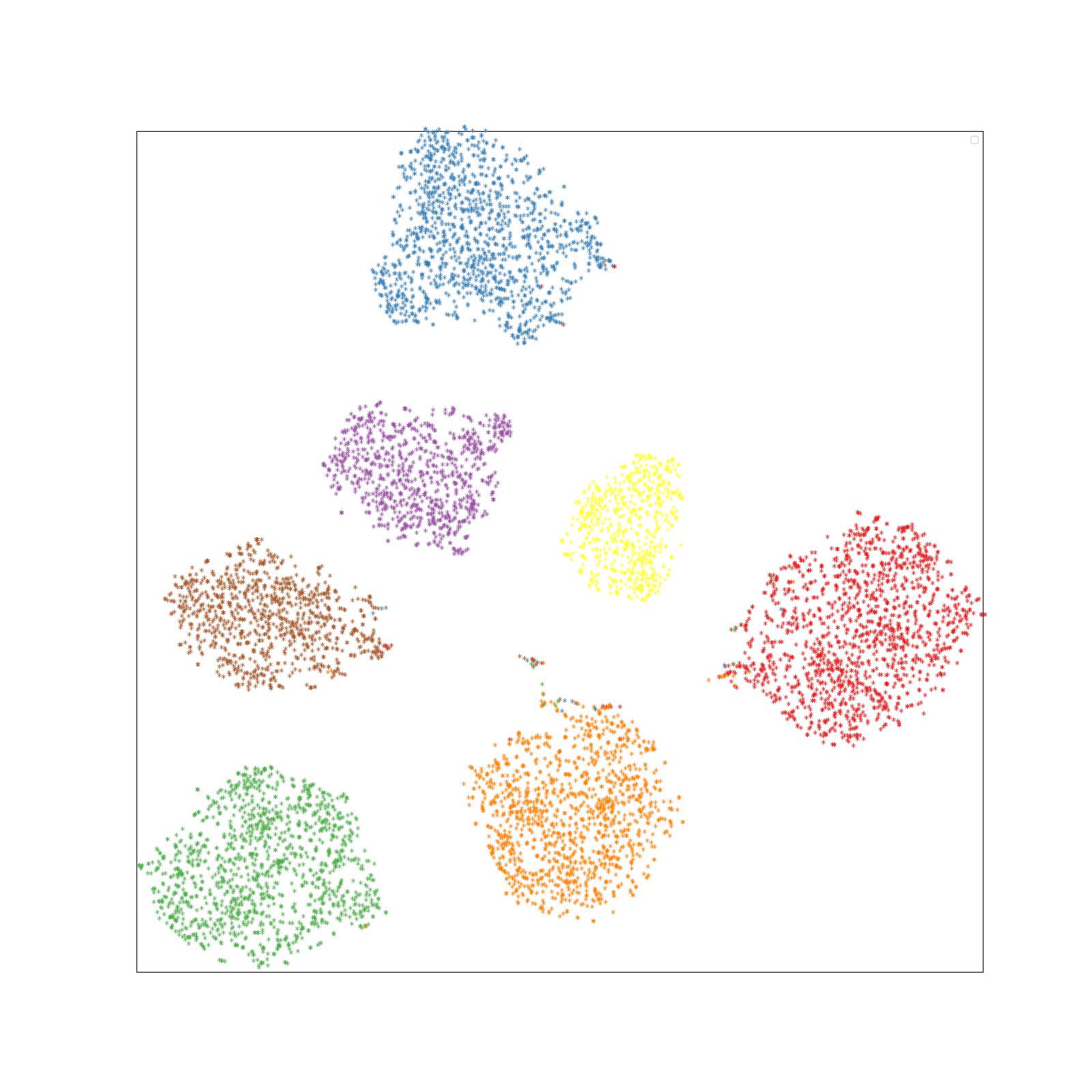}\vspace{4pt}
\includegraphics[width=1\linewidth,trim={5cm 2cm 2cm 4cm},clip]{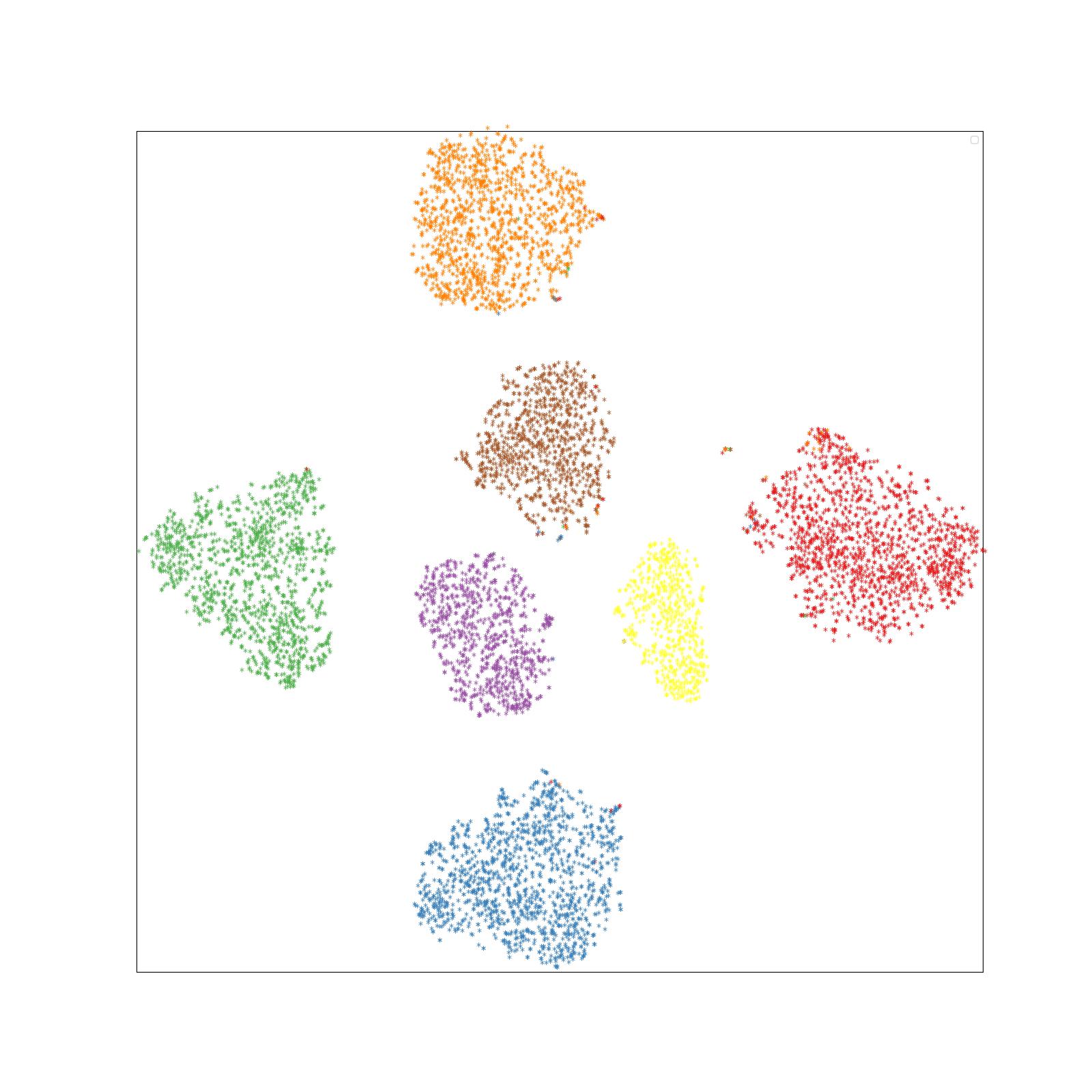}
\end{minipage}}
\subcaptionbox{Art,Cartoon,Photo}{
\begin{minipage}[b]{0.23\linewidth}
\includegraphics[width=0.8\linewidth,trim={0cm 0cm 0cm 0cm},clip]{images/classlabel.png}\vspace{4pt}
\includegraphics[width=1\linewidth,trim={5cm 2cm 2cm 4cm},clip]{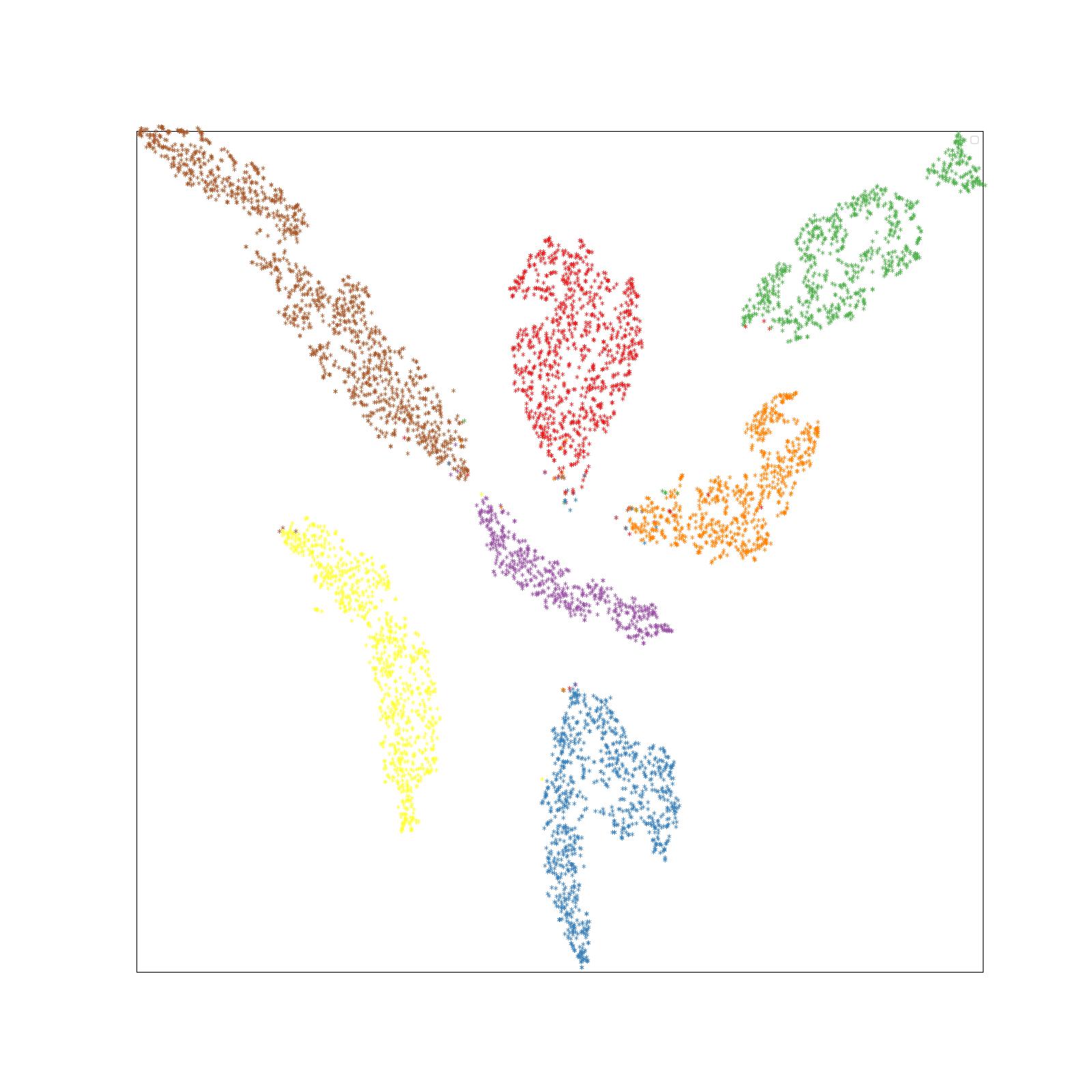}\vspace{4pt}
\includegraphics[width=1\linewidth,trim={5cm 2cm 2cm 4cm},clip]{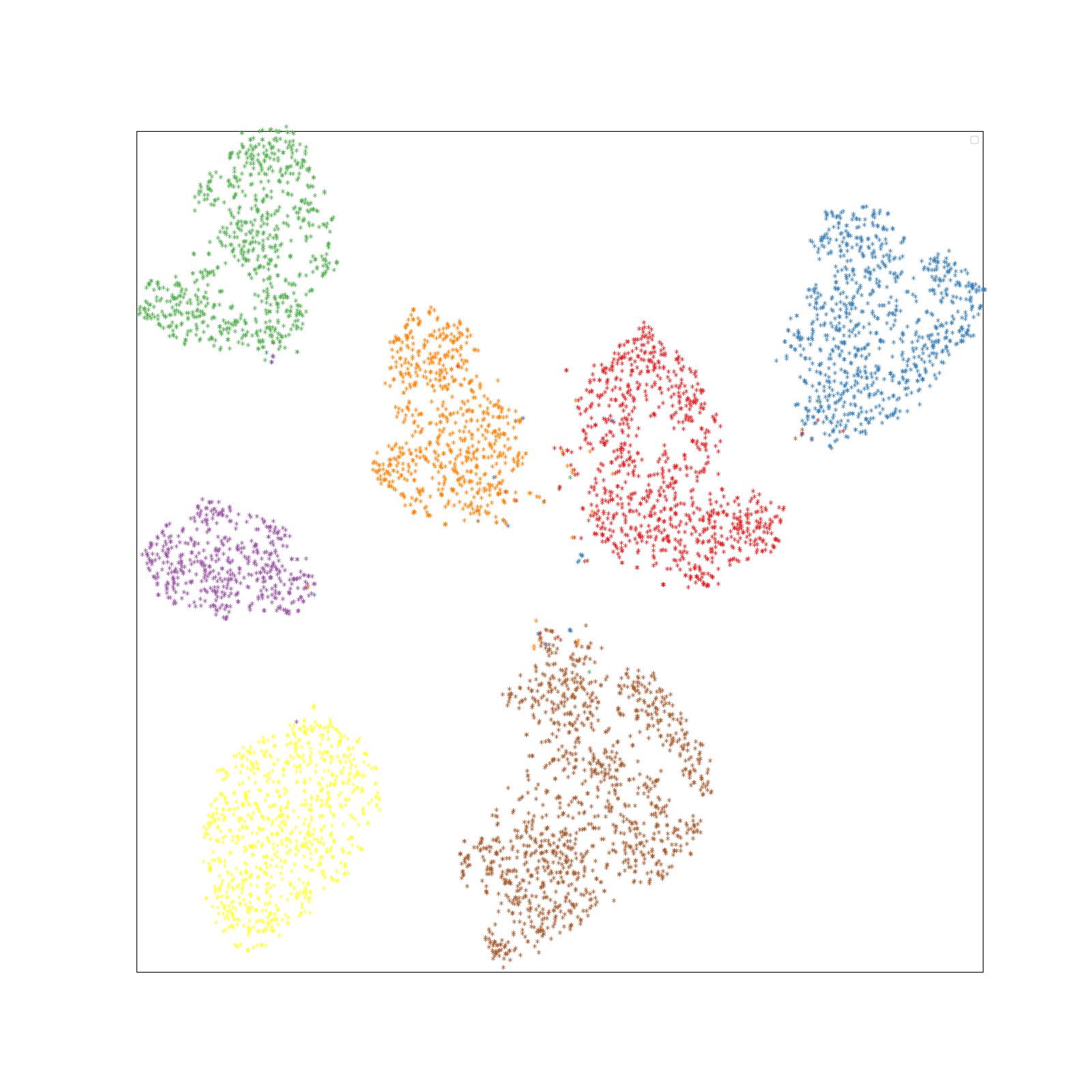}\vspace{4pt}
\includegraphics[width=1\linewidth,trim={5cm 2cm 2cm 4cm},clip]{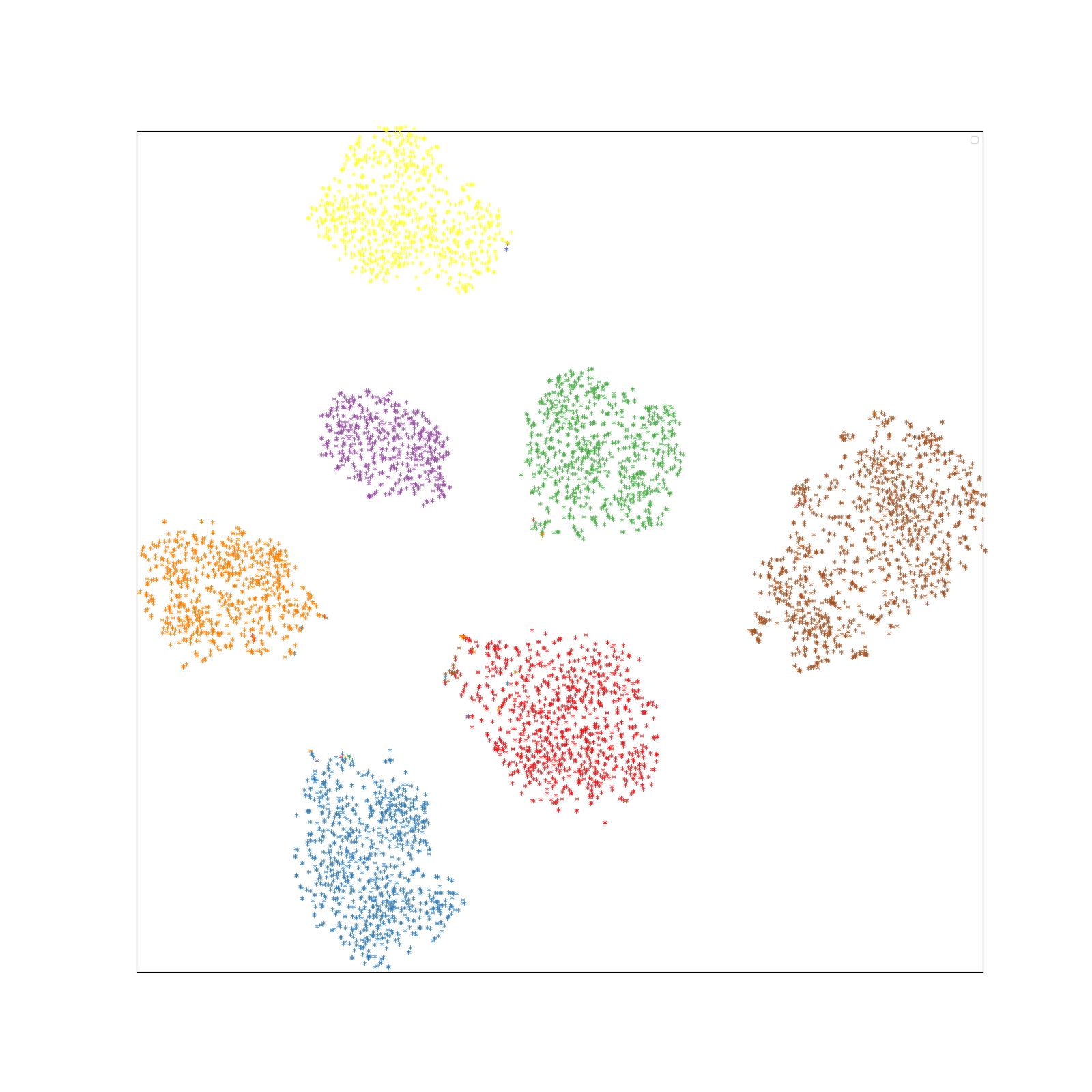}\vspace{4pt}
\includegraphics[width=1\linewidth,trim={5cm 2cm 2cm 4cm},clip]{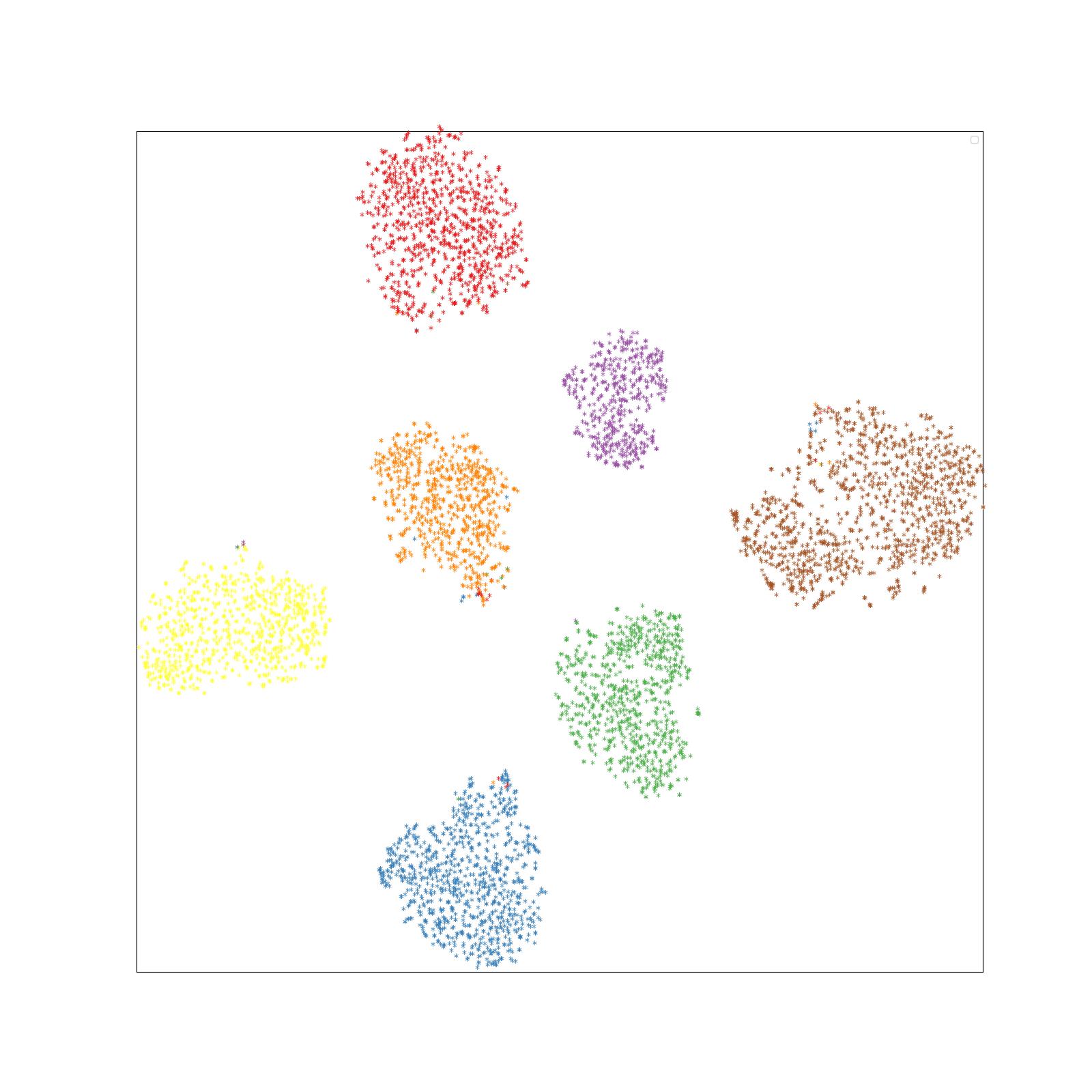}
\end{minipage}}
\end{figure*}
\subsection{Ablation Study}
The results in the ablation study are averaged over three matching trial seeds.\par
\begin{table}[h]
    \centering
    \caption{Ablation study over our algorithm on PACS.}
    \label{tab:ablation}
    \resizebox{0.45\textwidth}{10mm}{
    \begin{tabular}{c c c c c c}
    \hline
         &\textbf{Art}&\textbf{Cartoon}&\textbf{Photo}&\textbf{Sketch}&\textbf{avg}\\
         \hline
    CE&84.0&80.9& 96.5&77.1&84.6\\
    CE+BN&84.5&78.7&95.5&78.2&84.2\\
    CE+BN+Triplet&84.9&81.1&95.5&82.3&85.9 \\
    CE+BN+DCT&87.6&81.8& 97.0&83.1&87.3 \\
    \hline
    \end{tabular}}    
\end{table}
Table \ref{tab:ablation} shows the ablation study over different parts in our algorithm on the PACS dataset.  As the table shows, even though the normalized embedding space will help feature clustering in domain generalization, it has a negative effect on performance when testing on the Photo domain. We assume that the reason for this phenomenon is that the normalized space will change the original distribution in the embedding space, which will diminish the domain information in the pretrained model.  Also, Table \ref{tab:ablation} shows that the domain-aware triplet loss will significantly improve the ability of domain generalization in the model. In addition, \ref{FigPACScluser} and \ref{FigPACScluserid} presents the visualization of the embedding space corresponding to \ref{tab:ablation}\par
\begin{table}[h]
    \centering
    \caption{Ablation study over normalized embedding space on PACS.}
    \label{tab:ablation2}
    \resizebox{0.45\textwidth}{10mm}{
    \begin{tabular}{c c c c c c}
    \hline
         &\textbf{Art}&\textbf{Cartoon}&\textbf{Photo}&\textbf{Sketch}&\textbf{avg}\\
         \hline
    after&87.6&81.8& 97.0&83.1&87.3\\
    before&88.1 &81.8 &96.6&81.7&87.1\\
    FN& 85.1&75.8&98.3&72.1&85.3\\ 
    after+FN&86.1&76.8&97.9&73.5&83.6 \\
    before+FN& 85.5&76.6&98.1&74.9&83.8\\
    \hline
    \end{tabular}}
\end{table}
\begin{table}[h]
    \centering
    \caption{Ablation study over normalized embedding space on VLCS.}
    \label{tab:ablation3}
    \resizebox{0.45\textwidth}{9mm}{
    \begin{tabular}{c c c c c c}
    \hline         &\textbf{Caltech101}&\textbf{LabelMe}&\textbf{SUN09}&\textbf{VOC2007}&\textbf{avg}\\
         \hline    
    w/o FN&-&- &-&-&-\\
    FN & 98.0&66.1&74.2&76.7&78.8\\
    after+FN&98.9&65.2&73.0&77.9&78.7 \\
    before+FN&98.7&66.3&74.3&77.9&79.3 \\
    \hline
    \end{tabular}}
\end{table}
We also perform an ablation study over different normalized embedding spaces on VLCS and PACS. We also consider the situation of whether to use features before the batch normalization layer for the classifier. In our experiments, different datasets may require different embedding-space-normalization methods. PACS is a datasets with stylistic diversity and VLCS is a dataset containing only real images from different sources. For VLCS dataset, the feature normalization plays a critical role instead of batch normalization, but for PACS, the situation is quite the opposite. Actually, the loss can not converge without feature normalization on VLCS. Tables \ref{tab:ablation2} and \ref{tab:ablation3} show the results with different normalized embedding spaces. In the tables, `after' means the input of the classifier is batch normalized, and `before' means the input of the classifier is not batch normalized. 

\subsection{Comparison with other methods}
\begin{table}[h]
\caption{Comparison on PACS dataset with Resnet50 and RegnetY 16GF backbones}
    \label{PACS}
\resizebox{0.45\textwidth}{32mm}{ 
    \begin{tabular}{c c c c c c}
    \hline
    \textbf{Algorithm} &\textbf{Art}&\textbf{Cartoon}&\textbf{Photo}&\textbf{Sketch}&\textbf{avg}\\
    \hline
    \multicolumn{5}{c}{Resnet50}\\
    \hline
    PCL(w/o SWAD) \cite{pcl}&80.6&79.8&96.1&77.2&83.4\\
    IRM \cite{IRM}& 84.8&76.4&96.7&76.1&83.5\\
    ERM \cite{erm}& 85.7&77.1&97.4&76.6&84.2\\
    ERM(reprodeuced)&84.0&80.9& 96.5&77.1&84.6\\
    MMD \cite{mmd}&86.1&79.4&96.6&76.5&84.7\\
    MSL \cite{MSL}&84.2&80.0&95.7&80.1&85.0 \\
    Mixstyle \cite{mixstyle}& 86.8&79.0&96.6&78.5&85.2\\ 
    RSC \cite{rsc}& 85.4&79.7&97.6&78.2&85.2\\
    ER \cite{ER}& 87.5& 79.3& \textbf{98.3}& 76.3& 85.3\\  
    MIRO \cite{miro}&87.4&78.3&97.2&78.7&85.4	 \\
    SagNet \cite{sagnet}&87.4&80.7&97.1&80.0&86.3\\
    CORAL \cite{coral}& \textbf{88.3}&80.0&97.5&78.8&86.2\\
    DSON \cite{dson}& 87.0& 80.6& 96.0& 82.9& 86.6\\ 
    
    \hline
    DCT (ours)& 87.6&\textbf{81.8}& 97.0&\textbf{83.1}&\textbf{87.3}\\
    \hline
    \multicolumn{5}{c}{RegnetY 16GF}\\
    \hline
    ERM \cite{erm}&-&-&-&-&89.6\\
    ERM+SWAD\cite{swad}&-&-&-&-&94.7\\
    MIRO \cite{miro}&-&-&-&-&97.4\\
    \hline
    DCT (ours)&98.3&98.5&99.9&93.8&\textbf{97.6}\\
    
    \hline
    \end{tabular}}
\end{table}
\begin{table}[h]
\caption{Comparison on OfficeHome dataset with Resnet50 and RegnetY 16GF backbones}
    \label{officehome}
\resizebox{0.45\textwidth}{26mm}{ 
    \begin{tabular}{c c c c c c}
    \hline
    \textbf{Algorithm} &\textbf{Art}&\textbf{Clipart}&\textbf{Product}&\textbf{Real World}&\textbf{avg}\\
    \hline
    \multicolumn{5}{c}{Resnet50}\\
    \hline
    Mixstyle \cite{mixstyle}&  51.1& 53.2& 68.2& 69.2& 60.4\\ 
    IRM \cite{IRM}& 58.9&52.2&72.1&74.0&64.3\\
    ARM \cite{arm}& 58.9& 51.0& 74.1& 75.2& 64.8\\
    RSC \cite{rsc}&  60.7&51.4&74.8&75.1&65.5\\
    MMD \cite{mmd}& 60.4& 53.3& 74.3& 77.4& 66.4\\
    ERM \cite{erm}&  63.1& 51.9& 77.2& 78.1& 67.6\\
    PCL(w/o SWAD) \cite{pcl}&64.1&53.9&75.3&78.4&67.9\\
    SagNet \cite{sagnet}& 63.4&54.8&75.8&78.3&68.1\\
    CORAL \cite{coral}&  65.3&54.4&76.5&78.4&68.7\\    
    MIRO \cite{miro}&\textbf{67.5}&54.6&\textbf{78.0}&\textbf{81.6}	&\textbf{70.5}\\
    
    \hline
    DCT (ours) &67.0 &\textbf{54.9} &\textbf{78.0}&81.1&70.3\\
    \hline
    \multicolumn{5}{c}{RegnetY 16GF}\\
    \hline
    ERM \cite{erm}&-&-&-&-&71.9\\
    ERM+SWAD\cite{swad}&-&-&-&-&80.0\\
    MIRO \cite{miro}&-&-&-&-&80.4\\
    \hline
    DCT (ours)&81.3&69.9&89.1&89.9&\textbf{82.6}\\
    
    \hline
    \end{tabular}}
\end{table}
\begin{table}[h]
\caption{Comparison on VLCS dataset with Resnet50 and RegnetY 16GF}
    \label{VLCS}
\resizebox{0.50\textwidth}{26mm}{ 
    \begin{tabular}{c c c c c c}
    \hline
    \textbf{Algorithm} &\textbf{Caltech101}&\textbf{LabelMe}&\textbf{SUN09}&\textbf{VOC2007}&\textbf{avg}\\
    \hline
    \multicolumn{5}{c}{Resnet50}\\
    \hline
    IRM \cite{IRM}& 98.6& 66.0&69.3& 71.5&76.3\\
    ARM \cite{arm}& 97.2& 62.7& 70.6& 75.8& 76.6\\
    ERM \cite{erm}&  98.0& 62.6& 70.8& 77.5& 77.2\\
    MMD \cite{mmd}& 98.3& 65.6& 69.7& 75.7& 77.3\\
    SagNet \cite{sagnet}&  97.3&61.6& 73.4& 77.6&77.5\\
    RSC \cite{rsc}& 97.5&63.1&73.0&76.2&77.5\\
    CORAL \cite{coral}&  96.9 &65.7& 73.3&\textbf{78.7}&78.7\\ 
    Mixup \cite{mixup}&  98.4& 63.4& 72.9&76.1& 77.7\\ 
    MIRO \cite{miro}&98.3&64.7&\textbf{75.3}&77.8&79.0\\ 
    \hline
    DCT (ours) &\textbf{98.7}&\textbf{66.3}&74.3&77.9&\textbf{79.3}\\
    \hline
    \multicolumn{5}{c}{RegnetY 16GF}\\
    \hline
    ERM \cite{erm}&-&-&-&-&78.6\\
    ERM+SWAD\cite{swad}&-&-&-&-&79.7\\
    MIRO \cite{miro}&-&-&-&-&79.9\\
    \hline
    DCT (ours)&99.1&67.5&81.5&83.4&\textbf{82.9}\\    
    \hline
    \end{tabular}}
\end{table}
\textbf{Results on Domain Generalization benchmark:} DCT loss is an algorithm considering the domain discrepancy in domain generalization. To illustrate the competitiveness of DCT loss, we compare it with other methods similarly solving domain discrepancy in domain generalization. Tables \ref{PACS}, \ref{officehome} and \ref{VLCS} show this comparison on PACS, OfficeHome and VLCS dataset respectively.  The margin of DCT loss is set to 15 for the PACS dataset and 0 for the other two datasets. Notice that the result of PCL is reproduced from their released code without the SWAD algorithm and MSL is reproduced with the same hyperparameters claimed in \cite{MSL} on PACS dataset. The result of SWAD is from MIRO\cite{miro}. We evaluate our algorithm with two backbones, namely ImageNet pretrained resnet50 \cite{resnet} and the pretrained RegNet(SWAG) \cite{regnet}. Only feature normalization is used to normalize the embedding space on RegNet. The results of other methods in Tables \ref{PACS}, \ref{officehome} are from previous research\cite{pcl, miro} and the results of other methods in Table \ref{VLCS} are from  MIRO and Domainbed~\cite{miro, domainbed}. The methods we compare with include traditional methods ~\cite{erm, IRM}, data augmentation methods ~\cite{mixstyle, mixup}, representation learning methods~\cite{coral,IRM} and some state-of-the-art methdos~\cite{pcl,miro}. In Table \ref{PACS}, \ref{officehome} and \ref{VLCS}, our DCT loss performs in the first tier on average accuracy and outperforms with the backbone of RegNet. In addition, as a representation learning method, our algorithm also  outperforms other representation learning methods on DomainBed overall.  \par
\begin{figure*}
\centering 
\caption{Visualization of embedding space withDCT and MSL~\cite{MSL}. Settings are same with former visualization. The first two rows are labeled by domain and the last two rows are labeled by class. The first and third rows are the results  of MSL. The second and fourth rows are the results of DCT.} 
\label{FigPACSmsl} 
\subcaptionbox{Cartoon,Photo, Sketch}{
\begin{minipage}[b]{0.23\linewidth}
\includegraphics[width=0.8\linewidth,trim={0cm 0cm 0cm 0cm},clip]{images/label1.png}\vspace{1pt}
\includegraphics[width=1\linewidth,height=0.9\linewidth,trim={5cm 2cm 2cm 4cm},clip]{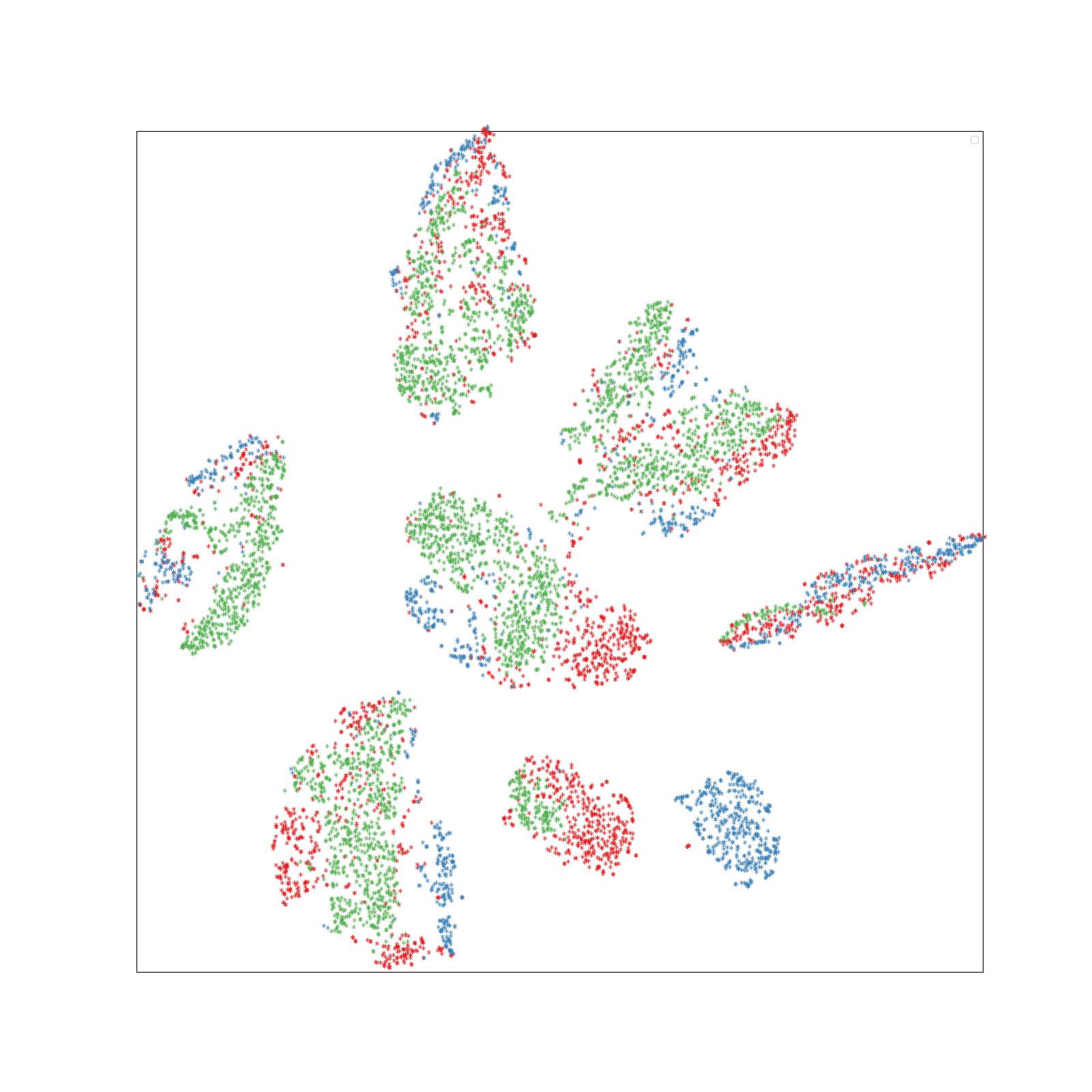}\vspace{1pt}
\includegraphics[width=1\linewidth,height=0.9\linewidth,trim={5cm 2cm 2cm 4cm},clip]{images/DCT-15/eval0_show_tsne.jpg}\vspace{1pt}
\includegraphics[width=0.8\linewidth,trim={0cm 0cm 0cm 0cm},clip]{images/classlabel.png}\vspace{1pt}
\includegraphics[width=1\linewidth,height=0.9\linewidth,trim={5cm 2cm 2cm 4cm},clip]{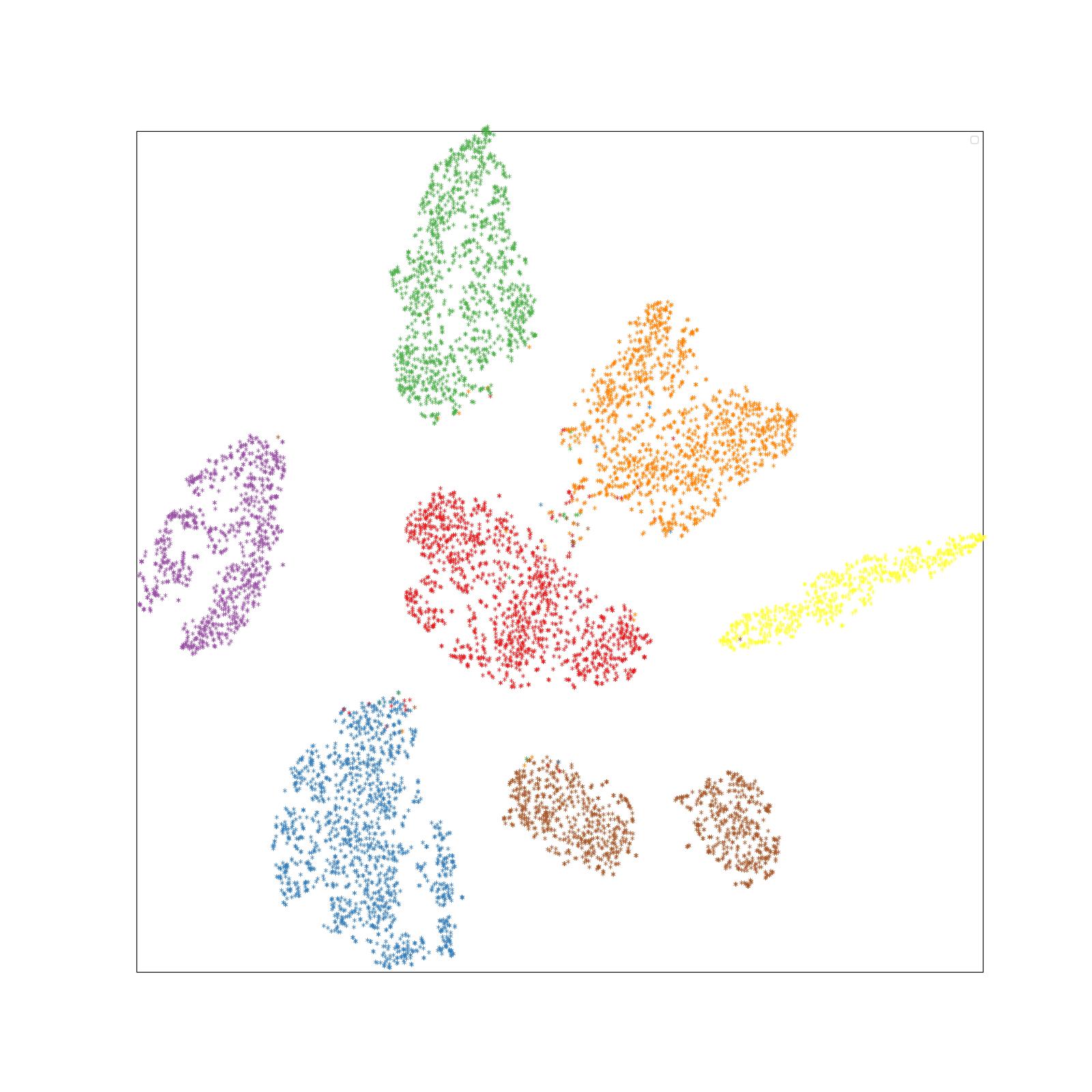}\vspace{1pt}
\includegraphics[width=1\linewidth,height=0.9\linewidth,trim={5cm 2cm 2cm 4cm},clip]{images/DCT-15-id/eval0_show_tsne.jpg}
\end{minipage}}
\subcaptionbox{Art,Photo, Sketch}{
\begin{minipage}[b]{0.23\linewidth}
\includegraphics[width=0.8\linewidth,trim={0cm 0cm 0cm 0cm},clip]{images/label2.png}\vspace{1pt}
\includegraphics[width=1\linewidth,height=0.9\linewidth,trim={5cm 2cm 2cm 4cm},clip]{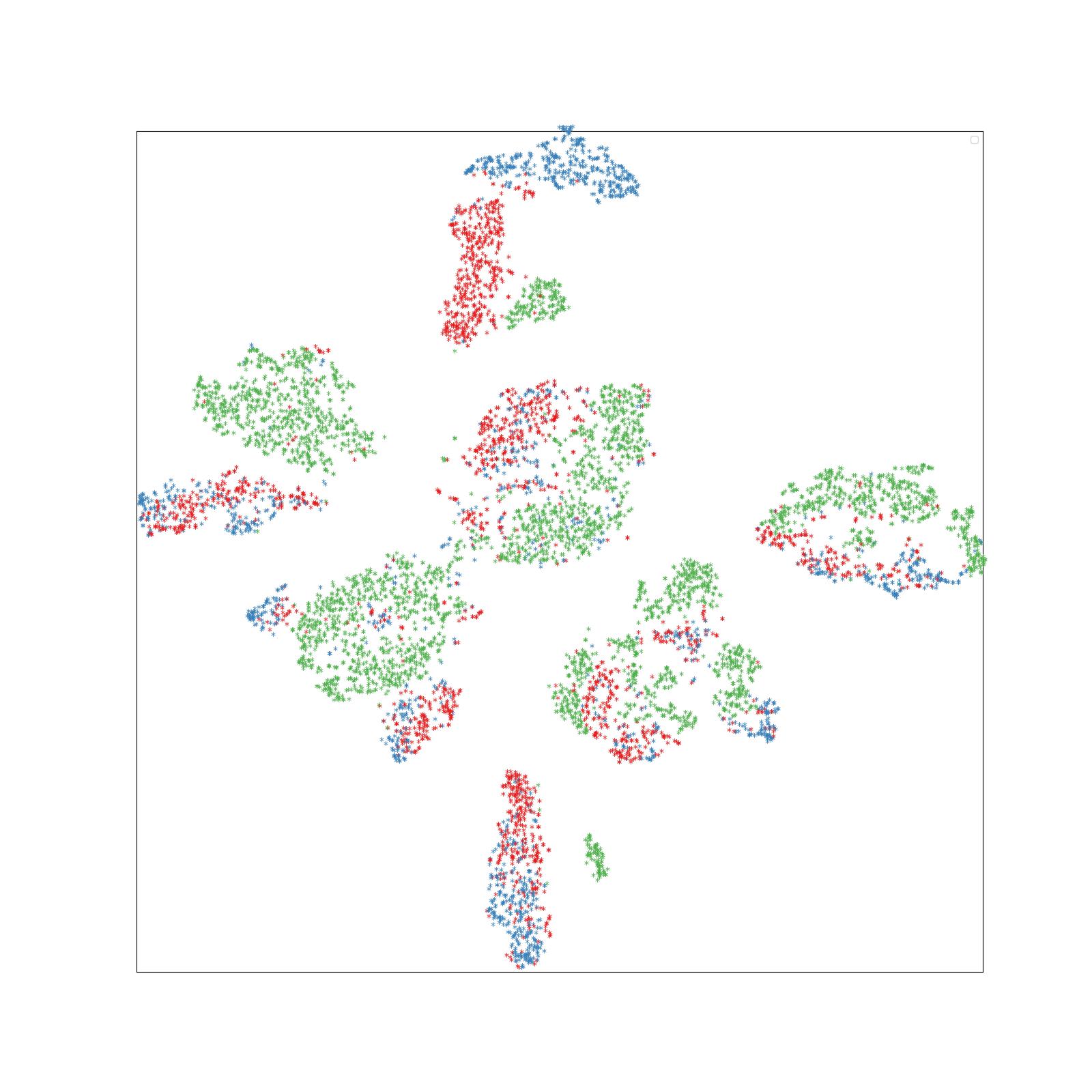}\vspace{1pt}
\includegraphics[width=1\linewidth,height=0.9\linewidth,trim={5cm 2cm 2cm 4cm},clip]{images/DCT-15/eval1_show_tsne.jpg}\vspace{1pt}
\includegraphics[width=0.8\linewidth,trim={0cm 0cm 0cm 0cm},clip]{images/classlabel.png}\vspace{1pt}
\includegraphics[width=1\linewidth,height=0.9\linewidth,trim={5cm 2cm 2cm 4cm},clip]{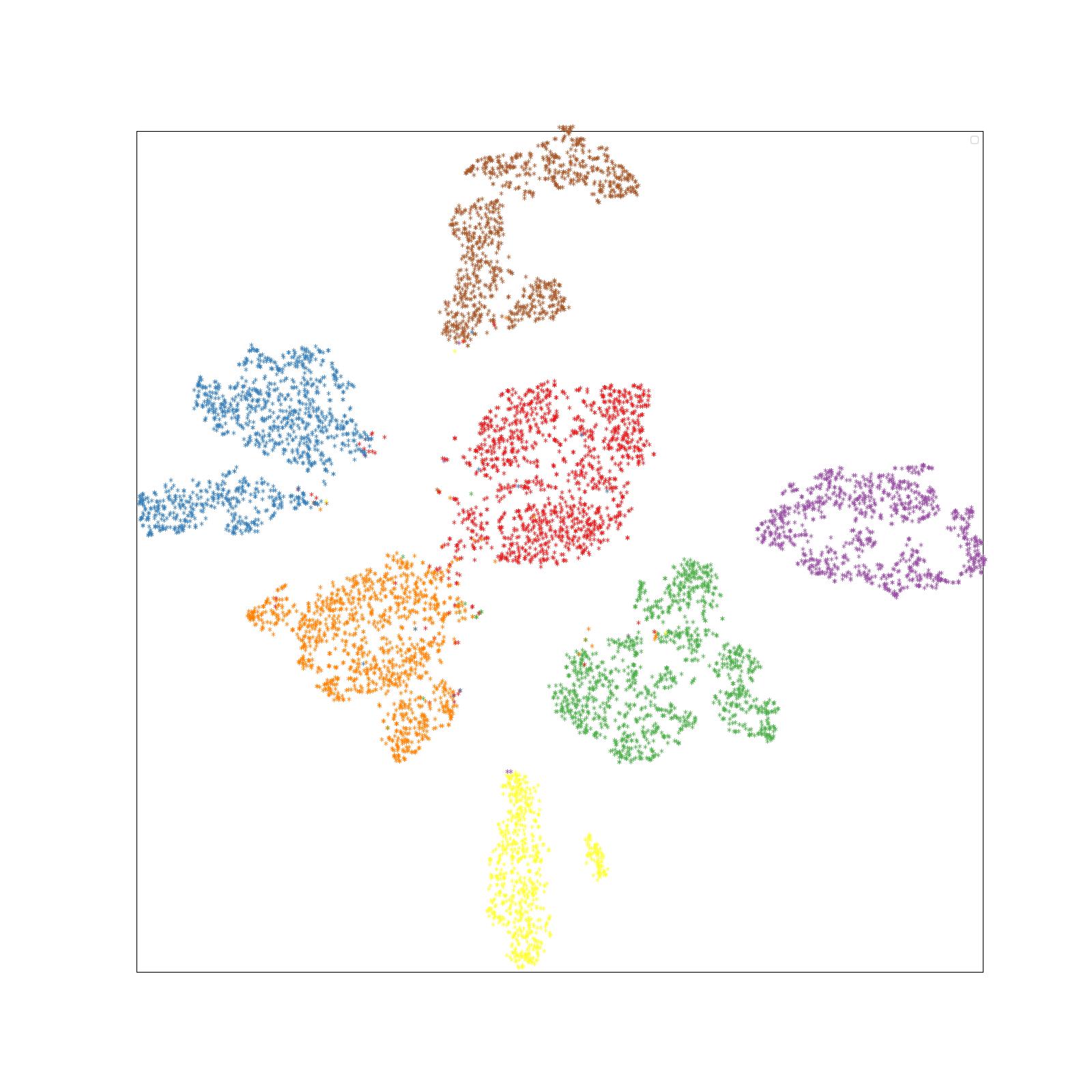}\vspace{1pt}
\includegraphics[width=1\linewidth,height=0.9\linewidth,trim={5cm 2cm 2cm 4cm},clip]{images/DCT-15-id/eval1_show_tsne.jpg}
\end{minipage}}
\subcaptionbox{Art,Cartoon,Sketch}{
\begin{minipage}[b]{0.23\linewidth}
\includegraphics[width=0.8\linewidth,trim={0cm 0cm 0cm 0cm},clip]{images/label3.png}\vspace{1pt}
\includegraphics[width=1\linewidth,height=0.9\linewidth,trim={5cm 2cm 2cm 4cm},clip]{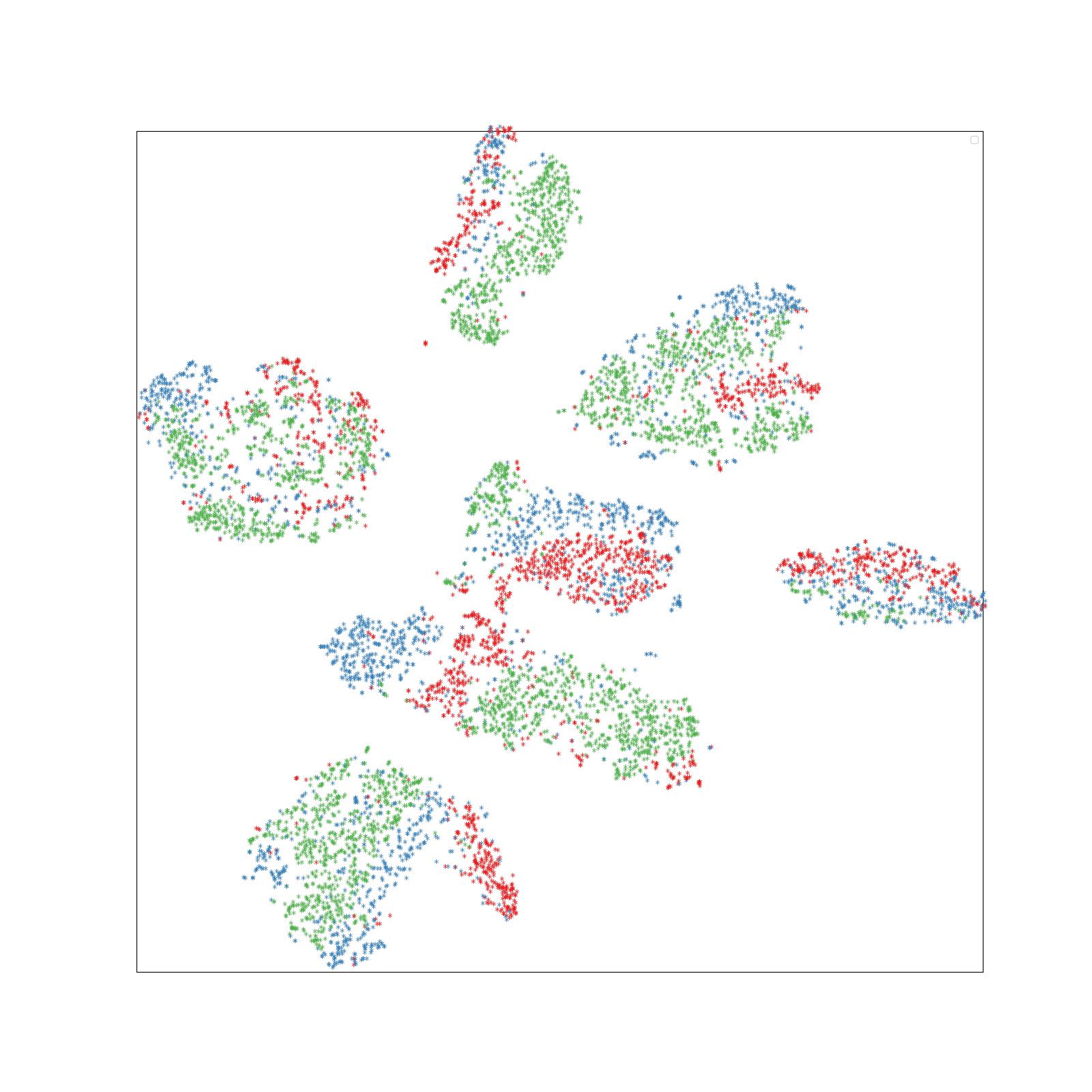}\vspace{1pt}
\includegraphics[width=1\linewidth,height=0.9\linewidth,trim={5cm 2cm 2cm 4cm},clip]{images/DCT-15/eval2_show_tsne.jpg}\vspace{1pt}
\includegraphics[width=0.8\linewidth,trim={0cm 0cm 0cm 0cm},clip]{images/classlabel.png}\vspace{1pt}
\includegraphics[width=1\linewidth,height=0.9\linewidth,trim={5cm 2cm 2cm 4cm},clip]{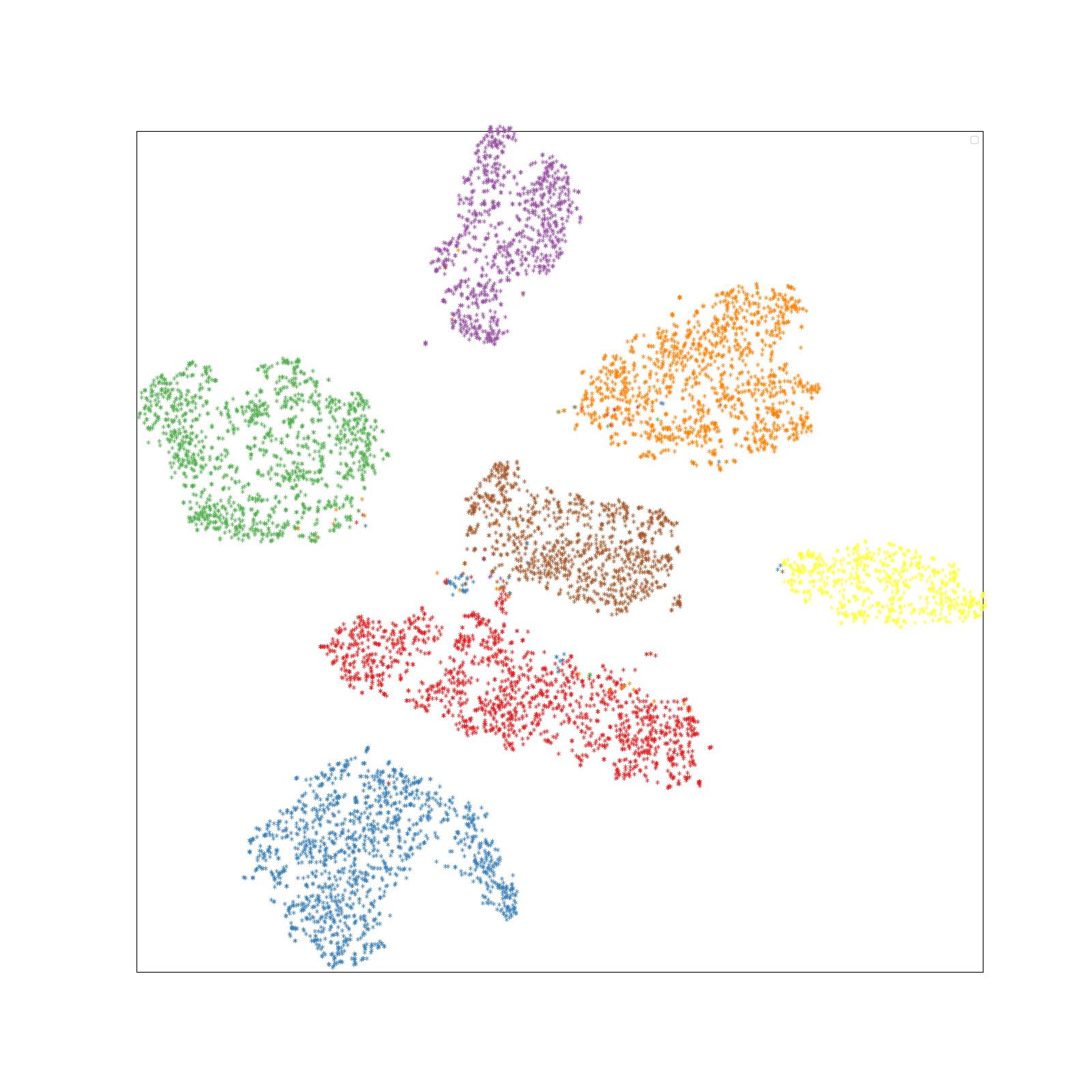}\vspace{1pt}
\includegraphics[width=1\linewidth,height=0.9\linewidth,trim={5cm 2cm 2cm 4cm},clip]{images/DCT-15-id/eval2_show_tsne.jpg}
\end{minipage}}
\subcaptionbox{Art,Cartoon,Photo}{
\begin{minipage}[b]{0.23\linewidth}
\includegraphics[width=0.8\linewidth,trim={0cm 0cm 0cm 0cm},clip]{images/label4.png}\vspace{1pt}
\includegraphics[width=1\linewidth,height=0.9\linewidth,trim={5cm 2cm 2cm 4cm},clip]{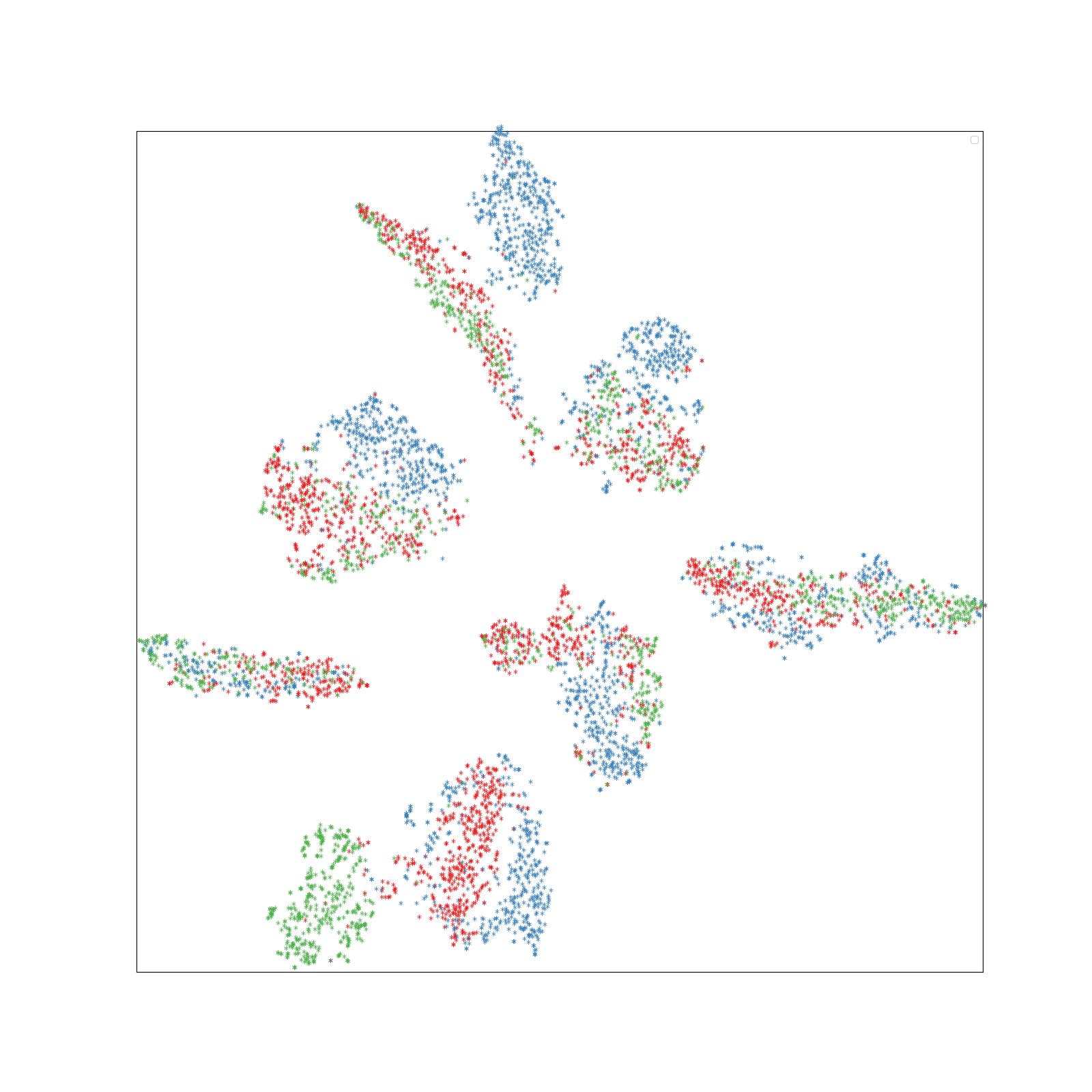}\vspace{1pt}
\includegraphics[width=1\linewidth,height=0.9\linewidth,trim={5cm 2cm 2cm 4cm},clip]{images/DCT-15/eval3_show_tsne.jpg}\vspace{1pt}
\includegraphics[width=0.8\linewidth,trim={0cm 0cm 0cm 0cm},clip]{images/classlabel.png}\vspace{1pt}
\includegraphics[width=1\linewidth,height=0.9\linewidth,trim={5cm 2cm 2cm 4cm},clip]{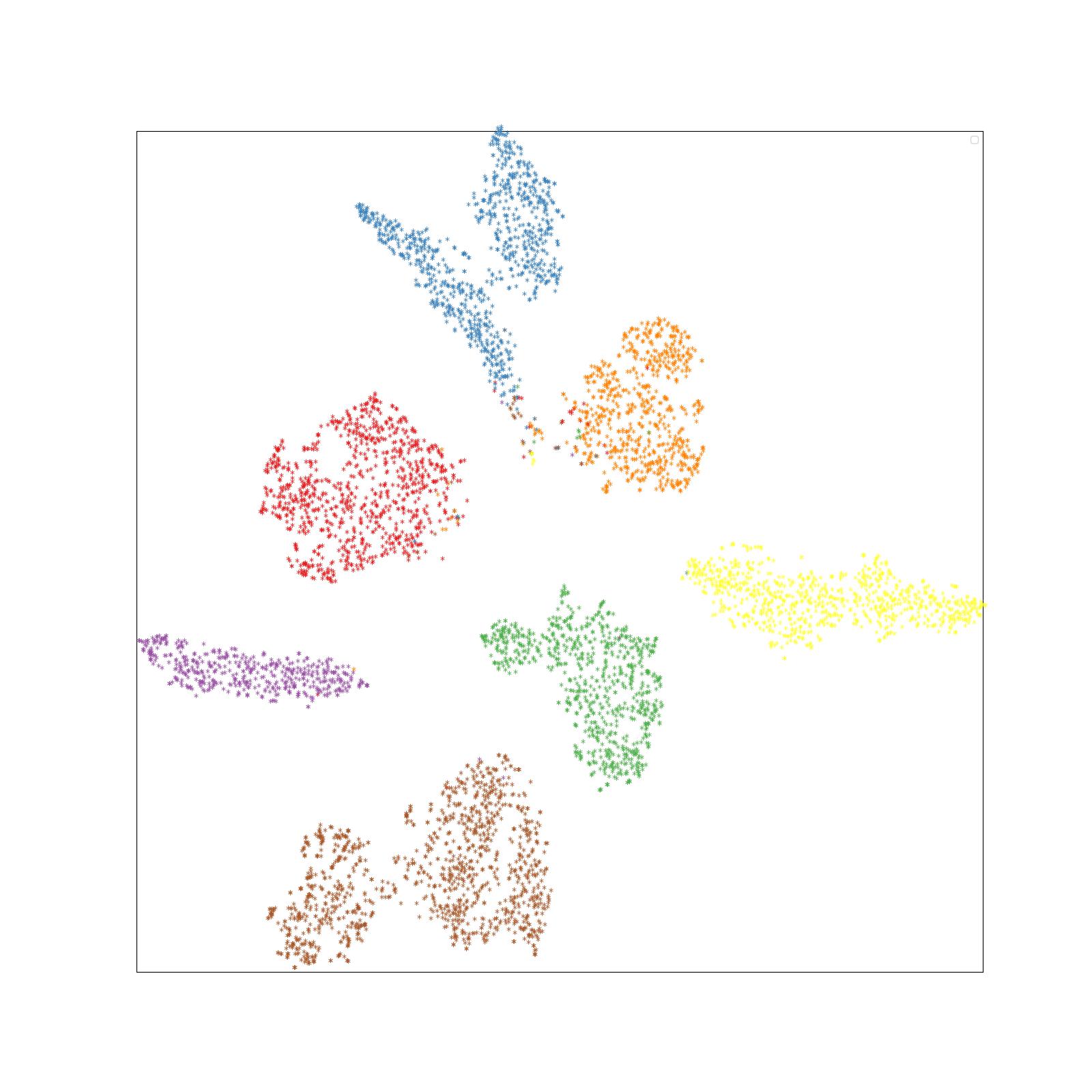}\vspace{1pt}
\includegraphics[width=1\linewidth,height=0.9\linewidth,trim={5cm 2cm 2cm 4cm},clip]{images/DCT-15-id/eval3_show_tsne.jpg}
\end{minipage}}
\end{figure*}

\textbf{Visualization on feature clustering:}
Triplet loss is a very efficient feature clustering penalty function in contrastive learning, but it cannot perform well in domain generalization. There are some research on domain alignment to help feature clustering in domain generalization~\cite{CDT, MSL}. CDT~\cite{CDT} is the triplet loss designed for multi-ethnicity face recognition and MSL~\cite{MSL} is updated from CDT for the task of cross-domain continual learning. The difference between CDT and former methods is that CDT is trying to disperse the original domain clusters or distributions but former methods are trying to align the different domain distributions. To show the performance of distribution alignment methods in feature clustering, we visualize the feature embedding of MSL~\cite{MSL} which is also designed on domain generalization datasets. The visualization of feature clustering with DCT and MSL on PACS is shown in Figure \ref{FigPACSmsl}. \par
As Figure \ref{FigPACSmsl} shows, MSL~\cite{MSL} can cluster the features based on the class labels but can not crush the domain clusters in the embedding space. Also, the compactness of the clusters with MSL is apparently looser than the ones with DCT. 
\section{Discussion and Limitations}
\label{sec:discussion}
The domain generalization gap is bounded by three factors: flat minima, domain discrepancy, and confidence bound \cite{swad}. Considering domain discrepancy, This paper proposes a simple but effective pair-mining method for contrastive learning in domain generalization. We have shown that this method can significantly improve the performance of domain generalization on three datasets. However, we find that the convergence of triplet loss is slow and  more complicated for domain generalization. In this way, the original algorithms seeking flat minima, like SWAD, may not be able to work efficiently with DCT loss, since the steps for updating the averaged model in SWAD may not be entirely suitable for triplet loss. The results of DCT with SWAD are shown in Table \ref{tabswad}. Despite this, the DCT loss with RegNet still can outperform on different domain generalization datasets like PACS, etc, which is shown in Table \ref{PACS}, \ref{officehome} and \ref{VLCS}. \par
\begin{table}[h]
    \centering
    \caption{Results of DCT and DCT with SWAD on PACS and OfficeHome on Resnet50.}
    \label{tabswad}
    \begin{tabular}{c c c}
    \hline
         &PACS&OfficeHome\\
         \hline
    DCT&87.3&70.3\\
    DCT+SWAD&87.7&70.2 \\
    \hline
    \end{tabular}

\end{table}
The motivation for exploring feature clustering in domain generalization is not trivial. With the appearance of cross-domain continual learning~\cite{MSL} and open-set domain generalization~\cite{opensetdg}, the combination of domain shift and semantic shift becomes a new topic in the field of out-of-distribution (OOD). As \cite{forward} mentions, the model should be not only domain-generalized but also growable and provident with proper compatibility for new classes, which can be achieved with compact feature clusters in the embedding space. So, we believe  cross-domain feature clustering is an important issue for the future of deep learning algorithms. 
\section{Conclusions}
\label{sec:concl}
In this paper, we give an explanation of feature clustering for domain generation.  A visualization of feature embedding at the initial training stage is also given to support the explanation. Motivated by this, we propose a novel pair mining method applied to triplet loss for domain generalization, which can lead to features generalized across different domains. Moreover, we also solve the weakness of feature clustering in positive pairs in domain generalization with two effective methods. Then, we show the effectiveness of our algorithm in feature clustering and compare our method with  other domain generalization methods on three datasets. Our DCT loss outperforms over all three datasets with the SWAG pretrained RegNet. We believe the insights behind this research on cross-domain feature clustering are interesting and promising. 









\bibliographystyle{cas-model2-names}
\bibliography{egbib}

\begin{thebibliography}{51}
\expandafter\ifx\csname natexlab\endcsname\relax\def\natexlab#1{#1}\fi
\providecommand{\url}[1]{\texttt{#1}}
\providecommand{\href}[2]{#2}
\providecommand{\path}[1]{#1}
\providecommand{\DOIprefix}{doi:}
\providecommand{\ArXivprefix}{arXiv:}
\providecommand{\URLprefix}{URL: }
\providecommand{\Pubmedprefix}{pmid:}
\providecommand{\doi}[1]{\href{http://dx.doi.org/#1}{\path{#1}}}
\providecommand{\Pubmed}[1]{\href{pmid:#1}{\path{#1}}}
\providecommand{\bibinfo}[2]{#2}
\ifx\xfnm\relax \def\xfnm[#1]{\unskip,\space#1}\fi
\bibitem[{Albuquerque et~al.(2019)Albuquerque, Monteiro, Falk and
  Mitliagkas}]{adv1}
\bibinfo{author}{Albuquerque, I.}, \bibinfo{author}{Monteiro, J.},
  \bibinfo{author}{Falk, T.H.}, \bibinfo{author}{Mitliagkas, I.},
  \bibinfo{year}{2019}.
\newblock \bibinfo{title}{Adversarial target-invariant representation learning
  for domain generalization}.
\newblock \bibinfo{journal}{ArXiv} \bibinfo{volume}{abs/1911.00804}.
\bibitem[{Arjovsky et~al.(2019)Arjovsky, Bottou, Gulrajani and Lopez-Paz}]{IRM}
\bibinfo{author}{Arjovsky, M.}, \bibinfo{author}{Bottou, L.},
  \bibinfo{author}{Gulrajani, I.}, \bibinfo{author}{Lopez-Paz, D.},
  \bibinfo{year}{2019}.
\newblock \bibinfo{title}{Invariant risk minimization}.
\newblock \bibinfo{journal}{arXiv preprint arXiv:1907.02893} .
\bibitem[{Blanchard et~al.(2011)Blanchard, Lee and Scott}]{domaingen}
\bibinfo{author}{Blanchard, G.}, \bibinfo{author}{Lee, G.},
  \bibinfo{author}{Scott, C.}, \bibinfo{year}{2011}.
\newblock \bibinfo{title}{Generalizing from several related classification
  tasks to a new unlabeled sample}, in: \bibinfo{editor}{Shawe-Taylor, J.},
  \bibinfo{editor}{Zemel, R.}, \bibinfo{editor}{Bartlett, P.},
  \bibinfo{editor}{Pereira, F.}, \bibinfo{editor}{Weinberger, K.} (Eds.),
  \bibinfo{booktitle}{Advances in Neural Information Processing Systems},
  \bibinfo{publisher}{Curran Associates, Inc.}
\newblock \URLprefix
  \url{https://proceedings.neurips.cc/paper/2011/file/b571ecea16a9824023ee1af16897a582-Paper.pdf}.
\bibitem[{Cha et~al.(2021)Cha, Chun, Lee, Cho, Park, Lee and Park}]{swad}
\bibinfo{author}{Cha, J.}, \bibinfo{author}{Chun, S.}, \bibinfo{author}{Lee,
  K.}, \bibinfo{author}{Cho, H.C.}, \bibinfo{author}{Park, S.},
  \bibinfo{author}{Lee, Y.}, \bibinfo{author}{Park, S.}, \bibinfo{year}{2021}.
\newblock \bibinfo{title}{Swad: Domain generalization by seeking flat minima}.
\newblock \bibinfo{journal}{Advances in Neural Information Processing Systems}
  \bibinfo{volume}{34}, \bibinfo{pages}{22405--22418}.
\bibitem[{Cha et~al.(2022)Cha, Lee, Park and Chun}]{miro}
\bibinfo{author}{Cha, J.}, \bibinfo{author}{Lee, K.}, \bibinfo{author}{Park,
  S.}, \bibinfo{author}{Chun, S.}, \bibinfo{year}{2022}.
\newblock \bibinfo{title}{Domain generalization by mutual-information
  regularization with pre-trained models}.
\newblock \bibinfo{journal}{European Conference on Computer Vision} .
\bibitem[{Chen et~al.(2021)Chen, Wang, Pan, Yao, Tian and Mei}]{steam}
\bibinfo{author}{Chen, Y.}, \bibinfo{author}{Wang, Y.}, \bibinfo{author}{Pan,
  Y.}, \bibinfo{author}{Yao, T.}, \bibinfo{author}{Tian, X.},
  \bibinfo{author}{Mei, T.}, \bibinfo{year}{2021}.
\newblock \bibinfo{title}{A style and semantic memory mechanism for domain
  generalization}, in: \bibinfo{booktitle}{Proceedings of the IEEE/CVF
  International Conference on Computer Vision}, pp.
  \bibinfo{pages}{9164--9173}.
\bibitem[{Deng et~al.(2009)Deng, Dong, Socher, Li, Li and Fei-Fei}]{imagenet}
\bibinfo{author}{Deng, J.}, \bibinfo{author}{Dong, W.},
  \bibinfo{author}{Socher, R.}, \bibinfo{author}{Li, L.J.},
  \bibinfo{author}{Li, K.}, \bibinfo{author}{Fei-Fei, L.},
  \bibinfo{year}{2009}.
\newblock \bibinfo{title}{Imagenet: A large-scale hierarchical image database},
  in: \bibinfo{booktitle}{2009 IEEE conference on computer vision and pattern
  recognition}, \bibinfo{organization}{Ieee}. pp. \bibinfo{pages}{248--255}.
\bibitem[{Dosovitskiy et~al.(2021)Dosovitskiy, Beyer, Kolesnikov, Weissenborn,
  Zhai, Unterthiner, Dehghani, Minderer, Heigold, Gelly, Uszkoreit and
  Houlsby}]{vit}
\bibinfo{author}{Dosovitskiy, A.}, \bibinfo{author}{Beyer, L.},
  \bibinfo{author}{Kolesnikov, A.}, \bibinfo{author}{Weissenborn, D.},
  \bibinfo{author}{Zhai, X.}, \bibinfo{author}{Unterthiner, T.},
  \bibinfo{author}{Dehghani, M.}, \bibinfo{author}{Minderer, M.},
  \bibinfo{author}{Heigold, G.}, \bibinfo{author}{Gelly, S.},
  \bibinfo{author}{Uszkoreit, J.}, \bibinfo{author}{Houlsby, N.},
  \bibinfo{year}{2021}.
\newblock \bibinfo{title}{An image is worth 16x16 words: Transformers for image
  recognition at scale}.
\newblock \bibinfo{journal}{ICLR} .
\bibitem[{Fang et~al.(2013)Fang, Xu and Rockmore}]{VLCS}
\bibinfo{author}{Fang, C.}, \bibinfo{author}{Xu, Y.},
  \bibinfo{author}{Rockmore, D.N.}, \bibinfo{year}{2013}.
\newblock \bibinfo{title}{Unbiased metric learning: On the utilization of
  multiple datasets and web images for softening bias}, in:
  \bibinfo{booktitle}{Proceedings of the IEEE International Conference on
  Computer Vision}, pp. \bibinfo{pages}{1657--1664}.
\bibitem[{Faraki et~al.(2021)Faraki, Yu, Tsai, Suh and Chandraker}]{CDT}
\bibinfo{author}{Faraki, M.}, \bibinfo{author}{Yu, X.}, \bibinfo{author}{Tsai,
  Y.H.}, \bibinfo{author}{Suh, Y.}, \bibinfo{author}{Chandraker, M.},
  \bibinfo{year}{2021}.
\newblock \bibinfo{title}{Cross-domain similarity learning for face recognition
  in unseen domains}, in: \bibinfo{booktitle}{Proceedings of the IEEE/CVF
  Conference on Computer Vision and Pattern Recognition}, pp.
  \bibinfo{pages}{15292--15301}.
\bibitem[{Gulrajani and Lopez{-}Paz(2020)}]{domainbed}
\bibinfo{author}{Gulrajani, I.}, \bibinfo{author}{Lopez{-}Paz, D.},
  \bibinfo{year}{2020}.
\newblock \bibinfo{title}{In search of lost domain generalization}.
\newblock \bibinfo{journal}{CoRR} \bibinfo{volume}{abs/2007.01434}.
\newblock \URLprefix \url{https://arxiv.org/abs/2007.01434},
  \href{http://arxiv.org/abs/2007.01434}{\tt arXiv:2007.01434}.
\bibitem[{Hadsell et~al.(2006)Hadsell, Chopra and LeCun}]{contrastiveloss}
\bibinfo{author}{Hadsell, R.}, \bibinfo{author}{Chopra, S.},
  \bibinfo{author}{LeCun, Y.}, \bibinfo{year}{2006}.
\newblock \bibinfo{title}{Dimensionality reduction by learning an invariant
  mapping}, in: \bibinfo{booktitle}{2006 IEEE Computer Society Conference on
  Computer Vision and Pattern Recognition (CVPR'06)}, pp.
  \bibinfo{pages}{1735--1742}.
\newblock \DOIprefix\doi{10.1109/CVPR.2006.100}.
\bibitem[{He et~al.(2016)He, Zhang, Ren and Sun}]{resnet}
\bibinfo{author}{He, K.}, \bibinfo{author}{Zhang, X.}, \bibinfo{author}{Ren,
  S.}, \bibinfo{author}{Sun, J.}, \bibinfo{year}{2016}.
\newblock \bibinfo{title}{Deep residual learning for image recognition}, in:
  \bibinfo{booktitle}{Proceedings of the IEEE conference on computer vision and
  pattern recognition}, pp. \bibinfo{pages}{770--778}.
\bibitem[{He et~al.(2021)He, Luo, Wang, Wang, Li and Jiang}]{transreid}
\bibinfo{author}{He, S.}, \bibinfo{author}{Luo, H.}, \bibinfo{author}{Wang,
  P.}, \bibinfo{author}{Wang, F.}, \bibinfo{author}{Li, H.},
  \bibinfo{author}{Jiang, W.}, \bibinfo{year}{2021}.
\newblock \bibinfo{title}{Transreid: Transformer-based object
  re-identification}, in: \bibinfo{booktitle}{Proceedings of the IEEE/CVF
  International Conference on Computer Vision (ICCV)}, pp.
  \bibinfo{pages}{15013--15022}.
\bibitem[{Hu et~al.(2020)Hu, Zhang, Chen and Chan}]{DMA}
\bibinfo{author}{Hu, S.}, \bibinfo{author}{Zhang, K.}, \bibinfo{author}{Chen,
  Z.}, \bibinfo{author}{Chan, L.}, \bibinfo{year}{2020}.
\newblock \bibinfo{title}{Domain generalization via multidomain discriminant
  analysis}, in: \bibinfo{booktitle}{Uncertainty in Artificial Intelligence},
  \bibinfo{organization}{PMLR}. pp. \bibinfo{pages}{292--302}.
\bibitem[{Huang et~al.(2020)Huang, Wang, Xing and Huang}]{rsc}
\bibinfo{author}{Huang, Z.}, \bibinfo{author}{Wang, H.}, \bibinfo{author}{Xing,
  E.P.}, \bibinfo{author}{Huang, D.}, \bibinfo{year}{2020}.
\newblock \bibinfo{title}{Self-challenging improves cross-domain
  generalization}, in: \bibinfo{booktitle}{European Conference on Computer
  Vision}, \bibinfo{organization}{Springer}. pp. \bibinfo{pages}{124--140}.
\bibitem[{Ioffe and Szegedy(2015)}]{batchnorm}
\bibinfo{author}{Ioffe, S.}, \bibinfo{author}{Szegedy, C.},
  \bibinfo{year}{2015}.
\newblock \bibinfo{title}{Batch normalization: Accelerating deep network
  training by reducing internal covariate shift}, in:
  \bibinfo{booktitle}{International conference on machine learning},
  \bibinfo{organization}{PMLR}. pp. \bibinfo{pages}{448--456}.
\bibitem[{Izmailov et~al.(2018)Izmailov, Podoprikhin, Garipov, Vetrov and
  Wilson}]{swa}
\bibinfo{author}{Izmailov, P.}, \bibinfo{author}{Podoprikhin, D.},
  \bibinfo{author}{Garipov, T.}, \bibinfo{author}{Vetrov, D.},
  \bibinfo{author}{Wilson, A.G.}, \bibinfo{year}{2018}.
\newblock \bibinfo{title}{Averaging weights leads to wider optima and better
  generalization}, in: \bibinfo{booktitle}{34th Conference on Uncertainty in
  Artificial Intelligence 2018, UAI 2018}, \bibinfo{organization}{Association
  For Uncertainty in Artificial Intelligence (AUAI)}. pp.
  \bibinfo{pages}{876--885}.
\bibitem[{Kantorovich(1960)}]{wdistance}
\bibinfo{author}{Kantorovich, L.V.}, \bibinfo{year}{1960}.
\newblock \bibinfo{title}{Mathematical methods of organizing and planning
  production}.
\newblock \bibinfo{journal}{Management science} \bibinfo{volume}{6},
  \bibinfo{pages}{366--422}.
\bibitem[{Katsumata et~al.(2021)Katsumata, Kishida, Amma and
  Nakayama}]{opensetdg}
\bibinfo{author}{Katsumata, K.}, \bibinfo{author}{Kishida, I.},
  \bibinfo{author}{Amma, A.}, \bibinfo{author}{Nakayama, H.},
  \bibinfo{year}{2021}.
\newblock \bibinfo{title}{Open-set domain generalization via metric learning},
  in: \bibinfo{booktitle}{2021 IEEE International Conference on Image
  Processing (ICIP)}, \bibinfo{organization}{IEEE}. pp.
  \bibinfo{pages}{459--463}.
\bibitem[{Kim et~al.(2021a)Kim, Yoo, Park, Kim and Lee}]{selfreg}
\bibinfo{author}{Kim, D.}, \bibinfo{author}{Yoo, Y.}, \bibinfo{author}{Park,
  S.}, \bibinfo{author}{Kim, J.}, \bibinfo{author}{Lee, J.},
  \bibinfo{year}{2021}a.
\newblock \bibinfo{title}{Selfreg: Self-supervised contrastive regularization
  for domain generalization}, in: \bibinfo{booktitle}{Proceedings of the
  IEEE/CVF International Conference on Computer Vision}, pp.
  \bibinfo{pages}{9619--9628}.
\bibitem[{Kim et~al.(2021b)Kim, Lee, Park, Min and Sohn}]{selfbalance}
\bibinfo{author}{Kim, J.}, \bibinfo{author}{Lee, J.}, \bibinfo{author}{Park,
  J.}, \bibinfo{author}{Min, D.}, \bibinfo{author}{Sohn, K.},
  \bibinfo{year}{2021}b.
\newblock \bibinfo{title}{Self-balanced learning for domain generalization},
  in: \bibinfo{booktitle}{2021 IEEE International Conference on Image
  Processing (ICIP)}, \bibinfo{organization}{IEEE}. pp.
  \bibinfo{pages}{779--783}.
\bibitem[{Li et~al.(2017)Li, Yang, Song and Hospedales}]{PACS}
\bibinfo{author}{Li, D.}, \bibinfo{author}{Yang, Y.}, \bibinfo{author}{Song,
  Y.Z.}, \bibinfo{author}{Hospedales, T.M.}, \bibinfo{year}{2017}.
\newblock \bibinfo{title}{Deeper, broader and artier domain generalization},
  in: \bibinfo{booktitle}{Proceedings of the IEEE international conference on
  computer vision}, pp. \bibinfo{pages}{5542--5550}.
\bibitem[{Li et~al.(2018)Li, Pan, Wang and Kot}]{mmd}
\bibinfo{author}{Li, H.}, \bibinfo{author}{Pan, S.J.}, \bibinfo{author}{Wang,
  S.}, \bibinfo{author}{Kot, A.C.}, \bibinfo{year}{2018}.
\newblock \bibinfo{title}{Domain generalization with adversarial feature
  learning}, in: \bibinfo{booktitle}{Proceedings of the IEEE conference on
  computer vision and pattern recognition}, pp. \bibinfo{pages}{5400--5409}.
\bibitem[{Luo et~al.(2019)Luo, Gu, Liao, Lai and Jiang}]{reidsb}
\bibinfo{author}{Luo, H.}, \bibinfo{author}{Gu, Y.}, \bibinfo{author}{Liao,
  X.}, \bibinfo{author}{Lai, S.}, \bibinfo{author}{Jiang, W.},
  \bibinfo{year}{2019}.
\newblock \bibinfo{title}{Bag of tricks and a strong baseline for deep person
  re-identification}, in: \bibinfo{booktitle}{2019 IEEE/CVF Conference on
  Computer Vision and Pattern Recognition Workshops (CVPRW)},
  \bibinfo{publisher}{IEEE Computer Society}, \bibinfo{address}{Los Alamitos,
  CA, USA}. pp. \bibinfo{pages}{1487--1495}.
\newblock \URLprefix
  \url{https://doi.ieeecomputersociety.org/10.1109/CVPRW.2019.00190},
  \DOIprefix\doi{10.1109/CVPRW.2019.00190}.
\bibitem[{Van~der Maaten and Hinton(2008)}]{tsne}
\bibinfo{author}{Van~der Maaten, L.}, \bibinfo{author}{Hinton, G.},
  \bibinfo{year}{2008}.
\newblock \bibinfo{title}{Visualizing data using t-sne.}
\newblock \bibinfo{journal}{Journal of machine learning research}
  \bibinfo{volume}{9}.
\bibitem[{Muandet et~al.(2013)Muandet, Balduzzi and Sch{\"o}lkopf}]{DICA}
\bibinfo{author}{Muandet, K.}, \bibinfo{author}{Balduzzi, D.},
  \bibinfo{author}{Sch{\"o}lkopf, B.}, \bibinfo{year}{2013}.
\newblock \bibinfo{title}{Domain generalization via invariant feature
  representation}, in: \bibinfo{booktitle}{International Conference on Machine
  Learning}, \bibinfo{organization}{PMLR}. pp. \bibinfo{pages}{10--18}.
\bibitem[{Nam et~al.(2021)Nam, Lee, Park, Yoon and Yoo}]{sagnet}
\bibinfo{author}{Nam, H.}, \bibinfo{author}{Lee, H.}, \bibinfo{author}{Park,
  J.}, \bibinfo{author}{Yoon, W.}, \bibinfo{author}{Yoo, D.},
  \bibinfo{year}{2021}.
\newblock \bibinfo{title}{Reducing domain gap by reducing style bias}, in:
  \bibinfo{booktitle}{Proceedings of the IEEE/CVF Conference on Computer Vision
  and Pattern Recognition}, pp. \bibinfo{pages}{8690--8699}.
\bibitem[{Pan and Yang(2009)}]{pan}
\bibinfo{author}{Pan, S.J.}, \bibinfo{author}{Yang, Q.}, \bibinfo{year}{2009}.
\newblock \bibinfo{title}{A survey on transfer learning}.
\newblock \bibinfo{journal}{IEEE Transactions on knowledge and data
  engineering} \bibinfo{volume}{22}, \bibinfo{pages}{1345--1359}.
\bibitem[{Paszke et~al.(2019)Paszke, Gross, Massa, Lerer, Bradbury, Chanan,
  Killeen, Lin, Gimelshein, Antiga et~al.}]{pytorch}
\bibinfo{author}{Paszke, A.}, \bibinfo{author}{Gross, S.},
  \bibinfo{author}{Massa, F.}, \bibinfo{author}{Lerer, A.},
  \bibinfo{author}{Bradbury, J.}, \bibinfo{author}{Chanan, G.},
  \bibinfo{author}{Killeen, T.}, \bibinfo{author}{Lin, Z.},
  \bibinfo{author}{Gimelshein, N.}, \bibinfo{author}{Antiga, L.}, et~al.,
  \bibinfo{year}{2019}.
\newblock \bibinfo{title}{Pytorch: An imperative style, high-performance deep
  learning library}.
\newblock \bibinfo{journal}{Advances in neural information processing systems}
  \bibinfo{volume}{32}.
\bibitem[{Rame et~al.(2022)Rame, Dancette and Cord}]{fishr}
\bibinfo{author}{Rame, A.}, \bibinfo{author}{Dancette, C.},
  \bibinfo{author}{Cord, M.}, \bibinfo{year}{2022}.
\newblock \bibinfo{title}{Fishr: Invariant gradient variances for
  out-of-distribution generalization}, in: \bibinfo{editor}{Chaudhuri, K.},
  \bibinfo{editor}{Jegelka, S.}, \bibinfo{editor}{Song, L.},
  \bibinfo{editor}{Szepesvari, C.}, \bibinfo{editor}{Niu, G.},
  \bibinfo{editor}{Sabato, S.} (Eds.), \bibinfo{booktitle}{Proceedings of the
  39th International Conference on Machine Learning},
  \bibinfo{publisher}{PMLR}. pp. \bibinfo{pages}{18347--18377}.
\bibitem[{Ren et~al.(2015)Ren, He, Girshick and Sun}]{fastrcnn}
\bibinfo{author}{Ren, S.}, \bibinfo{author}{He, K.}, \bibinfo{author}{Girshick,
  R.}, \bibinfo{author}{Sun, J.}, \bibinfo{year}{2015}.
\newblock \bibinfo{title}{Faster r-cnn: Towards real-time object detection with
  region proposal networks}.
\newblock \bibinfo{journal}{Advances in neural information processing systems}
  \bibinfo{volume}{28}.
\bibitem[{Schroff et~al.(2015)Schroff, Kalenichenko and Philbin}]{tripletloss}
\bibinfo{author}{Schroff, F.}, \bibinfo{author}{Kalenichenko, D.},
  \bibinfo{author}{Philbin, J.}, \bibinfo{year}{2015}.
\newblock \bibinfo{title}{Facenet: A unified embedding for face recognition and
  clustering}, in: \bibinfo{booktitle}{Proceedings of the IEEE conference on
  computer vision and pattern recognition}, pp. \bibinfo{pages}{815--823}.
\bibitem[{Seo et~al.(2020)Seo, Suh, Kim, Kim, Han and Han}]{dson}
\bibinfo{author}{Seo, S.}, \bibinfo{author}{Suh, Y.}, \bibinfo{author}{Kim,
  D.}, \bibinfo{author}{Kim, G.}, \bibinfo{author}{Han, J.},
  \bibinfo{author}{Han, B.}, \bibinfo{year}{2020}.
\newblock \bibinfo{title}{Learning to optimize domain specific normalization
  for domain generalization}, in: \bibinfo{booktitle}{European Conference on
  Computer Vision}, \bibinfo{organization}{Springer}. pp.
  \bibinfo{pages}{68--83}.
\bibitem[{Shankar et~al.(2018)Shankar, Piratla, Chakrabarti, Chaudhuri, Jyothi
  and Sarawagi}]{crossgrad}
\bibinfo{author}{Shankar, S.}, \bibinfo{author}{Piratla, V.},
  \bibinfo{author}{Chakrabarti, S.}, \bibinfo{author}{Chaudhuri, S.},
  \bibinfo{author}{Jyothi, P.}, \bibinfo{author}{Sarawagi, S.},
  \bibinfo{year}{2018}.
\newblock \bibinfo{title}{Generalizing across domains via cross-gradient
  training}.
\newblock \bibinfo{journal}{arXiv preprint arXiv:1804.10745} .
\bibitem[{Simon et~al.(2022)Simon, Faraki, Tsai, Yu, Schulter, Suh, Harandi and
  Chandraker}]{MSL}
\bibinfo{author}{Simon, C.}, \bibinfo{author}{Faraki, M.},
  \bibinfo{author}{Tsai, Y.H.}, \bibinfo{author}{Yu, X.},
  \bibinfo{author}{Schulter, S.}, \bibinfo{author}{Suh, Y.},
  \bibinfo{author}{Harandi, M.}, \bibinfo{author}{Chandraker, M.},
  \bibinfo{year}{2022}.
\newblock \bibinfo{title}{On generalizing beyond domains in cross-domain
  continual learning}, in: \bibinfo{booktitle}{Proceedings of the IEEE/CVF
  Conference on Computer Vision and Pattern Recognition}, pp.
  \bibinfo{pages}{9265--9274}.
\bibitem[{Singh et~al.(2022)Singh, Gustafson, Adcock, de~Freitas~Reis, Gedik,
  Kosaraju, Mahajan, Girshick, Doll{\'a}r and van~der Maaten}]{regnet}
\bibinfo{author}{Singh, M.}, \bibinfo{author}{Gustafson, L.},
  \bibinfo{author}{Adcock, A.}, \bibinfo{author}{de~Freitas~Reis, V.},
  \bibinfo{author}{Gedik, B.}, \bibinfo{author}{Kosaraju, R.P.},
  \bibinfo{author}{Mahajan, D.}, \bibinfo{author}{Girshick, R.},
  \bibinfo{author}{Doll{\'a}r, P.}, \bibinfo{author}{van~der Maaten, L.},
  \bibinfo{year}{2022}.
\newblock \bibinfo{title}{Revisiting weakly supervised pre-training of visual
  perception models}, in: \bibinfo{booktitle}{Proceedings of the IEEE/CVF
  Conference on Computer Vision and Pattern Recognition}, pp.
  \bibinfo{pages}{804--814}.
\bibitem[{Sun and Saenko(2016)}]{coral}
\bibinfo{author}{Sun, B.}, \bibinfo{author}{Saenko, K.}, \bibinfo{year}{2016}.
\newblock \bibinfo{title}{Deep coral: Correlation alignment for deep domain
  adaptation}, in: \bibinfo{booktitle}{European conference on computer vision},
  \bibinfo{organization}{Springer}. pp. \bibinfo{pages}{443--450}.
\bibitem[{Vapnik(1999)}]{erm}
\bibinfo{author}{Vapnik, V.N.}, \bibinfo{year}{1999}.
\newblock \bibinfo{title}{An overview of statistical learning theory}.
\newblock \bibinfo{journal}{IEEE transactions on neural networks}
  \bibinfo{volume}{10}, \bibinfo{pages}{988--999}.
\bibitem[{Venkateswara et~al.(2017)Venkateswara, Eusebio, Chakraborty and
  Panchanathan}]{officehome}
\bibinfo{author}{Venkateswara, H.}, \bibinfo{author}{Eusebio, J.},
  \bibinfo{author}{Chakraborty, S.}, \bibinfo{author}{Panchanathan, S.},
  \bibinfo{year}{2017}.
\newblock \bibinfo{title}{Deep hashing network for unsupervised domain
  adaptation}, in: \bibinfo{booktitle}{Proceedings of the IEEE conference on
  computer vision and pattern recognition}, pp. \bibinfo{pages}{5018--5027}.
\bibitem[{Wang et~al.(2022)Wang, Lan, Liu, Ouyang, Qin, Lu, Chen, Zeng and
  Yu}]{dgsurvey}
\bibinfo{author}{Wang, J.}, \bibinfo{author}{Lan, C.}, \bibinfo{author}{Liu,
  C.}, \bibinfo{author}{Ouyang, Y.}, \bibinfo{author}{Qin, T.},
  \bibinfo{author}{Lu, W.}, \bibinfo{author}{Chen, Y.}, \bibinfo{author}{Zeng,
  W.}, \bibinfo{author}{Yu, P.}, \bibinfo{year}{2022}.
\newblock \bibinfo{title}{Generalizing to unseen domains: A survey on domain
  generalization}.
\newblock \bibinfo{journal}{IEEE Transactions on Knowledge and Data
  Engineering} .
\bibitem[{Wang et~al.(2020)Wang, Wang, Lv, Cao and Fu}]{adv2}
\bibinfo{author}{Wang, Z.}, \bibinfo{author}{Wang, Q.}, \bibinfo{author}{Lv,
  C.}, \bibinfo{author}{Cao, X.}, \bibinfo{author}{Fu, G.},
  \bibinfo{year}{2020}.
\newblock \bibinfo{title}{Unseen target stance detection with adversarial
  domain generalization}, in: \bibinfo{booktitle}{2020 International Joint
  Conference on Neural Networks (IJCNN)}, \bibinfo{organization}{IEEE}. pp.
  \bibinfo{pages}{1--8}.
\bibitem[{Wen et~al.(2016)Wen, Zhang, Li and Qiao}]{centerloss}
\bibinfo{author}{Wen, Y.}, \bibinfo{author}{Zhang, K.}, \bibinfo{author}{Li,
  Z.}, \bibinfo{author}{Qiao, Y.}, \bibinfo{year}{2016}.
\newblock \bibinfo{title}{A discriminative feature learning approach for deep
  face recognition}, in: \bibinfo{booktitle}{European conference on computer
  vision}, \bibinfo{organization}{Springer}. pp. \bibinfo{pages}{499--515}.
\bibitem[{Wu et~al.(2018)Wu, Xiong, Yu and Lin}]{instancediscrimination}
\bibinfo{author}{Wu, Z.}, \bibinfo{author}{Xiong, Y.}, \bibinfo{author}{Yu,
  S.X.}, \bibinfo{author}{Lin, D.}, \bibinfo{year}{2018}.
\newblock \bibinfo{title}{Unsupervised feature learning via non-parametric
  instance discrimination}, in: \bibinfo{booktitle}{Proceedings of the IEEE
  conference on computer vision and pattern recognition}, pp.
  \bibinfo{pages}{3733--3742}.
\bibitem[{Yao et~al.(2022)Yao, Bai, Zhang, Zhang, Sun, Chen, Li and Yu}]{pcl}
\bibinfo{author}{Yao, X.}, \bibinfo{author}{Bai, Y.}, \bibinfo{author}{Zhang,
  X.}, \bibinfo{author}{Zhang, Y.}, \bibinfo{author}{Sun, Q.},
  \bibinfo{author}{Chen, R.}, \bibinfo{author}{Li, R.}, \bibinfo{author}{Yu,
  B.}, \bibinfo{year}{2022}.
\newblock \bibinfo{title}{Pcl: Proxy-based contrastive learning for domain
  generalization}, in: \bibinfo{booktitle}{Proceedings of the IEEE/CVF
  Conference on Computer Vision and Pattern Recognition}, pp.
  \bibinfo{pages}{7097--7107}.
\bibitem[{Zhang et~al.(2018)Zhang, Cisse, Dauphin and Lopez-Paz}]{mixup}
\bibinfo{author}{Zhang, H.}, \bibinfo{author}{Cisse, M.},
  \bibinfo{author}{Dauphin, Y.N.}, \bibinfo{author}{Lopez-Paz, D.},
  \bibinfo{year}{2018}.
\newblock \bibinfo{title}{mixup: Beyond empirical risk minimization}, in:
  \bibinfo{booktitle}{International Conference on Learning Representations}.
\bibitem[{Zhang et~al.(2020)Zhang, Marklund, Dhawan, Gupta, Levine and
  Finn}]{arm}
\bibinfo{author}{Zhang, M.}, \bibinfo{author}{Marklund, H.},
  \bibinfo{author}{Dhawan, N.}, \bibinfo{author}{Gupta, A.},
  \bibinfo{author}{Levine, S.}, \bibinfo{author}{Finn, C.},
  \bibinfo{year}{2020}.
\newblock \bibinfo{title}{Adaptive risk minimization: A meta-learning approach
  for tackling group distribution shift}.
\newblock \bibinfo{journal}{arXiv preprint arXiv:2007.02931}
  \bibinfo{volume}{1}.
\bibitem[{Zhao et~al.(2020)Zhao, Gong, Liu, Fu and Tao}]{ER}
\bibinfo{author}{Zhao, S.}, \bibinfo{author}{Gong, M.}, \bibinfo{author}{Liu,
  T.}, \bibinfo{author}{Fu, H.}, \bibinfo{author}{Tao, D.},
  \bibinfo{year}{2020}.
\newblock \bibinfo{title}{Domain generalization via entropy regularization},
  in: \bibinfo{editor}{Larochelle, H.}, \bibinfo{editor}{Ranzato, M.},
  \bibinfo{editor}{Hadsell, R.}, \bibinfo{editor}{Balcan, M.},
  \bibinfo{editor}{Lin, H.} (Eds.), \bibinfo{booktitle}{Advances in Neural
  Information Processing Systems}, \bibinfo{publisher}{Curran Associates,
  Inc.}. pp. \bibinfo{pages}{16096--16107}.
\newblock \URLprefix
  \url{https://proceedings.neurips.cc/paper/2020/file/b98249b38337c5088bbc660d8f872d6a-Paper.pdf}.
\bibitem[{Zhou et~al.(2022)Zhou, Wang, Ye, Ma, Pu and Zhan}]{forward}
\bibinfo{author}{Zhou, D.W.}, \bibinfo{author}{Wang, F.Y.},
  \bibinfo{author}{Ye, H.J.}, \bibinfo{author}{Ma, L.}, \bibinfo{author}{Pu,
  S.}, \bibinfo{author}{Zhan, D.C.}, \bibinfo{year}{2022}.
\newblock \bibinfo{title}{Forward compatible few-shot class-incremental
  learning}, in: \bibinfo{booktitle}{Proceedings of the IEEE/CVF Conference on
  Computer Vision and Pattern Recognition}, pp. \bibinfo{pages}{9046--9056}.
\bibitem[{Zhou et~al.(2021)Zhou, Yang, Qiao and Xiang}]{mixstyle}
\bibinfo{author}{Zhou, K.}, \bibinfo{author}{Yang, Y.}, \bibinfo{author}{Qiao,
  Y.}, \bibinfo{author}{Xiang, T.}, \bibinfo{year}{2021}.
\newblock \bibinfo{title}{Domain generalization with mixstyle}.
\newblock \bibinfo{journal}{arXiv preprint arXiv:2104.02008} .
\bibitem[{Zou et~al.(2020)Zou, Yang, Yu, Kumar and Kautz}]{DG-net-PP}
\bibinfo{author}{Zou, Y.}, \bibinfo{author}{Yang, X.}, \bibinfo{author}{Yu,
  Z.}, \bibinfo{author}{Kumar, B.}, \bibinfo{author}{Kautz, J.},
  \bibinfo{year}{2020}.
\newblock \bibinfo{title}{Joint disentangling and adaptation for cross-domain
  person re-identification}, in: \bibinfo{booktitle}{European Conference on
  Computer Vision}, \bibinfo{organization}{Springer}. pp.
  \bibinfo{pages}{87--104}.

\end{thebibliography}



\end{document}